\pdfoutput=1

\documentclass[11pt]{article}

\usepackage[]{acl}

\usepackage{times}
\usepackage{latexsym}

\usepackage[T1]{fontenc}

\usepackage[utf8]{inputenc}

\usepackage{microtype}

\usepackage{inconsolata}

\usepackage{hyperref}       
\usepackage{url}            
\usepackage{breakurl}
\usepackage{booktabs}       
\usepackage{amsfonts}       
\usepackage{nicefrac}       
\usepackage{microtype}      
\usepackage{todonotes}
\usepackage[scaled]{beramono} 
\usepackage{algpseudocode}
\usepackage{algorithm}
\usepackage{pifont}
\usepackage{balance}
\usepackage{xspace}

\newcommand{\model}{mTAD\xspace}

\usepackage{amsthm}
\usepackage{amsmath}
\usepackage{MnSymbol}%
\usepackage{subcaption}
\usepackage{graphicx}
\usepackage{longtable}

\usepackage{multirow}
\usepackage[scaled]{beramono} 

\usepackage[normalem]{ulem}
\useunder{\uline}{\ul}{}

\definecolor{verbosity_high}{RGB}{102, 153, 204}
\definecolor{verbosity_low}{RGB}{153, 204, 255}

\definecolor{fluency_high}{RGB}{204, 143, 82}
\definecolor{fluency_low}{RGB}{255, 179, 102}

\definecolor{default}{RGB}{240, 240, 240}
\definecolor{user}{RGB}{255, 255, 255}

\definecolor{intent}{RGB}{153, 204, 255}
\definecolor{profile}{RGB}{255, 179, 102}
\definecolor{llm}{RGB}{190, 190, 190}

\colorlet{default}{Azure2}
\colorlet{concise}{LightSkyBlue1}
\colorlet{fluent}{DarkSeaGreen2}
\colorlet{verbose}{Plum2}
\colorlet{tolerance}{Burlywood2}
\colorlet{cooperativeness}{OrangeRed1}
\colorlet{explorative}{LightGoldenrod1}
\colorlet{engagement}{DarkOrange1}
\colorlet{emotion}{Aquamarine1}
\colorlet{repetition}{Aquamarine3}

\title{Multi-trait User Simulation with Adaptive Decoding \\ for Conversational Task Assistants}

\author{Rafael Ferreira \and David Semedo \and João Magalhães \\
         NOVA University of Lisbon \and NOVA LINCS
         \\ \texttt{rah.ferreira@campus.fct.unl.pt}
         \\ \texttt{{\{df.semedo,jm.magalhaes\}}@fct.unl.pt}}

\begin{document}
\maketitle

\begin{abstract}
Conversational systems must be robust to user interactions that naturally exhibit diverse conversational traits. Capturing and simulating these diverse traits coherently and efficiently presents a complex challenge. 
This paper introduces Multi-Trait Adaptive Decoding (\textit{\model}), a method that generates diverse user profiles at decoding-time by sampling from various trait-specific Language Models (LMs). \textit{\model} provides an adaptive and scalable approach to user simulation, enabling the creation of multiple user profiles without the need for additional fine-tuning.
By analyzing real-world dialogues from the Conversational Task Assistant (CTA) domain, we identify key conversational traits and developed a framework to generate profile-aware dialogues that enhance conversational diversity.
Experimental results validate the effectiveness of our approach in modeling single-traits using specialized LMs, which can capture less common patterns, even in out-of-domain tasks. Furthermore, the results demonstrate that \textit{\model} is a robust and flexible framework for combining diverse user simulators.

\end{abstract}

\section{Introduction}
\label{sec_intro}
Thoroughly testing a conversational system with real users is a costly and time consuming process. Part of this cost lies in the myriad of conversational traits, reflecting user's behaviors, knowledge, and goals, which when diverse have shown to help improve performance of conversational systems~\cite{user_diversified_policy, simulator_multiwoz}.
User simulators have been a successful approach to model real user conversational traits and discover errors and limitations in conversational systems~\cite{simulator_multiwoz, schema_guided_dataset, cannot_stand_everyone_tod, kuai_sim_reccomender}.
Prior research investigated user simulators across diverse settings, including task-oriented dialogues~\cite{transformer_user_simulator, simulator_multiwoz, cannot_stand_everyone_tod} and recommendation~\cite{simulator_recommendation_systems, simulator_recommendation_toolkit, kuai_sim_reccomender} contexts.
However, designing user simulators that can effectively engage with a system is challenging,  due to the need for adaptability and controllability across various user conversational patterns.

\begin{table}[t]
    \centering
    \texttt{
    \tiny
    \begin{tabular}{|p{0.45\linewidth}p{0.45\linewidth}|}
    \hline 
      & \\
    \textbf{User Dialogue Traits} & \textbf{User Utterance Traits}\\ 
      & \\
    ${\textcolor{engagement}{\blacksquare}}$ Engagement & $\textcolor{verbose}{\blacksquare}$ Verbosity \\
    $\textcolor{cooperativeness}{\blacksquare}$ Cooperativeness & $\textcolor{emotion}{\blacksquare}$ Emotion \\ 
    $\textcolor{explorative}{\blacksquare}$ Exploration & $\textcolor{fluent}{\blacksquare}$ Fluency \\
    $\textcolor{tolerance}{\blacksquare}$ Tolerance & $\textcolor{repetition}{\blacksquare}$ Repetition\\
    & \\
    \hline
    \end{tabular}%
    \\
    \vspace{2mm}
    \tiny
    \begin{tabular}{|p{0.10\linewidth}p{0.80\linewidth}|}
    \hline
    & \\
    \textbf{Task:} & \textbf{Fresh homemade pasta} \\
    &\\
    \colorbox{default}{\textbf{USimul:}} & let's start    \\
    \colorbox{user}{\textbf{System:}} & let's get started! Step 1: Sift the flour and salt together on a clean working table.   \\ 
    &\multicolumn{1}{l|}{\ldots}     \\ 
    \colorbox{verbose}{\textbf{USimul:}} & next    \\
    \colorbox{user}{\textbf{System:}} & Step 3: Crack the large egg into the flour bowl and whish with a fork. \\ 
    \colorbox{fluent}{\textbf{USimul:}} & uhh read step again    \\
    \colorbox{user}{\textbf{System:}} & Sure, Step 3: Crack the large egg into the flour bowl and whish with a fork.   \\ 
    &\multicolumn{1}{l|}{\ldots}     \\ 
    \colorbox{verbose}{\textbf{USimul:}} & tell me the next step please    \\ 
    \colorbox{user}{\textbf{System:}} & Step 4: slowly add the flour and keep whisking until the flour has combined with the eggs.   \\ 
    \colorbox{explorative}{\textbf{USimul:}} & Why should I add the flour slowly\\ 
    \colorbox{user}{\textbf{System:}} & Adding the flour slowly, helps ensure a smooth and even mixture.  \\ 
    &\multicolumn{1}{l|}{\ldots}     \\ 
    \colorbox{cooperativeness}{\textbf{USimul:}} & what is your favorite pokemon    \\ 
    \colorbox{user}{\textbf{System:}} & Let's try to stay focused on the task.  \\ 
    &\multicolumn{1}{l|}{\ldots}     \\ 
    & \\ 
    \hline
    \end{tabular}%
    }
    \caption{Simulated dialogue example illustrating the impact of different user traits on conversational assistants, testing their ability to manage a variety of user behaviors.
    }
    \vspace{-10pt}
    \label{tab_sample_dialogue}
\end{table}

Simulating a user conversational profile entails a combination of diverse conversational traits, as shown in Table~\ref{tab_sample_dialogue}. 
In this work, we follow model-based approaches~\cite{cannot_stand_everyone_tod, transformer_user_simulator, simulator_multiwoz} by learning trait-specific Language Models (LMs) and combining them into multiple different user profiles.
Approaches to combine models include weight level approaches~\cite{weight_averaging_merging, task_arithmetic_merging, dare_merging, ties_merging}, and trainable Mixture-of-Experts (MoE)~\cite{mixture_of_loras, llava_mole, mixture_of_lora_experts, mixtral}.
Instead of combining LMs before inference time, we propose an adaptive and scalable method that combines user traits at decoding-time by sampling from distributions from each trait-specific LM.
With the proposed multi-Trait Adaptive Decoding method (\textit{\model}), we are able to sidestep the need for combinatory training datasets or extra model fine-tuning. In addition, new LM traits can be adaptively added to the pool of traits creating a new user profile, without the need to retrain existing LMs.
As we show, combining these traits is crucial for fostering more diverse conversational patterns. 

To identify these traits, we analyzed real-world dialogues in the novel Conversational Task Assistant (CTA) domain~\cite{taskbot_overview_year_1, taskbot_overview_year_2} and extracted the most relevant traits. 
In CTA scenarios, users actively engage in completing manual tasks (e.g., baking a cake) with the system's assistance, fostering mixed-initiative dialogues that pose unique user modeling challenges.

In summary, one of the core contributions of this paper is the \textbf{Multi-Trait Adaptive Decoding method (\textit{\model})}, which allows for the combination of LMs to simulate multiple different user profiles at decoding time, removing the need for combinatory data or additional fine-tuning.
The second contribution is the introduction of a set of \textbf{user conversational traits} at dialogue-level and utterance-level, derived from \textbf{real-world CTA data} collected during the Alexa TaskBot Challenge~\cite{taskbot_overview_year_1}.
Experimental results support the need for specialized LMs to accurately represent each trait. Additionally, the results show the scalability and performance of \textit{\model}, which can flexibly simulate users with arbitrary combinations of traits, without user profile-specific training.

\section{Related Work}
\label{sec_related_work}

\paragraph{Conversational Assistants}
Prior work has focused on task-oriented~\cite{multiwoz_original, schema_guided_dataset} and recommendation systems~\cite{recommendation_paper, path_reasoning_recommendation}, where the assistant performs tasks based on user input (e.g. buying a ticket). Our work departs from these traditional settings and explores CTAs~\cite{taskbot_overview_year_1}, where users complete tasks (e.g., cooking) with help from the assistant, presenting unique modeling challenges~\cite{mango_mango}, which we investigate in the context of user simulation.

\paragraph{User simulators}
Initial works~\cite{agenda_based_user_sim_original, agenda_based_2016} used rule-based agendas to model user actions. More recently, model-based approaches~\cite{simulator_multiwoz}, in specific with LLMs~\cite{transformer_user_simulator,simulators_in_context_learning,cannot_stand_everyone_tod} have been used. Additionally, diverse user simulators have shown to improve system performance~\cite{cannot_stand_everyone_tod,user_diversified_policy}. We believe more attention should be given to modeling specific user traits, which we address by modeling user profiles as combinations of specialized LMs.

\paragraph{Model Combination}
User simulators must exhibit diverse conversational patterns to effectively evaluate conversational systems. Model merging techniques~\cite{ties_merging, dare_merging, task_arithmetic_merging} combine models at a weight level, enhancing their capabilities across various NLP tasks. 
Alternatively, mixture-of-experts approaches~\cite{mole_experts, mixture_of_loras, mixtral} use a trainable router in the Transformer~\cite{vaswani_attention} to integrate information from multiple models. 
In a different vein \citet{classifier_guided_decoding} uses a classifier to adjust token probabilities, while \citet{collaborative_decoding} applies a classifier to determine when to activate the decoding process of different models.
In this work, we propose \textit{\model}, a controllable and adaptable method that combines token distributions from various models at decoding time without requiring additional fine-tuning.

\section{Adaptable Multi-Trait User Simulators}
\label{sec_user_simulator}
End-to-end simulators~\cite{transformer_user_simulator, simulate_users_tod} may overlook less common styles and language subtleties due to smoothing or potential forgetting~\cite{forgetting_llms}, while being limited w.r.t. generalization to novel traits. 
Therefore, we consider a trait-oriented model-based approach, in which specialized trait simulators are created and flexibly combined at decoding time.

\subsection{Trait-Specialized User Simulators}
\label{sub_user_profile_definition}
Given a dialogue domain $\mathcal{M}$ (e.g. TOD or CTA), we define a conversational trait $t_i$ in a discrete three-intensity level range $l_i \in \{low, \ neutral, \ high\}$. Each trait-intensity pair $(t_i,l_i)$  has an associated dialogue language modeling distribution:
\begin{equation}
    P_{\theta_{(t_i,l_i)}}(w_j|w_{<j}, H, \mathcal{M}, (t_i, l_i)),
\label{eq:trait_prob}
\end{equation} where $w_j$ is the next token to be generated, $H$ the dialogue history, and $\theta_{(t_i,l_i)}$ the distribution's parameters. Traits can encompass any conversational characteristic measurable across a dialogue, such as cooperativeness and fluency -- in Section~\ref{sec_simulator_cta}, we introduce the set of traits used in this work.
This formulation, allows us to categorize and simulate diverse user traits at different levels of intensity.

Given the set of all traits $T$, a user profile $U$ is defined as the set of traits: 
\begin{equation}
    U = \{(t_1, l_1), \ldots, (t_{|T|},l_{|T|})\}. 
\end{equation}
As an example, we can define an \textit{uncooperative} but \textit{fluent} user profile: $U_{e.g.} = \{(t_{cooperativeness}, low), \ (t_{fluency},high)\}$)\footnote{For simplicity, we omit all traits with $neutral$ intensity.}. 

A user profile $U$ is thus modeled as the conditional probability distribution:
\begin{equation}
P_{\theta_U}(w_j|w_{<j}, H, \mathcal{M}, \{(t_i,l_i)\}_{i=1}^{i=|T|}),
\end{equation}
with $\theta_U$ being the profile distribution parameters.
Our proposed simulator is designed to maximize the expectation $\mathbb{E}[P_{\theta_U}]$ in a zero-shot manner.

\subsection{User Simulator LMs}
\label{sub_model_formalization}

Due to the combinatory nature of user profiles, maximizing the expectation $\mathbb{E}[P_{\theta_U}]$ on demand, for every possible trait-combination, is infeasible.  
Instead, to provide an adaptable and controllable user simulator, given a target user profile $U$, we focus on  learning individual traits distributions $P_{\theta_{(t_i,l_i)}}$, and combine them at inference time to maximize $\mathbb{E}[P_{\theta_U}]$. 
This is accomplished with \textit{\model}, our proposed zero-shot trait-combination strategy (Section~\ref{sub_sub_mtd}).

\paragraph{Approximating $\boldsymbol{P_{\theta_{(t,l)}}}$ with a LoRA-based Specialized Trait Simulator (STS).}
The goal is to have one Specialized-Trait Simulator (\textit{STS}) per trait. 
This enables each model to capture the subtleties of specific traits, with minimal interference from others, by learning an independent distribution $P_{\theta_{(t_i,l_i)}}$ for each one. 
In addition, given our focus on delivering adaptable simulators, this strategy makes the incorporation of new traits seamless since each model operates independently.

To model the distribution $P_{\theta_{(t_i,l_i)}}$, we adopt a model-based approach~\cite{transformer_user_simulator, simulate_users_tod}, using an LLM to capture each trait.
In practice, maximizing the log-likelihood $P_{\theta_{(t_i,l_i)}}$ corresponds to minimizing the LLM's language modeling cross-entropy over trait-specific dialogues.

Moreover, to efficiently support the learning of multiple independent $P_{\theta_{(t_i,l_i)}}$ distributions, using specialized trait simulators, we use LoRA (Low-Rank Adaptation)~\cite{lora, qlora} adapters for each trait. 
With LoRA, we only update the low-rank matrices in specific layers while keeping the original model weights frozen. This reduces trainable parameters, enabling faster training and lower memory use. These characteristics make LoRA ideal for \textit{\model}, allowing diverse trait simulators on a shared LLM backbone at decoding time.

\begin{figure}[!t]
    \centering
    \includegraphics[width=1.0\linewidth]{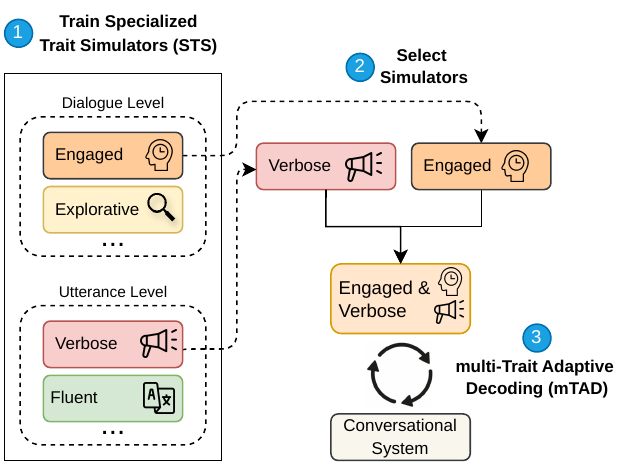}
    \caption{Multi-trait Adaptive Decoding (\textit{\model}), leveraging two Specialized Trait Simulators.}
    \label{fig_pipeline_image}
\end{figure}

\subsection{Approximating \texorpdfstring{$\boldsymbol{P_{\theta_U}}$}{P\_theta\_U} with Multi-Trait Adaptive Decoding (\textit{\model})}
\label{sub_sub_mtd}
To deliver an on-demand simulator of a user profile $U$, with a set of traits and corresponding intensities, while avoiding additional fine-tuning, we propose to approximate $P_{\theta_U}$  as a combination of independent trait distributions $P_{\theta_{(t_i,l_i)}}$, and implicitly maximize $\mathbb{E}[P_{\theta_U}]$. 
Namely, we propose \textbf{M}ulti-\textbf{T}rait \textbf{A}daptive \textbf{D}ecoding (\textit{\model}), to combine multiple user traits at decoding level as the following:
\begin{equation}
\begin{split}
    P_{\theta_U}&(w_j|w_{<j}, H, \mathcal{M}, (t_i,l_i)_{i=1}^{i=|T|}) = \\ &\sum_{i=1}^{|T|} \lambda_i\cdot P_{\theta_{(t_i,l_i)}} (w_j|w_{<j}, H, \mathcal{M}, (t_i, l_i)),
\end{split}
\end{equation}
where $\lambda_i$ are tunable trait weight parameters, which offer controllability in the profile modeling process, where one can easily select and activate each specialized trait simulator (\textit{STS}).

In practice, each $P_{\theta_{(t_i,l_i)}}$ is modeled by a language model $M_i$, where, to ensure compatibility, all models $M$ share the same vocabulary. 
During decoding, at each step, given the distribution $P_{\theta_{U}}$, we sample tokens using a given decoding strategy.
This implies sampling each specialized trait LM independently for each decoding step, which has minimal requirements due to the selective activation of different LoRA adapters.
Figure~\ref{fig_pipeline_image} represents an overview of the \textit{\model} framework.

\subsection{User Simulator Grounding}
\label{sub_single_user}
To materialize the probability distribution of Eq.~\ref{eq:trait_prob} and ground the behavior of each \textit{STS}, we define the input sequence:
\begin{equation}
    P \oplus H \oplus U \oplus S,
\end{equation}
where $\oplus$ denotes the concatenation operation, $P$ is an optional preamble, which varies based on the model used~\cite{vicuna2023}, $H$ represents the dialogue history, including $n$ previous turns (each consisting of a user and system utterance), $U$ is the user profile, encoded as a unique token sequence, and $S$ is a suffix used to prompt generation.

The learning objective becomes the causal language modeling task, which corresponds to minimizing the loss:
\begin{equation}
    \mathcal{L} = -\sum_{j=1}^{N} \log (P_{\theta_{(t_i, l_i)}}(w_j | w_{<j}, P, H, U, S)),
\end{equation}
where $j$ is the $j$-th token and $N$ the number of tokens in the response, which comprises both the a user intent and the utterance.
The model is trained on this dual-generation task to enhance interpretability by enabling the analysis of intent distribution probabilities.
In Appendix~\ref{app_input_format}, we present an example of the input and response formats. 

\section{User Simulation in Conversational Task Assistants (CTAs)}
\label{sec_simulator_cta}

Conversational Task Assistants (CTAs) guide users through tasks such as cooking or DIY~\cite{taskbot_overview_year_1, taskbot_overview_year_2}. 
This setting raises a number of challenges~\cite{mango_mango, wizard_of_tasks}, that are particularly well addressed by \textit{\model}: dialogues have mixed-initiative, users follow a task plan with the aim of completing a manual task, ask explorative questions, and engage in chit-chat as shown in Table~\ref{tab_sample_dialogue}.

\subsection{Real World Conversational Data}
To ensure the most representative user simulator, we ground the trait and models on real-world dialogue data. Specifically, we used the dataset from~\citet{plangpt} composed of 3.6k conversations collected during the Alexa TaskBot Challenge~\cite{taskbot_overview_year_1}. 
This dataset is composed of a generated dialogue graph based on intent transitions and ASR transcribed real user utterances.
We extend this pipeline to include a diverse profile-aware creation process detailed in Appendix~\ref{app_dataset_creation}.

\subsection{Modeling User Traits}
\label{sub_model_user_traits}
To define the set of traits ($T$) characterizing a user profile ($U$), we carefully analyzed human-system conversations (for statistics, refer to Appendix~\ref{app_real_data_stats}), identifying two categories at different levels: 1) \textit{Dialog-level}, and 2) \textit{Utterance-level}.

\begin{figure*}[t]
    \centering
    \begin{subfigure}{0.24\textwidth}
        \centering
        \includegraphics[width=\linewidth]{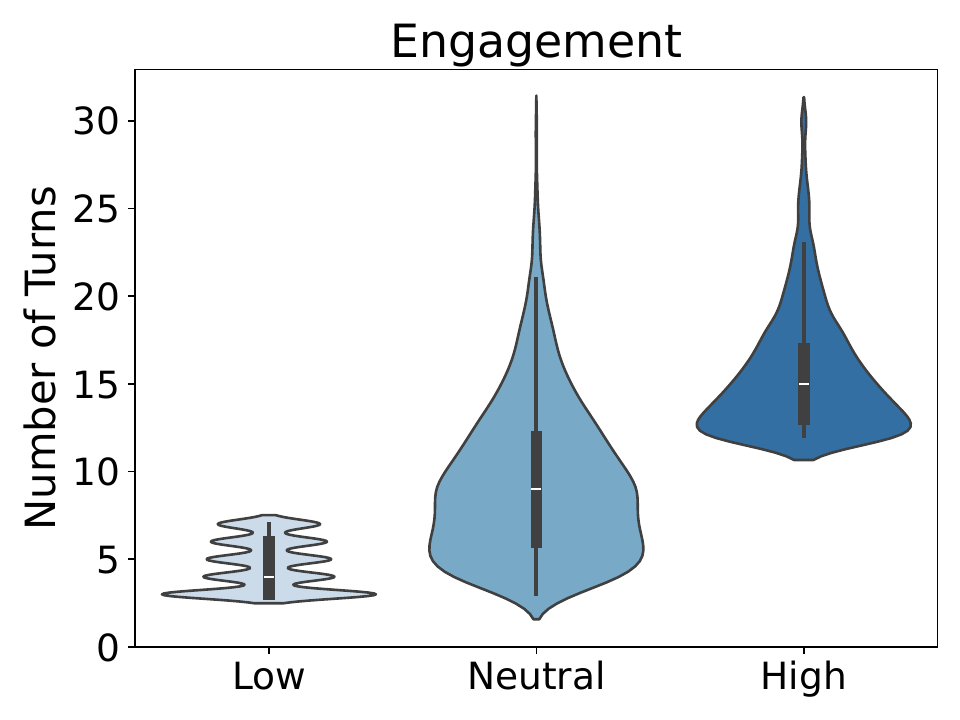}
        \vspace{-6mm}
        \label{fig_patient_violin}
    \end{subfigure}
    \begin{subfigure}{0.24\textwidth}
        \centering
        \includegraphics[width=\linewidth]{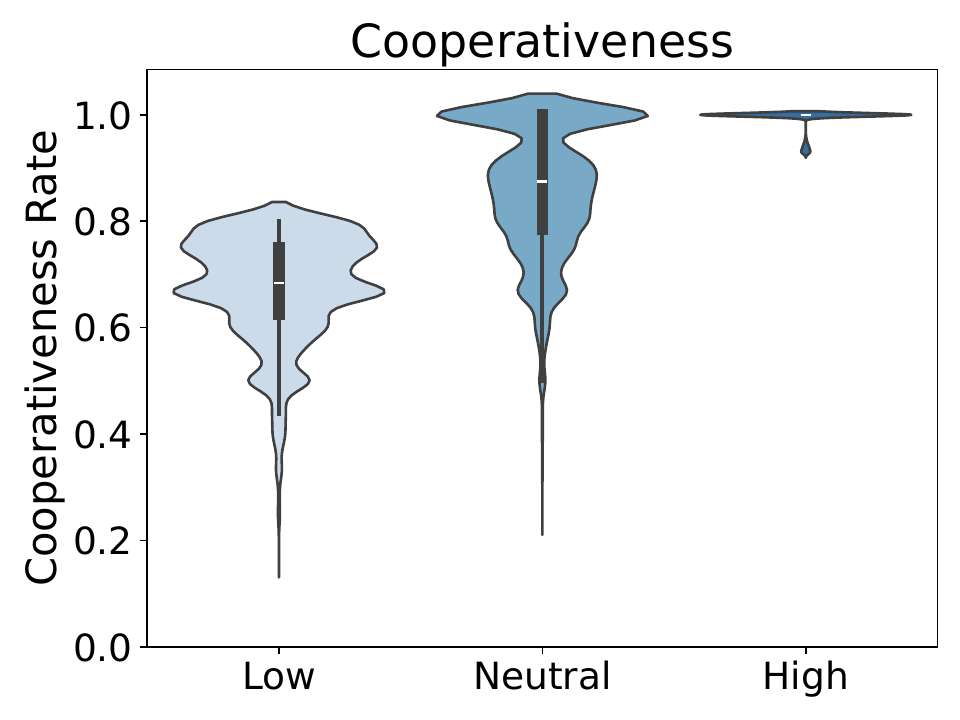}
        \vspace{-6mm}
        \label{fig_cooperativeness_violin}
    \end{subfigure}
    \begin{subfigure}{0.24\textwidth}
        \centering
        \includegraphics[width=\linewidth]{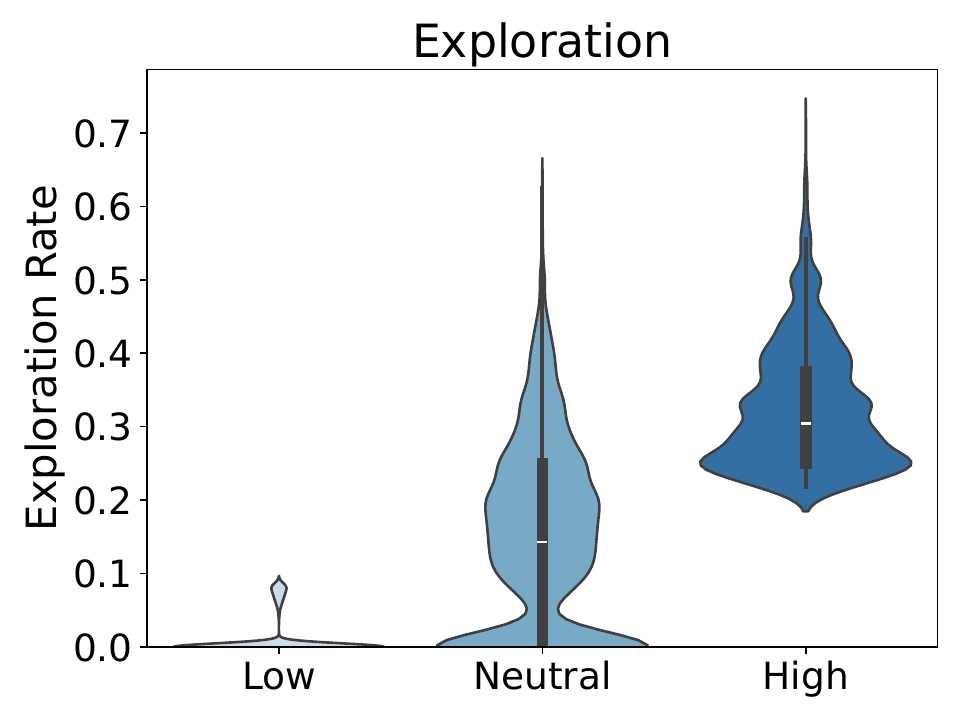}
        \vspace{-6mm}
        \label{fig_explorative_violin_dialogue_level}
    \end{subfigure}
        \begin{subfigure}{0.24\textwidth}
        \centering
        \includegraphics[width=\linewidth]{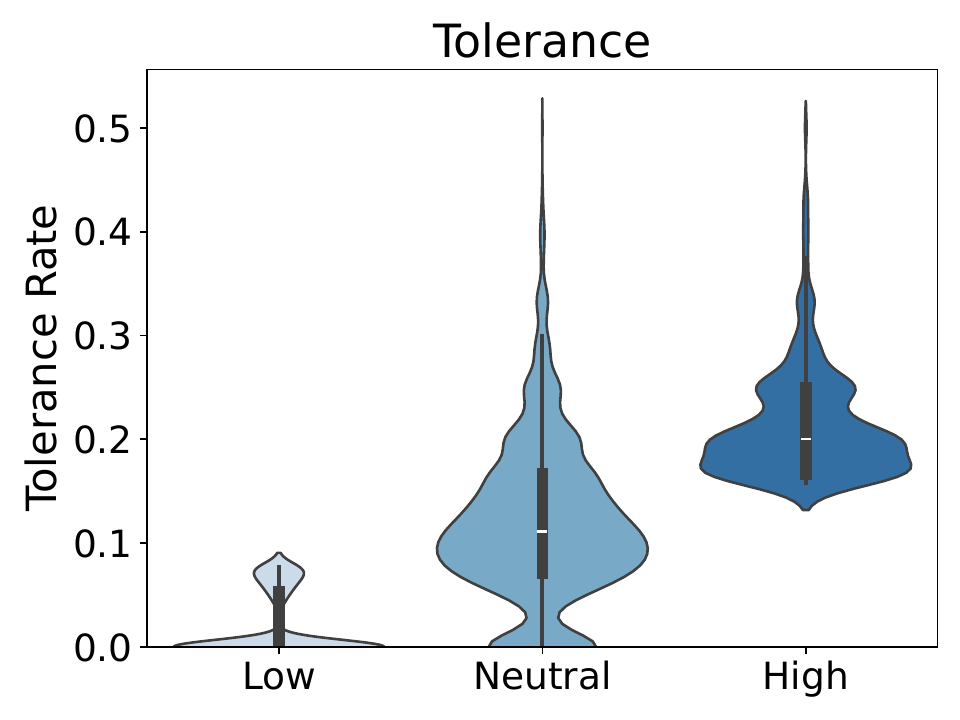}
        \vspace{-6mm}
        \label{fig_tolerance_violin}
    \end{subfigure}
    \\ \vspace{2mm}
    \begin{subfigure}{0.24\textwidth}
        \centering
        \includegraphics[width=\linewidth]{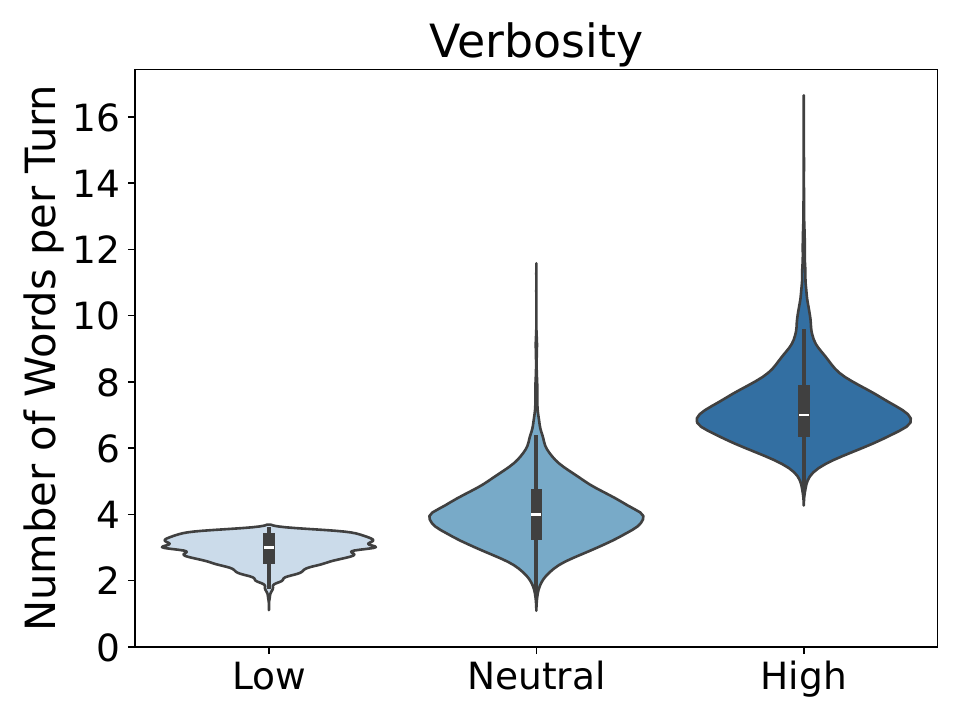}
        \vspace{-6mm}
        \label{fig_verbosity_violin}
    \end{subfigure}
    \begin{subfigure}{0.24\textwidth}
        \centering
        \includegraphics[width=\linewidth]{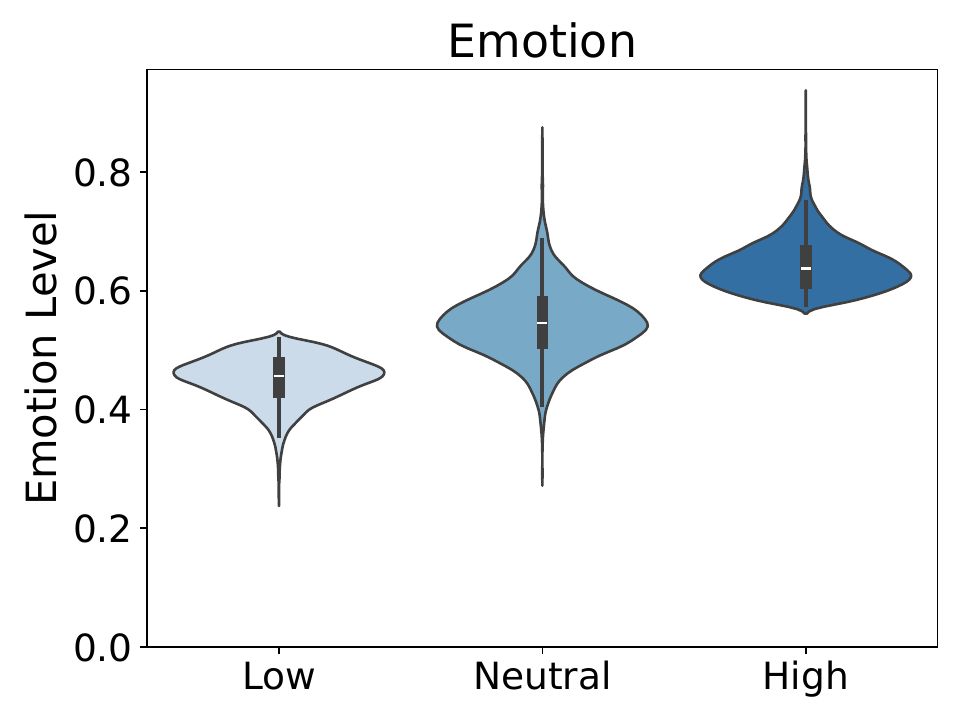}
        \vspace{-6mm}
        \label{fig_emotion_violin}
    \end{subfigure}
    \begin{subfigure}{0.24\textwidth}
        \centering
        \includegraphics[width=\linewidth]{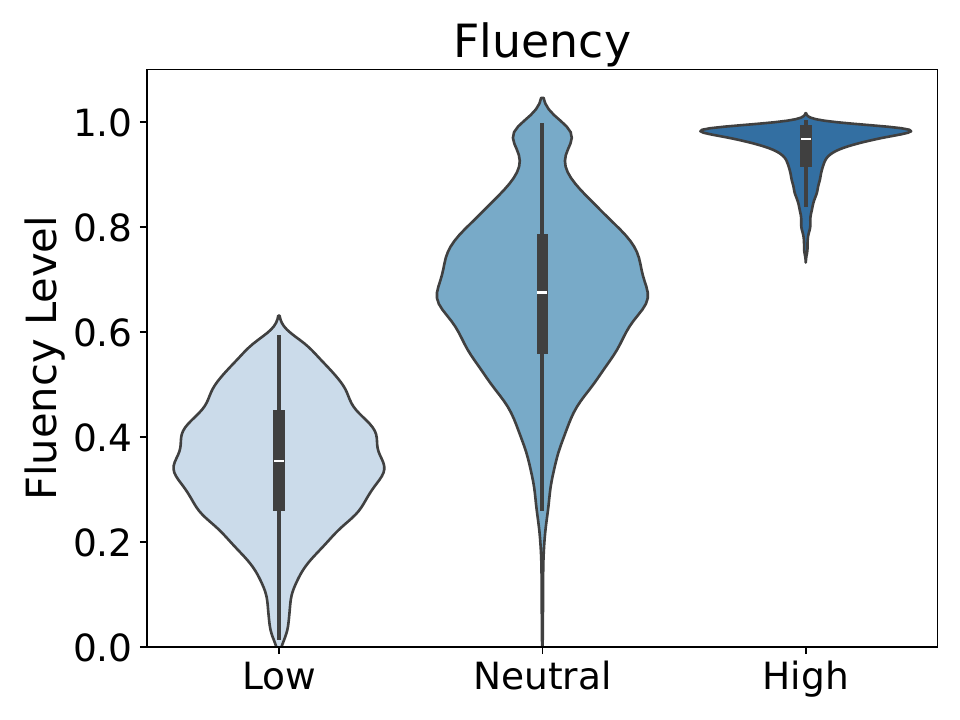}
        \vspace{-6mm}
        \label{fig_fluency_violin}
    \end{subfigure}
    \begin{subfigure}{0.24\textwidth}
        \centering
        \includegraphics[width=\linewidth]{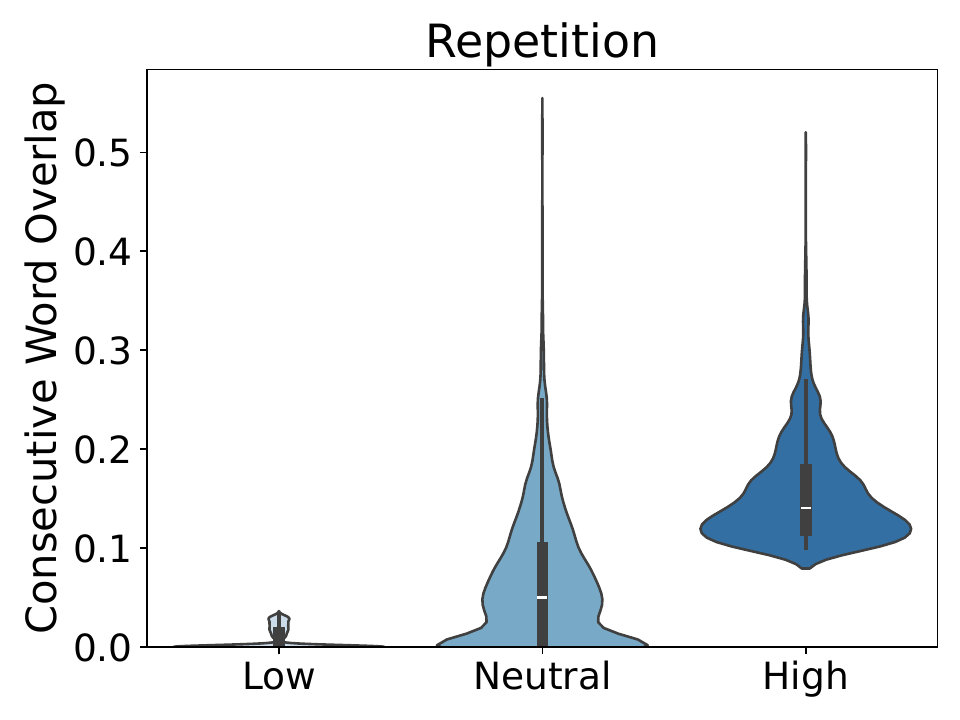}
        \vspace{-5mm}
        \label{fig_repetition_violin}
    \end{subfigure}
    \caption{Training data distribution for dialogue-level (top-row) and utterance-level (bottom-row) traits according to their identifying metric (y-axis) across all trait intensities.}
    \label{fig_training_violins}
\end{figure*}

\subsubsection{Dialog-Level Traits}
\label{sub_sub_dialog_level}
Dialog-Level traits influence the overall progression of a conversation 
by impacting the probability of transitioning between intents.

\paragraph{Engagement.} 
Reflects the user's willingness to engage with the system for a longer period of time as in~\cite{two_ints_simulator}. 
We measure engagement by the number of turns in the dialogue.

\paragraph{Cooperativeness.} 
Users' tendency to follow the system's instructions and reduce unrelated interactions. It is measured by the probability of in-domain intents in the dialogue.
Cooperativeness is a trait also discussed in~\cite{two_ints_simulator, non_cooperative_users_paradigm}, which we include here for task-guiding dialogues.

\paragraph{Exploration.}
This trait strikes the balance between exploitation, which involves moving forward in the task, and exploration, which entails engaging with optional system features~\cite{Learning-to-Ask}.
It is measured by the probability of explorative intents (e.g. QA) that differ from the default navigational requests (e.g. ``next step'').

\paragraph{Tolerance.} 
Represents the user's tolerance for system mistakes, where less tolerant users conclude the interaction sooner~\cite{errors_voice_assistants_book}.
Represented by the tolerance rate, defined as the number of system mistakes tolerated divided by the number of turns.

\subsubsection{Utterance-Level Traits}
\label{sub_sub_utterance_level}
These traits consider a more fine-grained behavior, which given an intent, chooses the style of the utterance according to the user profile.

\paragraph{Verbosity.}
Verboseness of the user's requests. 
Measured by the average number of words in each utterance.

\paragraph{Emotion.}
Refers to the overall tone (negative or positive) expressed towards the system. 
Emotion level is measured in a 0-1 scale using the model from~\cite{sentiment_model}. 

\paragraph{Fluency.} 
Represents the ability to express oneself clearly and coherently without hesitations, disruptions, or ASR errors.
The fluency level is measured in a 0-1 scale using model~\footnote{\url{https://huggingface.co/gchhablani/bert-base-cased-finetuned-cola}}. 

\paragraph{Repetition.}
Consistent use of the same lexicon throughout the dialogue. 
Measured through word overlap between consecutive utterances.

\subsection{Trait Distribution and Intensity}
\label{sub_distribution_per_trait}
Figure~\ref{fig_training_violins} shows the dialogue data distribution per trait and intensity pair.
It shows that certain traits exhibit longer tails and have more noticeable differences across intensities, which is expected to have a direct influence on their modeling challenges.

\section{Experimental Setup}
\label{sub_experimental_setting}

\subsection{Dataset}
\label{sub_sub_dataset}

\paragraph{Tasks and Training Dialogues.}
We focus on the task execution phase~\cite{taskbot_overview_year_1}, where the user has an ongoing task such as ``baking a cake''.
We focus on task assistance in the cooking domain ($\mathcal{M}$) and use 1000 unique recipes to ground the dialogue flow.
Given these, we randomly sample tasks to generate a total of 20k dialogues with an average of 9.6 turns. Finally, we use a 1000/100/100 dialogue split for each trait-intensity pair and ensure generalization by not sharing tasks between splits.

\paragraph{Simulator Inference.}
We evaluate various user simulators by interacting with a live system, \textit{PlanLLM}~\cite{plangpt}, known for its task-guiding capabilities. For each experiment, we generate 100 dialogues per profile, with interactions ending either by the simulator's side (i.e. the user) or upon reaching a maximum turn limit of 20.

\subsection{Models and Baselines.}
As the backbone for our simulators, we used the Transformer-based~\cite{vaswani_attention} Mistral-7B~\cite{mistral} and Vicuna-7B~\cite{vicuna2023}.
The results for Vicuna are in annex and support generalization across LLMs.

We evaluated two types of simulators: the proposed \textit{Specialized Trait Simulators} (\textit{\textbf{STS}}) from Section~\ref{sub_user_profile_definition}, which models each trait independently. 
And as a strong baseline, we consider a \textit{Joint Trait Simulator} (\textit{\textbf{JTS}}), that jointly learns a distribution for all traits in a single model.\footnote{Code available at \url{https://github.com/rafaelhferreira/mtad_cta}}

\paragraph{Implementation Details}
In all experiments, we use LoRA~\cite{lora} adapters and a 4-turn context size. To account for the variability of users, we employ sampling decoding. 
For detailed information about the training procedure, refer to Appendix~\ref{app_hyperparameters}.

\paragraph{Automatic Metrics}
\label{sub_sub_metrics}
Evaluating user simulators is challenging~\cite{simulator_multiwoz, simulators_in_context_learning, kuai_sim_reccomender} due to the absence of a single correct response and multiple dialogue directions, making metrics like BLEU~\cite{Papineni02bleu}, and ROUGE~\cite{lin-2004-rouge} unsuitable~\cite{bleu_not_good}. 
To address this, we analyze the simulator-generated dialogues using each user profiles' identifying characteristics, as depicted in Figure~\ref{fig_training_violins}.
Namely, we measure each distributions' proximity to the training data using Wasserstein's distance~\cite{wasserstein} for discrete trait distributions (i.e. engagement and verbosity), and Kolmogorov-Smirnov's (K-S) distance~\cite{kl_stat} for continuous ones (refer to Appendix~\ref{app_hyperparameters} for detailed metrics' definitions).
In both, lower values indicate a closer approximation to the reference distributions.

\begin{figure*}[!tbp]
    \centering
    \begin{subfigure}{0.24\textwidth}
        \centering
        \includegraphics[width=\linewidth]{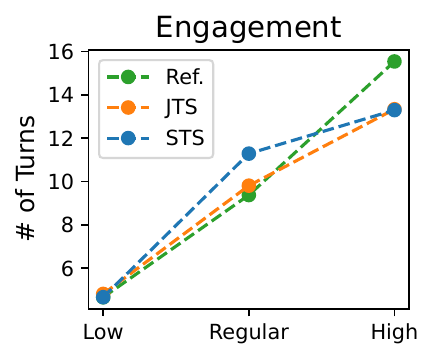}
        \vspace{-4mm}
        \label{fig_patient_2d_mistral}
    \end{subfigure}
        \begin{subfigure}{0.24\textwidth}
        \centering
        \includegraphics[width=\linewidth]{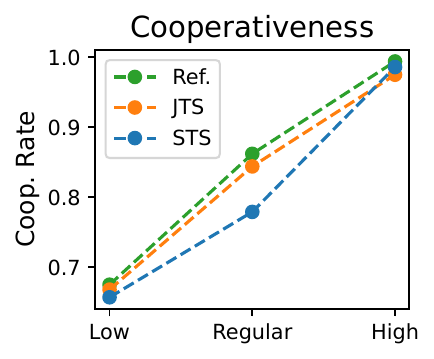}
        \vspace{-4mm}
        \label{fig_cooperativeness_2d_mistral}
    \end{subfigure}
    \begin{subfigure}{0.24\textwidth}
        \centering
        \includegraphics[width=\linewidth]{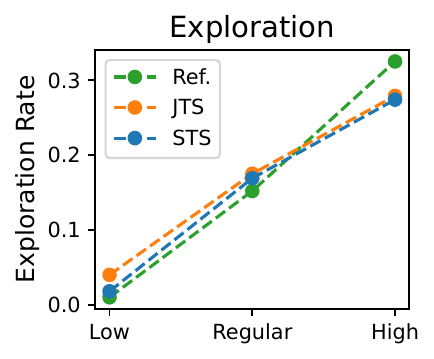}
        \vspace{-4mm}
        \label{fig_exploration_2d_mistral}
    \end{subfigure}
        \begin{subfigure}{0.24\textwidth}
        \centering
        \includegraphics[width=\linewidth]{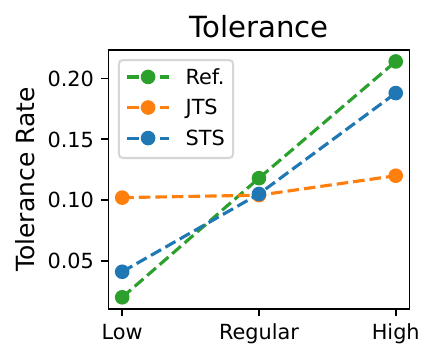}
        \vspace{-4mm}
        \label{fig_tolerance_2d_mistral}
    \end{subfigure}
    \\
    \begin{subfigure}{0.24\textwidth}
        \centering
        \includegraphics[width=\linewidth]{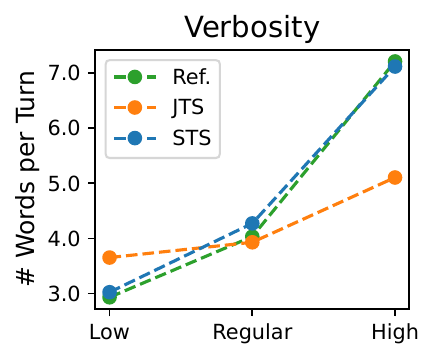}
        \vspace{-4mm}
        \label{fig_verbosity_2d_mistral}
    \end{subfigure}
    \begin{subfigure}{0.24\textwidth}
        \centering
        \includegraphics[width=\linewidth]{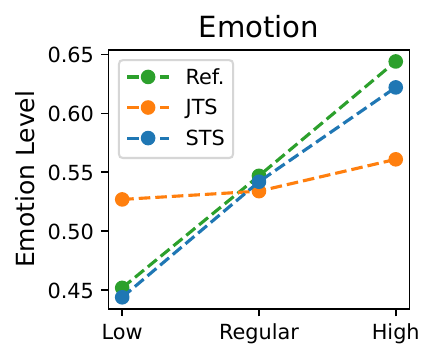}
        \vspace{-4mm}
        \label{fig_emotion_2d_mistral}
    \end{subfigure}
    \begin{subfigure}{0.24\textwidth}
        \centering
        \includegraphics[width=\linewidth]{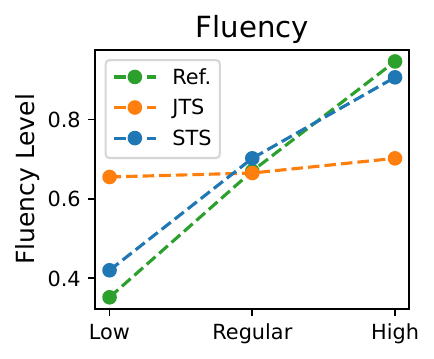}
        \vspace{-4mm}
        \label{fig_fluency_2d_mistral}
    \end{subfigure}
    \begin{subfigure}{0.24\textwidth}
        \centering
        \includegraphics[width=\linewidth]{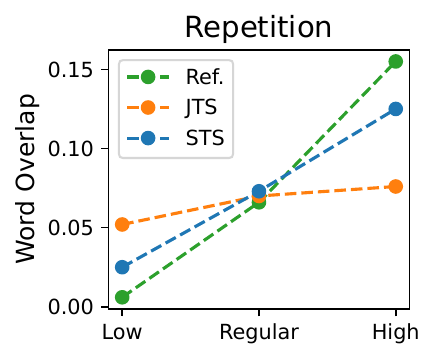}
        \vspace{-4mm}
        \label{fig_repetition_2d_mistral}
    \end{subfigure}
    \caption{Single-trait results for dialogue-level (top-row) and utterance-level (bottom row) traits across all intensities comparing Reference, Joint Trait Simulator (\textit{JTS}), and Specialized Trait Simulators (\textit{STS}).}
    \label{fig_2d_plot_mistral}
\end{figure*}

\section{Results and Discussion}
\label{sec_results}

\subsection{Single-Trait Evaluation}
\label{sub_single_trait_eval}
In this setting, we create user profiles by adjusting each trait from Section~\ref{sub_model_user_traits} to a non-neutral (i.e. \textit{Low} or \textit{High}) intensity level. Additionally, we create a profile called \textit{\textbf{Regular}}, where all traits are set to \textit{neutral}.\footnote{Total of 17 profiles ($8 \ traits \times 2 \ intensities + Regular$)}

\subsubsection{Trait Matching}
\label{sub_sub_single_trait_eval_trend}

In Figure~\ref{fig_2d_plot_mistral}, we compare the performance of \textit{STS} and \textit{JTS} across trait intensities, aiming for values closer to the reference.
As expected, both methods exhibit increasing metrics with trait intensity. 
However, \textit{JTS} shows a greater deviation from the reference at both \textit{Low} and \textit{High} intensities especially noticeable in utterance-level traits.

In Table~\ref{tab_distances_mistral}, we use distance-based metrics to assess the simulators' modeling of the reference distributions.
\textit{STS} generally represents trait behaviors more closely, especially at \textit{Low} and \textit{High} intensities, as each model learns specific profiles independently.
For some traits, \textit{JTS} represents the \textit{Regular} profile (\textit{neutral} intensity for all traits) closer. This is a result of having been exposed to all profiles during training, resulting in an overall smoother distribution that fails to capture extreme trait-intensity subtleties.
%
See Appendix~\ref{app_generated_dialogues_example} for examples of generated dialogues.

\paragraph{Modeling Traits Difficulty.}
Each trait poses unique modeling challenges.
\textit{JTS} struggles with \textit{Low Fluency} since LLMs are trained to generate coherent text. It also struggles with \textit{High Fluency} due to smoothing effects from the \textit{Regular} profile, while STS is able to learn these patterns.

Both \textit{JTS} and \textit{STS} find it difficult to model \textit{Low Tolerance} and \textit{Low Repetition}. \textit{Low Tolerance} is rare, as it depends on system errors, which are beyond the control of the simulator. \textit{Low Repetition} is challenging because it requires generating unique utterances each time, whereas models often copy parts of the dialogue history.

In summary, these results show the difficulty in modeling traits at diverse intensities, highlighting the advantage of specialized trait simulators (\textit{STS}) over a joint trait simulator (\textit{JTS}), particularly for the less common and more extreme user patterns.

\begin{table}[tbp]
\centering
\resizebox{\linewidth}{!}{%
\begin{tabular}{@{}l|cc|cc|cc@{}}
\toprule
 &
  \multicolumn{2}{c|}{\textbf{Low}} &
  \multicolumn{2}{c|}{\textbf{Regular}} &
  \multicolumn{2}{c}{\textbf{High}} \\ \midrule
 &
  \multicolumn{1}{l}{\textbf{JTS}} &
  \multicolumn{1}{l|}{\textbf{STS}} &
  \multicolumn{1}{l}{\textbf{JTS}} &
  \multicolumn{1}{l|}{\textbf{STS}} &
  \multicolumn{1}{l}{\textbf{JTS}} &
  \multicolumn{1}{l}{\textbf{STS}} \\ \midrule
Engagement*       & 0.25 & \textbf{0.10} & \textbf{0.58} & 2.18          & 2.89 & \textbf{2.70} \\
Cooperativeness & 0.24 & \textbf{0.16} & \textbf{0.10} & 0.26          & 0.17 & \textbf{0.10} \\
Exploration     & 0.26 & \textbf{0.12} & 0.09          & \textbf{0.07} & 0.34 & \textbf{0.33} \\
Tolerance       & 0.66 & \textbf{0.28} & 0.20          & \textbf{0.12} & 0.70 & \textbf{0.34} \\ \midrule
Verbosity*      & 0.72 & \textbf{0.17} & \textbf{0.12} & 0.26          & 2.10 & \textbf{0.12} \\
Emotion         & 0.62 & \textbf{0.11} & 0.17          & \textbf{0.06} & 0.66 & \textbf{0.20} \\
Fluency         & 0.74 & \textbf{0.22} & \textbf{0.07} & 0.15          & 0.75 & \textbf{0.27} \\
Repetition      & 0.57 & \textbf{0.35} & 0.09          & \textbf{0.08} & 0.69 & \textbf{0.45} \\ \bottomrule
\end{tabular}%
}
\caption{Results for \textit{JTS} and \textit{STS} in all trait intensities considering a single-trait setting. * Wasserstein for discrete and K-S distance for continuous.}
\label{tab_distances_mistral}
\end{table}

\subsubsection{Generalization to Unseen Domains}
\label{sub_unseen_domains}
To assess how user simulators adapt to new domains $\mathcal{M}$, with out-of-domain tasks,
we randomly sampled 100 DIY tasks from WikiHow\footnote{\url{https://www.wikihow.com/}}, generated 100 more dialogues per profile, and evaluated whether the trend follows the correct pattern\footnote{Since the evaluation is out-of-domain, we omit distance-based metrics due to the lack of a reference distribution.}.

The results in Table~\ref{tab_mistral_wikihow} indicate good generalization, showing an upward tendency in all cases. As before, \textit{STS} achieves more pronounced metrics over intensity ranges, demonstrating the simulator's ability to generalize to novel tasks.

\begin{table}[!t]
\centering
\resizebox{\linewidth}{!}{%
\begin{tabular}{@{}l|ccccc|ccccc@{}}
\toprule
                & \multicolumn{5}{c|}{\textbf{JTS}} & \multicolumn{5}{c}{\textbf{STS}} \\ \midrule
 & \textbf{Low} &  & \textbf{Regular} &  & \textbf{High} & \textbf{Low} &  & \textbf{Regular} &  & \textbf{High} \\ \midrule
Eng        & 4.5   & <  & 8.77  & <  & 12.96  & 5.08  & <  & 9.96  & <  & 12.73  \\
Coop & 0.7   & <  & 0.87  & <  & 0.97   & 0.66  & <  & 0.81  & <  & 0.99   \\
Expl     & 0.04  & <  & 0.18  & <  & 0.24   & 0.02  & <  & 0.13  & <  & 0.28   \\
Tol       & 0.09  & <  & 0.12  & <  & 0.17   & 0.07  & <  & 0.11  & <  & 0.22   \\ \midrule
Verb       & 3.48  & <  & 4.08  & <  & 5.15   & 3.12  & <  & 4.28  & <  & 7.32   \\
Emot         & 0.51  & <  & 0.53  & <  & 0.55   & 0.43  & <  & 0.55  & <  & 0.63   \\
Flu         & 0.64  & <  & 0.70  & <  & 0.72   & 0.46  & <  & 0.72  & <  & 0.90    \\
Rep      & 0.05  & <  & 0.06  & <  & 0.07   & 0.03  & <  & 0.08  & <  & 0.11   \\ \bottomrule
\end{tabular}%
}
\caption{Trend analysis in unseen DIY domain, where each value represents its identifying metric. There is an increasing trend across all traits' intensities.}
\label{tab_mistral_wikihow}
\end{table}

\subsection{Multi-Trait Combination Evaluation}
\label{sub_multi_trait_eval}
We now evaluate trait combinations, using \textit{STS} to model individual traits.
We defined a total of 14 profiles: 8 with 2 traits, 4 with 3 traits, and 2 with 4 traits (full list in Appendix~\ref{app_other_profile_combinations}). For evaluation purposes, we create dialogs for these combinations and use the same evaluation protocol, generating 100 dialogues for each profile and method.

\subsubsection{Combination Baselines and Methods}

\paragraph{Sampling.} 
This method randomly samples an active user profile at each turn. 

\paragraph{\textit{\model}.}
Combines token probabilities from multiple trait simulators as proposed in Section~\ref{sub_sub_mtd}. We use a uniform weight distribution ($\lambda$) across all active trait simulators.

\paragraph{\textit{\model-LA} (Level Aware).}
Extension to \textit{\model} that generates parts of the response using different models, similar to collaborative decoding~\cite{collaborative_decoding}. Specifically, generates the intent using \textit{Dialogue}-level profiles and the utterance using \textit{Utterance}-level profiles.

\vspace{2mm}
Focusing on no-training combination methods, we tested weight-level merging~\cite{dare_merging, ties_merging}. However, outputs did not follow the structure and produced generic utterances. 
Hence, we exclude them from the results.

\begin{table*}[tbhp]
\centering
\small
\begin{tabular}{@{}cl|cccc|cccc@{}}
\toprule
\multicolumn{1}{l}{} &
  \multicolumn{1}{c|}{} &
  \textbf{Enga*} &
  \textbf{Coop} &
  \textbf{Expl} &
  \textbf{Tol} &
  \textbf{Verb*} &
  \textbf{Emot} &
  \textbf{Flu} &
  \multicolumn{1}{l}{\textbf{Rep}} \\ \midrule
\multirow{3}{*}{\rotatebox[origin=c]{90}{\textbf{Avg.}}} &
  Sampling &
  2.61 &
  0.28 &
  0.29 &
  0.18 &
  0.87 &
  0.26 &
  0.33 &
  0.27 \\
 &
  \model &
  2.76 &
  0.27 &
  0.30 &
  \textbf{0.16} &
  0.81 &
  0.26 &
  0.31 &
  0.26 \\
 &
  \model-LA &
  \textbf{2.06} &
  \textbf{0.26} &
  \textbf{0.25} &
  0.18 &
  \textbf{0.48} &
  \textbf{0.24} &
  \textbf{0.30} &
  \textbf{0.20} \\ \midrule
\multirow{3}{*}{\rotatebox[origin=c]{90}{\textbf{2 Traits}}} &
  Sampling &
  1.70 &
  0.25 &
  0.26 &
  0.16 &
  0.88 &
  \textbf{0.20} &
  0.28 &
  0.23 \\
 &
  \model &
  1.76 &
  0.25 &
  0.28 &
  \textbf{0.14} &
  0.80 &
  \textbf{0.20} &
  \textbf{0.26} &
  0.22 \\
 &
  \model-LA &
  \textbf{1.21} &
  0.25 &
  \textbf{0.22} &
  0.16 &
  \textbf{0.58} &
  0.21 &
  \textbf{0.26} &
  \textbf{0.19} \\ \midrule
\multirow{3}{*}{\rotatebox[origin=c]{90}{\textbf{3 Traits}}} &
  Sampling &
  2.27 &
  0.29 &
  0.29 &
  0.21 &
  1.05 &
  0.21 &
  \textbf{0.27} &
  0.35 \\
 &
  \model &
  2.66 &
  0.28 &
  0.31 &
  \textbf{0.18} &
  1.06 &
  0.18 &
  \textbf{0.27} &
  0.34 \\
 &
  \model-LA &
  \textbf{1.91} &
  \textbf{0.26} &
  \textbf{0.26} &
  0.19 &
  \textbf{0.38} &
  \textbf{0.17} &
  0.28 &
  \textbf{0.24} \\ \midrule
\multirow{3}{*}{\rotatebox[origin=c]{90}{\textbf{4 Traits}}} &
  Sampling &
  6.93 &
  0.36 &
  0.40 &
  0.23 &
  0.47 &
  0.65 &
  0.65 &
  0.28 \\
 &
  \model &
  6.98 &
  0.36 &
  0.36 &
  0.23 &
  0.34 &
  0.65 &
  0.62 &
  0.29 \\
 &
  \model-LA &
  \textbf{5.79} &
  \textbf{0.32} &
  \textbf{0.32} &
  \textbf{0.18} &
  \textbf{0.29} &
  \textbf{0.50} &
  \textbf{0.55} &
  \textbf{0.19} \\ \bottomrule
\end{tabular}%
\caption{Results for combining various \textit{STS} models. * Wasserstein for discrete and K-S distance for continuous.}
\label{tab_mistral_profile_combination}
\end{table*}

\subsubsection{Multi-Trait Combination Results}
\label{sub_sub_combination_results}
Table~\ref{tab_mistral_profile_combination} presents the average results for all profiles, and grouped by the number of combined traits. 
\textit{\model}-based methods outperform \textit{Sampling}, as they combine multiple models instead of relying on a single one per turn. \textit{Sampling} also produces inconsistent behavior by mimicking different profiles across turns. 

Using the \textit{\model} framework, it is important to consider the current decoding step, evident by the best performing method \textit{\model-LA}, which activates relevant models depending on the current generation step. 
As the number of traits increases, modeling difficulty also rises due to distribution dilution across profiles.
In these experiments, \textit{\model} weights were fixed and uniformly split across each profile, in Appendix~\ref{sub_sub_multi_trait_weights_analysis}, we show \textit{\model}'s controllability by varying these weights.

In summary, we show models can be combined at decoding time using \textit{\model} without additional training, and that activating models at the appropriate time enhances profile modeling performance.

\subsection{Multi-Level Simulator Evaluation}
\label{sub_human_eval}

To further validate the simulators' behavior, we use the best methods for \textit{Single-Trait} (\textit{STS}) and \textit{Multi-Trait} (\textit{\model-LA}), sampling 10 dialogues from each user profile, resulting in a total of 300 dialogues.

\subsubsection{Multi-Level Evaluation Metrics}
For turn-level evaluation, we considered the following metrics:
\begin{itemize}
    \item \textit{Degeneration} - A rule-based binary metric that flags degeneration if: 1) the output does not have a valid user intent, or 2) special tokens (e.g., speaker labels, \textit{SOS}, \textit{EOS}) appear within the middle of a generated utterance.
\item \textit{Uniqueness} - A binary metric that checks if the generated utterance is novel, i.e., it does not appear in the training data, indicating the model’s ability to generate original responses.
\item \textit{System Response Quality} - A rating from 0 to 2, assessing the quality of the system's response in relation to the ongoing dialogue.
\end{itemize}

At the dialogue level, we measured \textit{Trait Modeling Accuracy} by comparing each profile's generated dialogue across its defining traits (i.e. $l_i \neq neutral$), with a dialogue from the test set with the same task but an opposite trait intensity, as well as comparing with a \textit{Regular} dialogue.

Based on previous research~\cite{llm-as-judge, dpo_paper} showing that LLMs align closely with human judgments, we used GPT-4o\footnote{gpt-4o-2024-05-13} to annotate 2776 examples for \textit{System Response Quality} and 1040 for \textit{Trait Modeling Accuracy}. To ensure reliability, we conducted a study with 5 human annotators, achieving an average of 83\% agreement and 0.67 Fleiss Kappa score with GPT-4o, confirming its suitability for this task (full details and prompts in Appendix~\ref{app_human_eval_details}).

\subsection{Turn Level Results.}
Analyzing turn-level metrics in Table~\ref{tab_single_trait_and_multi_trait_turn_level}, we observe that degenarations are rare ($\leq 4\%$), indicating the methods' ability to generate and combine profiles effectively. 

Regarding uniqueness, both methods have similar values, with over 50\% of the utterances not appearing in the training set, demonstrating good generalization to other requests and tasks.

The system's response quality is generally good with the model achieving an average score over 1.4.
A closer analysis by trait revealed that the system achieves higher response quality scores ($\geq$1.57) handling higher \textit{fluency}, \textit{exploration}, and \textit{emotion}, while scoring lower ($\leq1.34$) for low \textit{engagement}, \textit{tolerance} and high \textit{verbosity}. 
These results allow to understand system limitation and in turn create more robust systems, as it has been shown that simulator variability can improve system performance~\cite{cannot_stand_everyone_tod, user_diversified_policy}.

\begin{table}[t]
\centering
\small
\begin{tabular}{@{}lcc@{}}
\toprule
                  & \textbf{Single-Trait} & \textbf{Multi-Trait} \\ \midrule
Degeneration      & 0.03                  & 0.04                 \\
Unique Utterances & 0.48                  & 0.46                 \\
System RQ (0-2)   & 1.46                  & 1.43                      \\
\bottomrule
\end{tabular}%
\caption{Turn-level metrics for Degeneration, Unique utterances and System Response Quality.}
\label{tab_single_trait_and_multi_trait_turn_level}
\end{table}

\subsection{Trait Modeling Accuracy Results.}
Table~\ref{tab_single_trait_and_multi_trait} presents the results of trait modeling accuracy. On average, we observe a small drop in performance when moving from single to multi-trait settings. As expected, model accuracy is higher when comparing opposite intensity traits, due to the more pronounced differences between them compared to a \textit{Regular} dialogue.
Performance significantly declines in the multi-trait setting with \textit{Regular} dialogues, as the model must integrate multiple profiles, resulting in a smoothing effect that makes it harder to distinguish traits compared to the single-trait setting.

Examining individual traits, modeling low \textit{verbosity} and high \textit{emotion} is particularly challenging since their \textit{Regular} values are similar to these traits. Conversely, \textit{engagement} is easier to model because it relies on the number of turns, and low \textit{emotion} is detectable due to utterances indicating low motivation. Full results by trait are provided in Appendix~\ref{app_model_eval_per_trait}.

In summary, the results show good performance in both single-trait and multi-trait settings, allowing for diverse simulators that can be effectively combined at decoding time.

\begin{table}[!t]
\centering
\small
\begin{tabular}{@{}lcc@{}}
\toprule
                  & \textbf{Single-Trait} & \textbf{Multi-Trait} \\ \midrule
Average Comparison          & 0.73                  & 0.70                      \\ \midrule
Comparison w/ Opposite          & 0.76                  & 0.78                     \\
Comparison w/ Regular          & 0.70                  & 0.62                     \\
\bottomrule
\end{tabular}%
\caption{Trait Modeling Accuracy Average, and considering only comparisons between a profile and its opposite intensity profile or a \textit{Regular} profile.}
\label{tab_single_trait_and_multi_trait}
\end{table}

\section{Conclusions}
\label{sec_conclusions}
This paper addresses the challenge of simulating diverse user traits and effectively combining them in a conversational setting.
Using real-world data to identify user traits, we developed a framework to generate profile-aware conversations, being one of the first to deliver user simulators for the Conversational Task Assistance (CTA) setting.

Our results demonstrate the need for specialized simulators for each trait (\textit{STS}), which produce a more accurate trait adherence when compared to jointly learning all traits with a single model. The latter converges to highly smoothed trait distributions that tend to overlook subtle trait characteristics. With \textit{STS}, trait-specific characteristics are better preserved, including in unseen domains with different tasks, highlighting its generalizability.

To flexibly combine diverse user traits and profiles without extra fine-tuning, we proposed Multi-Trait Adaptive Decoding (\textit{\model}). The results show that \textit{\model} effectively combines multiple user traits. This shows that our framework provides an adaptable and controllable approach to create user simulators that can accommodate a wide range of profiles and tasks.
%


\section*{Limitations}
\label{sec_limitations}
Our analysis focuses on a particular set of conversational traits. Given the versatility and broadness of this setting, additional traits could be considered to have a more comprehensive user simulator.
Additionally, given the combinatorial nature of possible trait combinations, despite our approach supporting it, we did not exhaustively explore all combinations of the various traits. 
%
Furthermore, our simulator focuses on the task execution level, leaving room for future work in the retrieval and grounding processes of Conversational Task Assistants~\cite{mango_mango}. 

To conclude, we believe our work presents a step forward in creating adaptable user simulators, however, subsequent studies should broaden the scope to allow for a more comprehensive understanding of user simulation dynamics.

\section*{Ethical Considerations}
\label{sec_ethical_considerations}
All human interactions considered in this study were obtained voluntarily, with users having the ability to withdraw at any point. Prior to each interaction with the system, users were informed that all information would be saved and shared with the authors. Subsequently, users were given the option to proceed with the interaction, indicating their understanding of the terms and conditions.
Moreover, the responses collected were anonymous and devoid of user demographics.
Additionally, we conducted a thorough review of the data to ensure that no personal information was included.

Regarding the annotators, all individuals volunteered and provided consent to participate in the experiment. They were fully informed about the study and its implications.

\section*{Acknowledgements}
This work has been partially funded by the Amazon Science - TaskBot Prize Challenge 2021 and 2022, by the NOVA LINCS project Ref. UIDP/04516/2020 and by FCT Ref. UI/BD/151261/2021.

\bibliography{custom}

\appendix

\clearpage

\section{Real-world Data Analysis}
\label{app_real_data_stats}
We leverage the interaction data from~\citet{plangpt} composed of real-world interactions collected in the context of the Alexa Prize TaskBot Challenge~\cite{taskbot_overview_year_1}, where a user interacts with an Alexa device by voice to complete a manual task in the cooking or DIY domain. 
We focus on conversations with at least three turns that reached the start of a task, and we identified traits at two levels: \textit{Dialogue}, controlling the flow using intents, and \textit{Utterance}, representing the way users express these intents. Table~\ref{tab_trait_and_metric} provides a summarized overview of the identified traits and their identifying characteristic.

\paragraph{Dialogue Level}

\begin{figure*}[htbp]
    \centering
    \begin{subfigure}{0.22\textwidth}
        \centering
        \includegraphics[width=\linewidth]{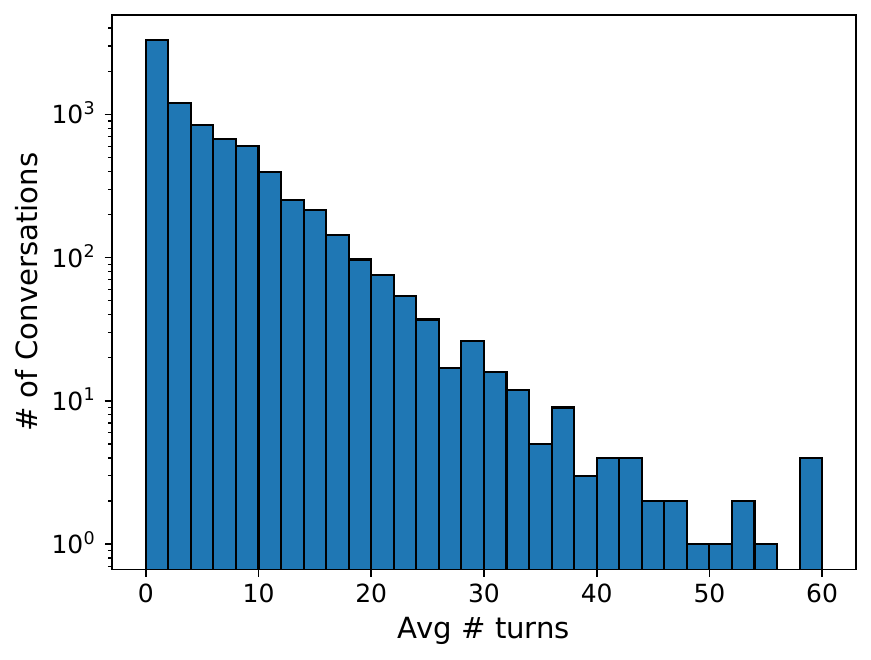}
        \caption{Engagement}
        \label{fig_patient_real_data}
    \end{subfigure}
        \begin{subfigure}{0.22\textwidth}
        \centering
        \includegraphics[width=\linewidth]{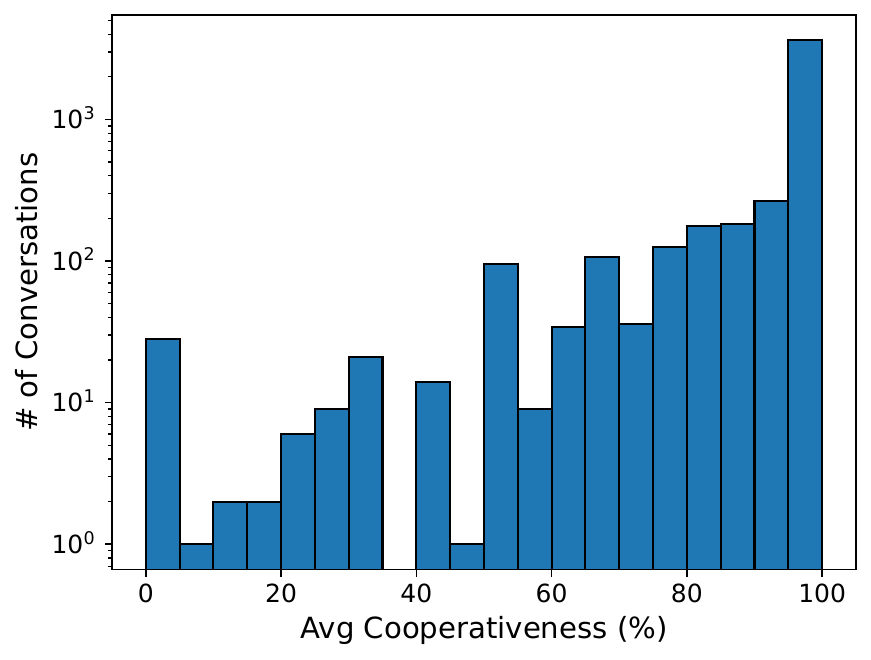}
        \caption{Cooperativeness}
        \label{fig_cooperativeness_real_data}
    \end{subfigure}
    \begin{subfigure}{0.22\textwidth}
        \centering
        \includegraphics[width=\linewidth]{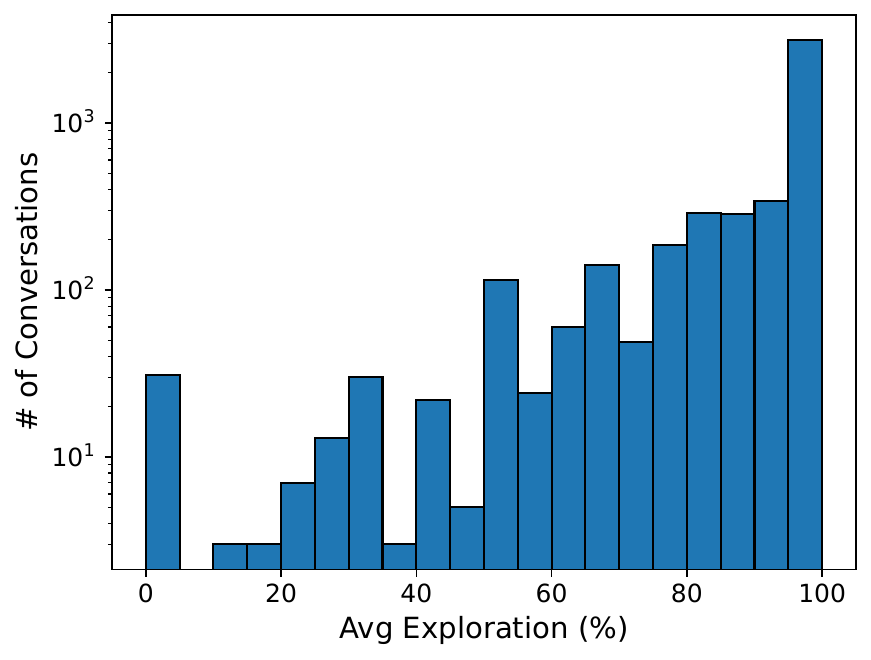}
        \caption{Exploration}
        \label{fig_explorative_real_data}
    \end{subfigure}
    \begin{subfigure}{0.22\textwidth}
        \centering
        \includegraphics[width=\linewidth]{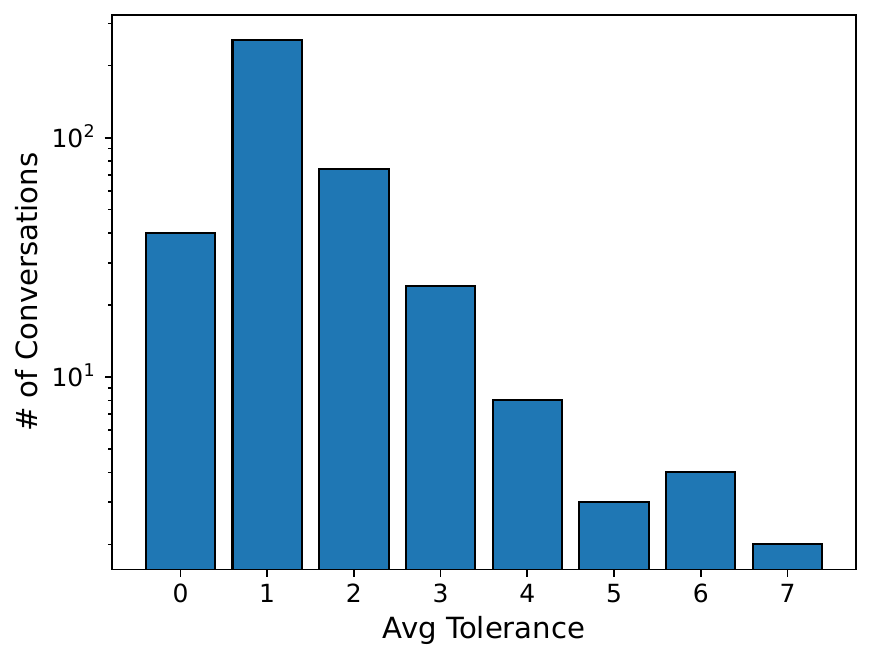}
        \caption{Tolerance}
        \label{fig_tolerance_real_data}
    \end{subfigure}
    \caption{Real user conversations statistics for dialogue level traits in the Alexa TaskBot Dataset.}
    \label{fig_real_data_dialogue_level}
\end{figure*}

Figure~\ref{fig_real_data_dialogue_level} shows the statistics for dialogue-level traits.
\textit{Engagement} predictably decreases as the number of turns increases, with most conversations being short as users explore the system or find tasks misaligned with their goals.
Most users exhibit high \textit{cooperativeness}, close to 100\%, since they primarily follow the task by stating ``next''. Without ``next'' intents, cooperativeness drops from 94\% to 60\%. 
Similarly, users show a strong tendency toward task \textit{exploration} going deeper into the task, however, by excluding ``next'' intents, explorativeness decreases from 92\% to 57\%.
Finally, \textit{tolerance} is analyzed in sessions with at least one system error or an unrecognized request, showing users typically endure at least one error, with a maximum tolerance of seven errors.

\begin{table}[!t]
\centering
\small
\resizebox{\linewidth}{!}{%
\begin{tabular}{@{}lll@{}}
\toprule
                                 & \textbf{Trait}           & \textbf{Identifying Characteristic}   \\ \midrule
\multirow{4}{*}{\rotatebox[origin=c]{90}{Dialogue}}  & Engagement        & Number of Turns                 \\
                                 & Cooperativeness & Cooperativeness Rate          \\
                                 & Tolerance       & Tolerance Rate                \\
                                 & Exploration     & Exploration Rate              \\ \midrule
\multirow{4}{*}{\rotatebox[origin=c]{90}{Utterance}} & Verbosity       & Number of Words per Turn        \\
                                 & Emotion         & Emotion Level            \\
                                 & Fluency         & Fluency Level            \\
                                 & Repetition      & Consecutive Word Overlap \\ \cmidrule(l){1-3} 
\end{tabular}%
}
\caption{Dialogue and Utterance Level traits and corresponding characteristic.}
\label{tab_trait_and_metric}
\end{table}

\paragraph{Utterance Level}

\begin{figure*}[htbp]
    \centering
    \begin{subfigure}{0.22\textwidth}
        \centering
        \includegraphics[width=\linewidth]{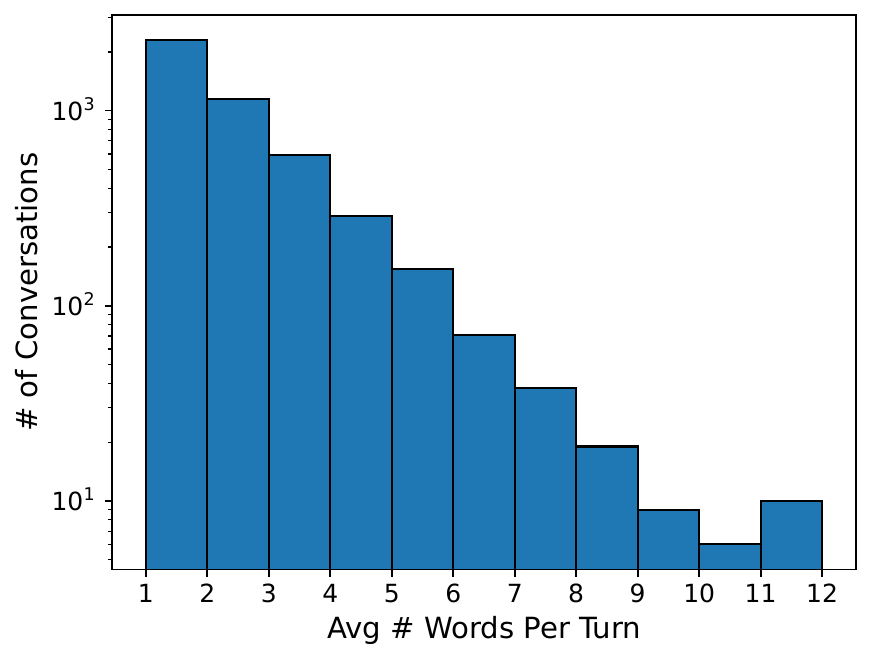}
        \caption{Verbosity}
        \label{fig_verbosity_real_data}
    \end{subfigure}
    \begin{subfigure}{0.22\textwidth}
        \centering
        \includegraphics[width=\linewidth]{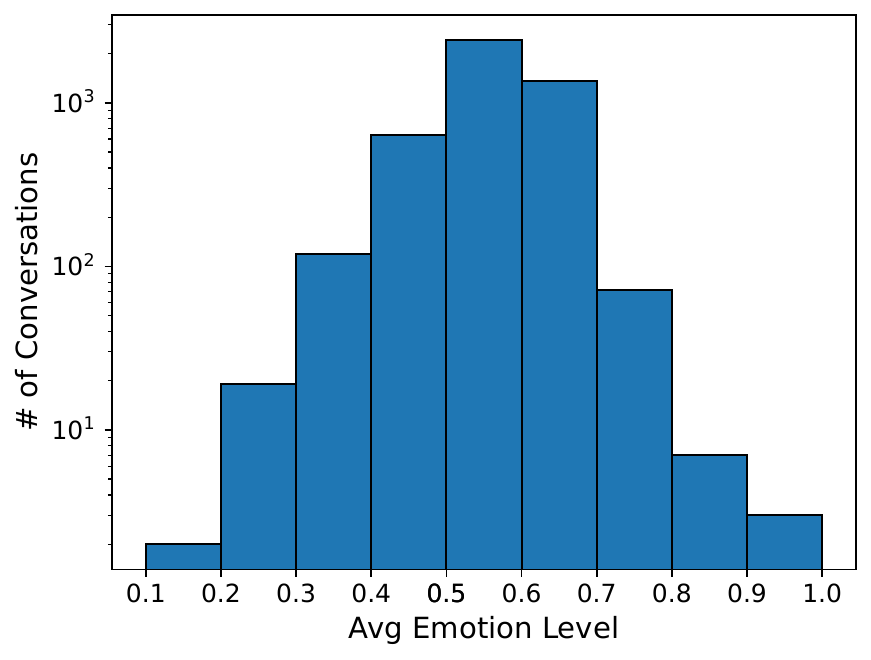}
        \caption{Emotion}
        \label{fig_emotion_real_data}
    \end{subfigure}
    \begin{subfigure}{0.22\textwidth}
        \centering
        \includegraphics[width=\linewidth]{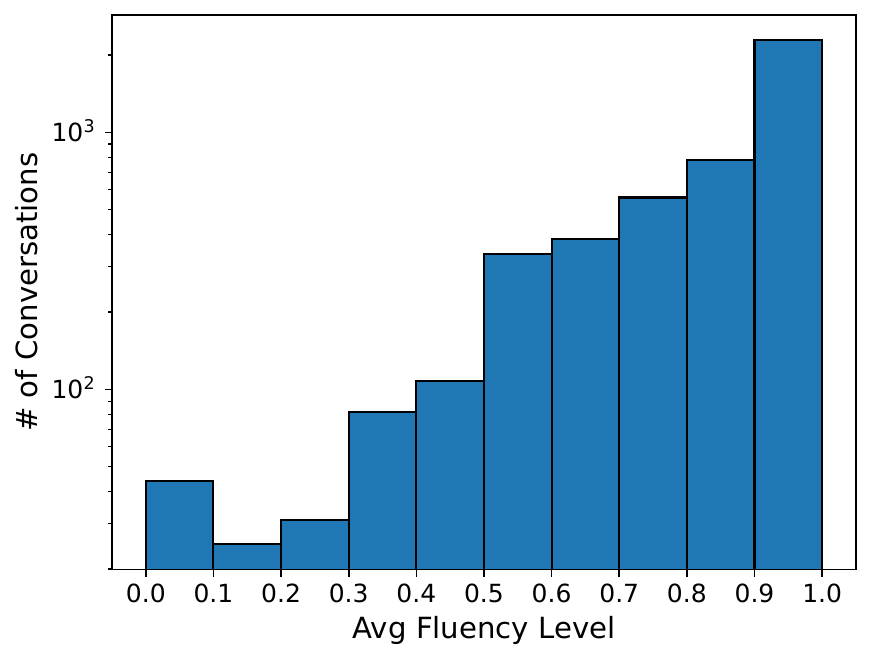}
        \caption{Fluency}
        \label{fig_fluency_real_data}
    \end{subfigure}
    \begin{subfigure}{0.22\textwidth}
        \centering
        \includegraphics[width=\linewidth]{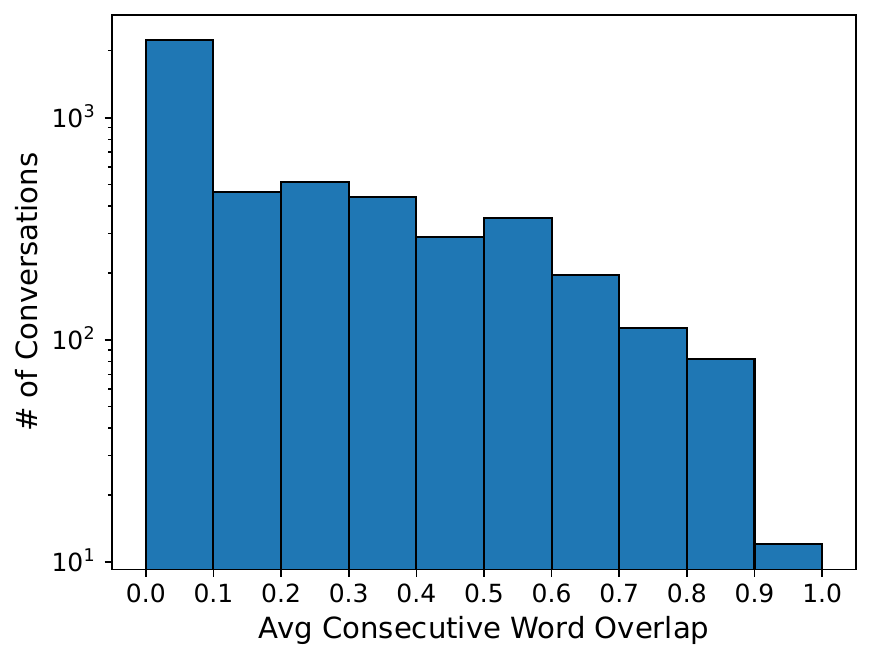}
        \caption{Repetition}
        \label{fig_repetition_real_data}
    \end{subfigure}
    \caption{Real user conversations statistics for dialogue level traits in the Alexa TaskBot Dataset.}
    \label{fig_real_data_utterance_level}
\end{figure*}

Figure~\ref{fig_real_data_utterance_level} illustrates utterance-level traits. 
Regarding \textit{verbosity}, users predominantly employ short, straightforward utterances focused on continuing the task.
This observation shows the disparity between natural-sounding utterances, as depicted in~\cite{wizard_of_tasks}, and real-world user interactions, emphasizing the importance of utilizing authentic user data over crowdsourced alternatives.
Most utterances exhibit mid-range \textit{emotion} levels, aligning with the nature of the task. Instances of lower and higher emotional sentiment generally indicate user frustration or satisfaction. 
\textit{Fluency} is generally high, influenced by the prevalence of short utterances. 
Lastly, \textit{repetition} analysis shows minimal consecutive word overlap, indicating diverse vocabulary usage, though occasional redundancy is observed during task navigation.

\section{Dataset Creation Details}
\label{app_dataset_creation}

\subsection{Conversational flow}
Following the approach of~\citet{plangpt}, we utilized the same dataset of 3.6k real-world interactions from the Alexa Prize TaskBot Challenge~\cite{taskbot_overview_year_1}. Using this data, we generated a directed graph based on intent transition probabilities, detailed in Table~\ref{tab_intents_list}.

\begin{table*}[tbhp]
\centering
\resizebox{\textwidth}{!}{%
\begin{tabular}{@{}llllccc@{}}
\toprule
             & \textbf{Intent}       & \textbf{Description}                                                & \textbf{Example}                   & \textbf{Stop} & \textbf{Expl.} & \textbf{Coop.} \\ \midrule
\multirow{6}{*}{\rotatebox{90}{\textbf{\ \ Navigational}}} 
& Start     & User asks to start the task.                          & please start                 &      &      &          \\
& NextStep     & User asks to go to the next step.                          & next step                 &      & \ding{51}     & \ding{51}            \\
             & PreviousStep & User asks to go to the previous step.                      & previous step             &      &      & \ding{51}             \\
             & Resume       & User asks the system to continue the current step.         & resume                    &      &      & \ding{51}             \\
             & Repeat       & User asks the system to repeat the previous response.      & repeat that               &      &      & \ding{51}             \\
             & Stop         & User ends the interaction.                                 & finish this task                      & \ding{51}    &       & \ding{51}             \\ \midrule
\multirow{4}{*}{\rotatebox{90}{\textbf{\ Open QA}}}  & Question     & User asks a recipe-related question.                       & how much salt do i need  &      & \ding{51}     & \ding{51}              \\
 &
  Definition &
  User asks for an explanation of a concept. &
  what is a spatula &
   &
  \ding{51} &
  \ding{51}  \\
 &
  Replacement &
  User asks for replacements of a tool/ingredient. &
  i do not have sugar &
   &
  \ding{51} &
  \ding{51}  \\
             & GetFunFact   & User asks for a fun fact.    & tell me a fun fact        &      & \ding{51}     & \ding{51}       \\ \midrule
\multirow{4}{*}{\rotatebox{90}{\textbf{\ Other}}} &
  NewTask &
  User asks for a new unrelated task. &
  how to change a tire &
   &
   &
   \\
  & ChitChat &
  The basic conversation norms, e.g., thanks, chitchat. &
  how are you today &
   &
   &
 \\
 &
  Sensitive &
  Dangerous or inappropriate requests (answer is denied). &
  how to make a nuke &
   &
  &
 \\
             & Fallback     & Any request where user intention is not clear or OOD. & find restaurants near me &      &     &            \\ \bottomrule
\end{tabular}%
}
\caption{Intents list, description, and example utterances. We categorize intents into the groups: ``Stop'', ``Exploration'', and ``Cooperative'', which influence their occurrence probability when creating profile-aware dialogues.}
\label{tab_intents_list}
\end{table*}

\subsection{System Responses}
For the system responses, we utilize the framework provided by~\citet{plangpt}, which is composed of both template-based and contextual LLM-generated responses. 

\subsection{User Utterances}
Regarding user utterances, these are obtained from the dataset in~\cite{plangpt}, where each of the top-100 utterances was manually classified into an intent, which we further verified for personally identifiable information.

This process resulted in a dataset grounded in reality, as users interacted voluntarily and naturally with the system. Consequently, the dataset includes various instances of noisy utterances (e.g. ASR errors and arbitrary requests), enabling the simulator to capture these real-world patterns.

\subsubsection{Profile-Aware Dataset Creation Pipeline}
Our aim is to generate dialogues given a specific user profile. This user profile, as defined in Section~\ref{sub_user_profile_definition}, is composed of pairs of traits (Table~\ref{tab_trait_and_metric}) and intensities (\textit{low}, \textit{neutral} and \textit{high}).
Algorithm~\ref{alg_dialogue_creation} details our method for incorporating this information during dataset generation: 1) apply the user profile to the intent transitions, 2) sample an intent, 3) change weights of utterances according to the user profile, and 4) sample an utterance.

\begin{figure*}[tbhp]
    \centering
    \includegraphics[width=0.98\linewidth]{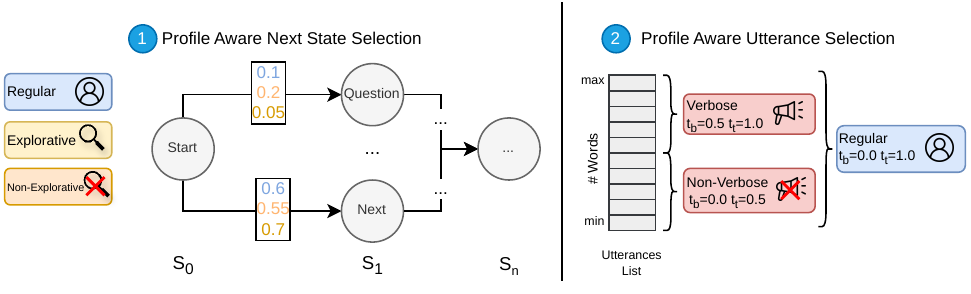}
    \caption{User profile application stages. First step is to choose the next state (dialogue-level) which has different probabilities for state transitions depending on the profile. In a second stage after choosing a state the goal is too choose an utterance (utterance-level) according to the user profile.}
    \label{fig_user_profile_application}
\end{figure*}

\begin{algorithm*}[tbhp]
\centering
\caption{Profile-Aware Dialogue Creation Algorithm}
\label{alg_dialogue_creation}
\begin{algorithmic}
\Require
    \State $\mathcal{T}$: List of tasks
    \State $u$: User profile
    \State $n$: Turn limit
    \State $\mathcal{S}$: State transition dictionary
    \State $\mathcal{U}_s$: Utterance list for each state $s \in \mathcal{S}$
    \State $apply\_dialogue\_level\_traits$: Function that applies user profile to state transition dict
    \State $apply\_utterance\_level\_traits$: Function that applies user profile to utterances list
    \State $get\_system\_utterance$: Function that returns a system response given state, utterance and task

\Ensure
\State $\mathcal{D}$ $\leftarrow []$: Current Dialog
\State $s_t$ $\leftarrow Start$: Current state
\State Randomly select a task: $t \sim \mathcal{T}$
\State Apply user profile to state dictionary: $\mathcal{S}'_t = apply\_dialogue\_level\_traits(u, \mathcal{S})$

\While{$n > 0$ and $s_t$ != $Stop$}
    \State Update and Sample state: $s_t \sim \mathcal{S}'_t$
    \State Apply user profile to utterance list: $\mathcal{U}'_{s_t} = apply\_utterance\_level\_traits(u, \mathcal{U}_{s_t})$ 
    \State Sample utterance: $u_t \sim \mathcal{U}'_{s_t}$
    \State Sample system response and new state $r_t, s_t = get\_system\_utterance(s_t, u_t, t)$
    \State Update state: $s_t \sim \mathcal{S}'_t(s_t)$
    \State Add turn to dialog: $\mathcal{D} \leftarrow \mathcal{D} + (s_t, u_t, r_t)$
    \State Decrement turn limit: $n \leftarrow n - 1$
\EndWhile
\State \Return $\mathcal{D}$
\end{algorithmic}
\end{algorithm*}

\subsubsection{Dialogue Level Traits Integration}
Given a profile, the transitions between intents are modified (Figure~\ref{fig_user_profile_application} left).

Generically, the probability $P(i|i_c)$ of transitioning to an intent $i$ given the current intent $i_c$  is computed as:
\begin{equation}
  P(i|i_c) = 
  \begin{cases}
    P(i|i_c) \cdot f, & \text{if } i \ \in \ Ints \\
    P(i|i_c), & \text{otherwise}
  \end{cases}
  \label{eq_dialogue_trait}
\end{equation}
where $f$ is a factor that increases or decreases the probability of that intent occurring which is only applied if the intent $i$ is part of a particular set of considered intents $Ints$.

\paragraph{Engagement.} We multiply the probability of the set of stop intents ($Ints =$ \textit{Stop}) by a factor $f$, which in turn decreases or increases the probability of stopping early. 

\paragraph{Cooperativeness.} 
This is modeled in the same way as Engagement (Eq.~\ref{eq_dialogue_trait}) but using a different factor $f$ and the set of uncooperative intents ($Ints \neq$ \textit{Coop.}).

\paragraph{Tolerance.} 
We follow a similar approach by increasing the probability of intents in the (\textit{Stop}) set when a system error occurs (e.g. incorrect answer, refusal to answer, fallback response):
\begin{equation}
    P(i|i_c) = P(i|i_c) \cdot f^{(\# \ errors)}, \ i \in Stop
\end{equation}
where, $\# \ errors$ denotes the current number of errors during the dialogue. Tolerance is designed to impose a more significant penalty for system mistakes.
During dialogue generation, we intentionally select incorrect system responses a percentage of the time to simulate errors.

\paragraph{Exploration.} 
We calculate the sum of the probabilities for the top-k intents and for the explorative intents (\textit{Expl.}) not in the \textit{top-k}:

\small
\begin{equation}
    P_{\text{top-k}} = \sum_{i \in \text{top-k}} P_{i} \ \ \ \ ; \ \ \ \ P_{E} = \sum_{i \in (Expl - \text{top-k})} P_{i}.
\end{equation}
\normalsize

We then remove part of the probability from the \textit{top-k} intents, and share it with the others: 

\small
\begin{equation}
  P(i|i_c) = 
  \begin{cases}
    P(i|i_c) - (P_{\text{top-k}} * f) * (P(i|i_c) / P_{\text{top-k}}), & \\ \quad \quad \text{if } i \in \text{top-k} \\
    P(i|i_c) + (P_{\text{top-k}} * f) * (P(i|i_c) / P_{Expl}), & \\ \quad \quad \text{if } i \ \in Expl - \text{top-k} \\
    P(i|i_c), & \\ \quad \quad \text{otherwise}
  \end{cases}
\end{equation}
\normalsize

We present the hyperparameters for dialogue-level traits in Table~\ref{tab_dialogue_level_parameters}.

\subsubsection{Utterance-Level Traits Integration}
We change the weights of choosing an utterance according to a user profile.

\paragraph{Verbosity, Sentiment, and Fluency.}
For each trait, we first attribute a value to each utterance, considering a specific function $F$.
For verbosity, the function ($F$) used is the number of words, while for sentiment and fluency, we use the Transformer-based models from~\footnote{\url{https://huggingface.co/cardiffnlp/twitter-roberta-base-sentiment-latest}} and~\footnote{\url{https://huggingface.co/gchhablani/bert-base-cased-finetuned-cola}}, respectively, which outputs we transform into a 0-1 scale.
After having these values for all the utterances, we apply a bottom-p ($t_b$) and top-p ($t_t$) threshold to select from this pool  
(Figure~\ref{fig_user_profile_application} right).

\paragraph{Repetition}
For repetition, we use word-overlap. This involves two probabilities: the likelihood of selecting a previously mentioned utterance via an exact match ($r_e$) and the likelihood of selecting utterances with some level of overlap ($r_o$).  If neither option applies (e.g. dialogue start), we sample from all available utterances without constraints.

We present the hyperparameters for utterance-level traits in Table~\ref{tab_turn_level_parameters}.

\begin{table}[tbhp]
\centering
\resizebox{\linewidth}{!}{%
\begin{tabular}{@{}lcc@{}}
\toprule
\textbf{Profile}     & \textbf{Intents Considered ($I$)} & \textbf{Factor ($f$)}                 \\ \midrule
Engagement=Low            & $Stop$                    & 2.0                                   \\
Engagement=High          & $Stop$                    & 0.5                                   \\
Cooperativeness=Low        & $I \notin Coop.$           & 2..0                                   \\
Cooperativeness=High    & $I \notin Coop.$           & 0.5                                   \\
Exploration=Low        &  $Expl.$             & 0.2 (other to top-1 intents)     \\
Exploration=High    &  $Expl.$             & 0.2 (top-1 to other intents) \\
Tolerance=Low           & $Stop$                    & 10.0                                   \\
Tolerance=High       & $Stop$                    & 1.0                                  \\
All Other Profiles & -                           & -                                     \\ \bottomrule
\end{tabular}%
}
\caption{Dialogue-level parameters used to create the dialogues for a particular type. ``All Other Profiles'' indicates the parameters used to create the profiles not attached to that particular metric.}
\label{tab_dialogue_level_parameters}
\end{table}

\begin{table}[tbhp]
\centering
\resizebox{\linewidth}{!}{%
\begin{tabular}{@{}lccc@{}}
\toprule
\textbf{Profile} & \textbf{Metric} & \textbf{Bottom Threshold ($t_b$)} & \textbf{Top Threshold ($t_t$)}    \\ \midrule
Verbosity=Low   & Avg \# Words & 0.0  & 0.5  \\
Verbosity=High   & Avg \# Words     & 0.5  & 1.0  \\
Emotion=Low & Avg Emotion  & 0.0  & 0.5  \\
Emotion=High & Avg Emotion  & 0.5  & 1.0  \\
Fluency=Low     & Avg Fluency  & 0.0  & 0.5  \\
Fluency=High & Avg Fluency  & 0.5  & 1.0  \\ 
All Other Profiles & (All Above)  & 0.0  & 1.0  \\ \midrule
\textbf{Profile} & -               & \textbf{Exact Match ($r_e$)}   & \textbf{Overlap Match ($r_o$)} \\ \midrule
Repetition=Low           & -            & 0.0  & 0.0  \\
Repetition=High       & -            & 1.0  & 1.0  \\ 
All Other Profiles              & -            & 0.15 & 0.15 \\
\bottomrule
\end{tabular}%
}
\caption{Utterance-level parameters used to create the dialogues for a particular type. ``All Other Profiles'' indicates the parameters used to create the profiles not attached to that particular metric.}
\label{tab_turn_level_parameters}
\end{table}

\subsection{Dialogue Filtering}
To guarantee the dialogues have diversity and follow the intended behavior, we first generate 10k dialogues for the \textit{Regular} profile to collect average statistics.
After this, we apply half a standard deviation, up for \textit{high}, and down for \textit{low} intensities to obtain dialogues within the required characteristics. 

For a summarized example showing the distinction between profiles, see Table~\ref{tab_dialogue_level_dialogue} and \ref{tab_utterance_level_dialogue} for dialogue and utterance-level profiles, respectively.

\begin{table}[thbp]
\centering
\small
\texttt{
\begin{tabular}{@{}|ll|@{}}
\toprule
\multicolumn{1}{|c}{\textbf{Intent}} & \multicolumn{1}{c|}{\textbf{Utterance}} \\ \midrule
\multicolumn{2}{|c|}{Profile: {\textbf{Exploration High}}}                                            \\ \midrule
Start                                & i want to start the task                \\
Next                                 & next                                    \\
Question                             & how much sugar do i need                \\
Next                                 & next step                               \\
Replacement                          & i do not have orange juice              \\ \midrule
\multicolumn{2}{|c|}{Profile: {\textbf{Exploration Low}}}                                           \\ \midrule
Start                                & start task                              \\
Next                                 & next                                    \\
Next                                 & next                                    \\
Next                                 & next step                               \\
stop                                 & stop now                                \\ \midrule
\multicolumn{2}{|c|}{Profile: {\textbf{Cooperativeness High}}}                                           \\ \midrule
Start                                & start task                              \\
Next                                 & next step                               \\
Next                                 & i'm done                                \\
Curiosity                            & tell me a fun fact                      \\
Next                                 & next                                    \\ \midrule
\multicolumn{2}{|c|}{Profile: {\textbf{Cooperativeness Low}}}                                           \\ \midrule
Start                                & start                                   \\
Sensitive                            & you are stupid                          \\
Next                                 & next                                    \\
Fallback                             & what can you see                        \\
Stop                                 & turn off                                \\ \bottomrule
\end{tabular}%
}
\caption{Dialogue-level example transitions for various simulators. System answers are omitted.}
\label{tab_dialogue_level_dialogue}
\end{table}

\begin{table}[tbhp]
\centering
\small
\texttt{
\resizebox{\linewidth}{!}{%
\begin{tabular}{@{}|p{0.99\linewidth}|@{}}
\toprule
\textbf{Recipe:} Easy Taco Pie                                \\ \midrule
\multicolumn{1}{|c|}{\textbf{Intent:} Start}                  \\ \midrule
{\textbf{Verbosity High:}} i want to start the task             \\
{\textbf{Verbosity Low:}} start                                 \\
{\textbf{Fluency High:}} start the task                         \\
{\textbf{Fluency Low:}} start cats                              \\
{\textbf{System:}} Let's start! Step 1: Preheat the oven to 350ºF (180 degrees C). \\ \midrule
\multicolumn{1}{|c|}{\textbf{Intent:} Next}                   \\ \midrule
{\textbf{Verbosity High:}} i am done what's next                  \\
{\textbf{Verbosity Low:}} next                                   \\
{\textbf{Fluency High:}} next step please                        \\
{\textbf{Fluency Low:}} next p                              \\
{\textbf{System:}} Step 2: Heat up your ground Lean Ground Beef in a frying pan on stove ...                     \\ \midrule
\multicolumn{1}{|c|}{\textbf{Intent:} Stop}                   \\ \midrule
{\textbf{Verbosity High:}} can you stop                           \\
{\textbf{Verbosity Low:}} stop                                   \\
{\textbf{Fluency High:}} let's stop                              \\
{\textbf{Fluency Low:}} hey off                             \\
{\textbf{System:}} Happy to help! See you again soon!              \\ \bottomrule
\end{tabular}%
}
}
\caption{Utterance-level example excerpts for various simulators.}
\label{tab_utterance_level_dialogue}
\end{table}

\subsection{Dataset Statistics}

Table~\ref{tab_filtered_stats} provides a comprehensive overview of the generated data. Profiles effectively capture distinct traits, as seen from the range of values achieved. After filtering, traits like \textit{Fluency} and \textit{Verbosity} retain most dialogues, while most \textit{Tolerance} dialogues are filtered given these rely on errors triggered by the system.
Trait interactions show behaviors like \textit{Low Exploration} and \textit{Low Tolerance} leading to shorter dialogues. In contrast, \textit{High Patience} and \textit{High Tolerance} profiles explore more compared to \textit{Low Patience}. 
Regarding utterance-level traits, we see minimal effects on each other.

Overall, these findings demonstrate how different user profiles impact various aspects of dialogues, including intents, utterances, and dialogue flow.

\begin{table*}[tb]
\centering
\resizebox{\textwidth}{!}{%
\begin{tabular}{@{}lcccccccccc@{}}
\toprule
 \textbf{Trait} & 
  \multicolumn{1}{c}{\textbf{Intensity}} &
  \textbf{\# Turns} &
  \textbf{Coop. Rate} &
  \textbf{Expl. Rate} &
  \textbf{Tolerance Rate} &
  \textbf{\# Words} &
  \textbf{Emotion} &
  \textbf{Fluency} &
  \textbf{Word Overlap} &
  \textbf{\# Dial} \\ \midrule
Regular &
  - &
  9.38 $\pm$ 4.4 &
  0.86 $\pm$ 0.12 &
  0.15 $\pm$ 0.13 &
  0.12 $\pm$ 0.08 &
  4.03 $\pm$ 0.96 &
  0.55 $\pm$ 0.06 &
  0.67 $\pm$ 0.16 &
  0.07 $\pm$ 0.07 &
  9000 \\ \midrule
\multirow{2}{*}{Engagement} &
  Low &
  {\ul 4.65 $\pm$ 1.44} &
  0.86 $\pm$ 0.16 &
  0.08 $\pm$ 0.12 &
  0.14 $\pm$ 0.1 &
  3.83 $\pm$ 1.07 &
  0.55 $\pm$ 0.08 &
  0.68 $\pm$ 0.2 &
  0.04 $\pm$ 0.06 &
  5731 \\
 &
  High &
  \textbf{15.55 $\pm$ 3.51} &
  0.87 $\pm$ 0.09 &
  0.2 $\pm$ 0.11 &
  0.09 $\pm$ 0.05 &
  4.18 $\pm$ 0.77 &
  0.55 $\pm$ 0.04 &
  0.65 $\pm$ 0.11 &
  0.08 $\pm$ 0.05 &
  3911 \\ \midrule
\multirow{2}{*}{Cooperativeness} &
  Low &
  9.71 $\pm$ 5.0 &
  {\ul 0.68 $\pm$ 0.1} &
  0.12 $\pm$ 0.11 &
  0.14 $\pm$ 0.09 &
  4.1 $\pm$ 1.01 &
  0.55 $\pm$ 0.06 &
  0.7 $\pm$ 0.15 &
  0.05 $\pm$ 0.05 &
  5510 \\
 &
  High &
  8.9 $\pm$ 4.12 &
  \textbf{0.99 $\pm$ 0.02} &
  0.17 $\pm$ 0.14 &
  0.09 $\pm$ 0.07 &
  3.96 $\pm$ 0.9 &
  0.55 $\pm$ 0.06 &
  0.64 $\pm$ 0.16 &
  0.09 $\pm$ 0.08 &
  5222 \\ \midrule
\multirow{2}{*}{Exploration} &
  Low &
  7.6 $\pm$ 3.95 &
  0.84 $\pm$ 0.14 &
  {\ul 0.01 $\pm$ 0.02} &
  0.12 $\pm$ 0.08 &
  3.56 $\pm$ 0.89 &
  0.56 $\pm$ 0.06 &
  0.67 $\pm$ 0.18 &
  0.08 $\pm$ 0.08 &
  3830 \\
 &
  High &
  10.58 $\pm$ 4.5 &
  0.88 $\pm$ 0.1 &
  \textbf{0.32 $\pm$ 0.08} &
  0.12 $\pm$ 0.08 &
  4.6 $\pm$ 0.88 &
  0.53 $\pm$ 0.05 &
  0.67 $\pm$ 0.15 &
  0.06 $\pm$ 0.06 &
  3543 \\ \midrule
\multirow{2}{*}{Tolerance} &
  Low &
  9.9 $\pm$ 5.09 &
  0.86 $\pm$ 0.12 &
  0.18 $\pm$ 0.12 &
  {\ul 0.02 $\pm$ 0.03} &
  4.12 $\pm$ 0.94 &
  0.53 $\pm$ 0.05 &
  0.69 $\pm$ 0.14 &
  0.06 $\pm$ 0.06 &
  1800 \\
 &
  High &
  10.96 $\pm$ 5.83 &
  0.78 $\pm$ 0.13 &
  0.2 $\pm$ 0.13 &
  \textbf{0.21 $\pm$ 0.05} &
  4.14 $\pm$ 0.94 &
  0.54 $\pm$ 0.05 &
  0.68 $\pm$ 0.15 &
  0.06 $\pm$ 0.06 &
  1122 \\ \midrule
\multirow{2}{*}{Verbosity} &
  Low &
  9.15 $\pm$ 4.26 &
  0.87 $\pm$ 0.12 &
  0.12 $\pm$ 0.11 &
  0.12 $\pm$ 0.08 &
  {\ul 2.93 $\pm$ 0.4} &
  0.56 $\pm$ 0.05 &
  0.66 $\pm$ 0.16 &
  0.06 $\pm$ 0.07 &
  6214 \\
 &
  High &
  9.31 $\pm$ 4.32 &
  0.86 $\pm$ 0.12 &
  0.15 $\pm$ 0.13 &
  0.12 $\pm$ 0.08 &
  \textbf{7.21 $\pm$ 1.13} &
  0.53 $\pm$ 0.07 &
  0.62 $\pm$ 0.16 &
  0.1 $\pm$ 0.07 &
  8992 \\ \midrule
\multirow{2}{*}{Emotion} &
  Low &
  9.53 $\pm$ 4.32 &
  0.86 $\pm$ 0.12 &
  0.16 $\pm$ 0.13 &
  0.11 $\pm$ 0.08 &
  4.26 $\pm$ 0.93 &
  {\ul 0.45 $\pm$ 0.04} &
  0.68 $\pm$ 0.15 &
  0.07 $\pm$ 0.06 &
  8206 \\
 &
  High &
  9.55 $\pm$ 4.41 &
  0.86 $\pm$ 0.12 &
  0.15 $\pm$ 0.12 &
  0.12 $\pm$ 0.07 &
  3.82 $\pm$ 0.91 &
  \textbf{0.64 $\pm$ 0.04} &
  0.66 $\pm$ 0.15 &
  0.07 $\pm$ 0.07 &
  8115 \\ \midrule
\multirow{2}{*}{Fluency} &
  Low &
  9.55 $\pm$ 4.38 &
  0.87 $\pm$ 0.12 &
  0.15 $\pm$ 0.12 &
  0.12 $\pm$ 0.07 &
  3.92 $\pm$ 0.98 &
  0.56 $\pm$ 0.06 &
  {\ul 0.35 $\pm$ 0.12} &
  0.07 $\pm$ 0.07 &
  8318 \\
 &
  High &
  9.26 $\pm$ 4.36 &
  0.86 $\pm$ 0.12 &
  0.15 $\pm$ 0.13 &
  0.12 $\pm$ 0.08 &
  4.14 $\pm$ 0.89 &
  0.54 $\pm$ 0.06 &
  \textbf{0.95 $\pm$ 0.05} &
  0.07 $\pm$ 0.07 &
  8918 \\ \midrule
\multirow{2}{*}{Repetition} &
  Low &
  8.37 $\pm$ 4.0 &
  0.85 $\pm$ 0.13 &
  0.14 $\pm$ 0.12 &
  0.12 $\pm$ 0.08 &
  3.83 $\pm$ 0.87 &
  0.55 $\pm$ 0.06 &
  0.68 $\pm$ 0.15 &
  {\ul 0.01 $\pm$ 0.01} &
  5224 \\
 &
  High &
  10.02 $\pm$ 4.34 &
  0.9 $\pm$ 0.11 &
  0.13 $\pm$ 0.13 &
  0.11 $\pm$ 0.07 &
  3.94 $\pm$ 0.94 &
  0.55 $\pm$ 0.05 &
  0.65 $\pm$ 0.15 &
  \textbf{0.16 $\pm$ 0.05} &
  3103 \\ \bottomrule
\end{tabular}%
}
\caption{Training data statistics. For each metric, maximum values are bold, and minimum values are underlined.}
\label{tab_filtered_stats}
\end{table*}


\section{Model Input Format}
\label{app_input_format}
\begin{figure}[!t]
    \centering
    \includegraphics[width=1.0\linewidth]{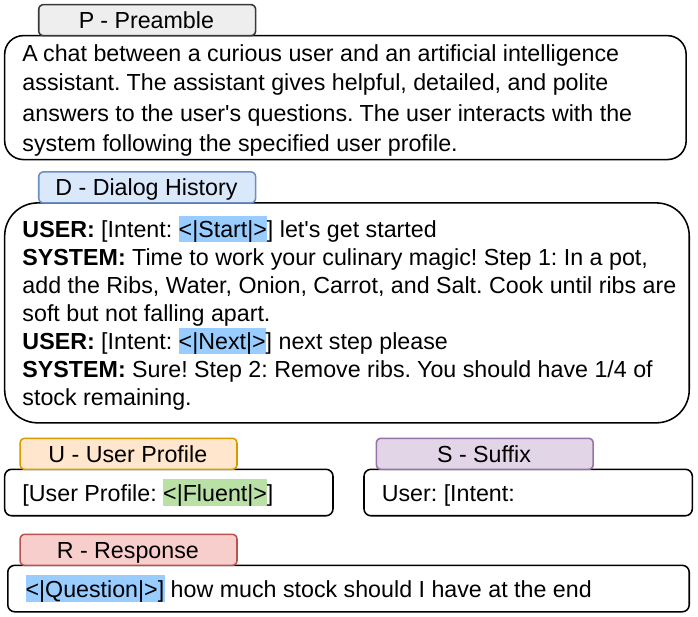}
    \caption{Model input and its various components. We also highlight the \colorbox{intent}{intents} and the \colorbox{fluent}{profile}. The simulator should generate the response (R).}
    \label{fig_prompt}
\end{figure}

The prompt used is shown in Figure~\ref{fig_prompt}, which consists of four components:

\begin{itemize}
    \item \textbf{Preamble} - Establishes the context and defines the task.
    \item \textbf{Dialogue History} - Includes the previous $n$ turns of dialogue, comprising both user intent and utterance, and system responses. 
    \item \textbf{User Profile}: Specifies the profile for the model to follow, identified by a unique token sequence.
    \item \textbf{Suffix} - A suffix that prompts the model to generate a response, since the loss should not be calculated over these tokens.
\end{itemize}

The model is trained in a dual-task, generating responses with both an intent and a user utterance, enhancing interpretability and controllability by allowing analysis of intent probabilities and injection of specific intents.

\section{Implementation Details}
\label{app_hyperparameters}

\paragraph{Handling Profile Imbalance.} 
Profiles exhibit varying average numbers of turns, causing imbalance. To address this, we use a stratified approach, ensuring an equal number of samples for training each user profile. 

\paragraph{Handling Intent Class Imbalance.} 
There exists a significant imbalance in intent distribution, with the dominant intent, \textit{NextStep}, representing 37\% of interactions.
To avoid overemphasizes on this intent, we undersample \textit{NextStep} by 50\% in the training dataset, resulting in a more balanced intent distribution.

\paragraph{Libraries.} We used Pytorch~\cite{pytorch} and the Huggingface library~\cite{huggingface}. 

\paragraph{Training Details and Hyperparameters.} 
For all runs, we trained a Low-Rank adapter (LoRa)~\cite{lora} on the embedding, query, key, value, and language modeling head layers, using bf16 precision and 4 turns of dialogue history. Other hyperparameters are in Table~\ref{tab_hyperparameters}. 
For training, we use the cross-entropy loss and select the best model based on the lowest loss in validation.

\paragraph{Hardware and Training Time.} 
For training, we use an A100-40GB GPU.
Regarding training times, the Joint Trait Simulator (\textit{JTS}) model takes around 20 hours, while each Specialized Trait Simulator (\textit{STS}) takes takes around 2 hours.

\paragraph{Simulator Inference.}
For inference, we use an A100-40GB GPU to load the user simulator model and the system model~\cite{plangpt} with which the simulator interacts in bf16 precision.
\textit{\model} uses LoRA adapters and leveraging approaches such as those in~\cite{mole_experts, mixture_of_loras}, it enables the simultaneous use of multiple active LoRA adapters with minimal impact on speed and memory usage compared to their original LLM backbones.
Inference speed results show that Mistral-7B~\cite{mistral} and Vicuna-7B~\cite{vicuna2023} generate an average of 250 and 480 tokens per second, respectively. We attribute these differences to variations in architecture and library optimizations, despite both models having the same number of parameters.

\begin{table}[!t]
\centering
\small
\begin{tabular}{@{}lc@{}}
\toprule
\multicolumn{1}{l}{\textbf{Parameter}} & \textbf{Value} \\ \midrule
Batch Size                             & 4              \\
Grad Acc.                              & 4              \\
Epochs                                 & 15              \\
Max Input Size                             & 1024           \\
Model DType                            & bf16           \\ \midrule
Learning Rate                          & $1 \cdot 10^{-5}$      \\
Scheduler                              & Linear         \\
Optimizer                              & AdamW          \\ \midrule
LoRa Rank                              & 16             \\
LoRa $\alpha$                          & 32             \\
LoRa Dropout                           & 0.1            \\ \bottomrule
\end{tabular}%
\caption{Training Hyperparameters.}
\label{tab_hyperparameters}
\end{table}

\paragraph{Automatic Metrics}
\label{app_metrics}

Evaluation of user simulators is challenging due to the absence of a single correct response and the potential for dialogues variations.
To address this, we analyze the identifying characteristics of each user profile (Table~\ref{tab_trait_and_metric}). 

In particular, we consider the distance metrics: Wasserstein Distance~\cite{wasserstein} for discrete metrics (i.e., engagement - number of turns, verbosity - number of words), and Kolmogorov-Smirnov (K-S) distance~\cite{kl_stat} for continuous metrics.
Wasserstein Distance measures the minimum ``work'' needed to transform one distribution into another. In the case of K-S distance, it measures the maximum distance between cumulative distribution functions (CDFs), indicating how much model-generated distributions deviate from the actual distributions.
In both metrics, lower values indicate better alignment with the reference distribution.

%

\paragraph{Decoding Strategy}
Selecting a greedy approach 
results in repetitive dialogues with minimal variation.
To address this, we adopted a sampling-based decoding strategy, leveraging the probabilistic nature of user simulation. 


\section{Distribution Analysis}
\begin{figure*}[!t]
    \centering
    \begin{subfigure}{0.24\textwidth}
        \centering
        \includegraphics[width=\linewidth]{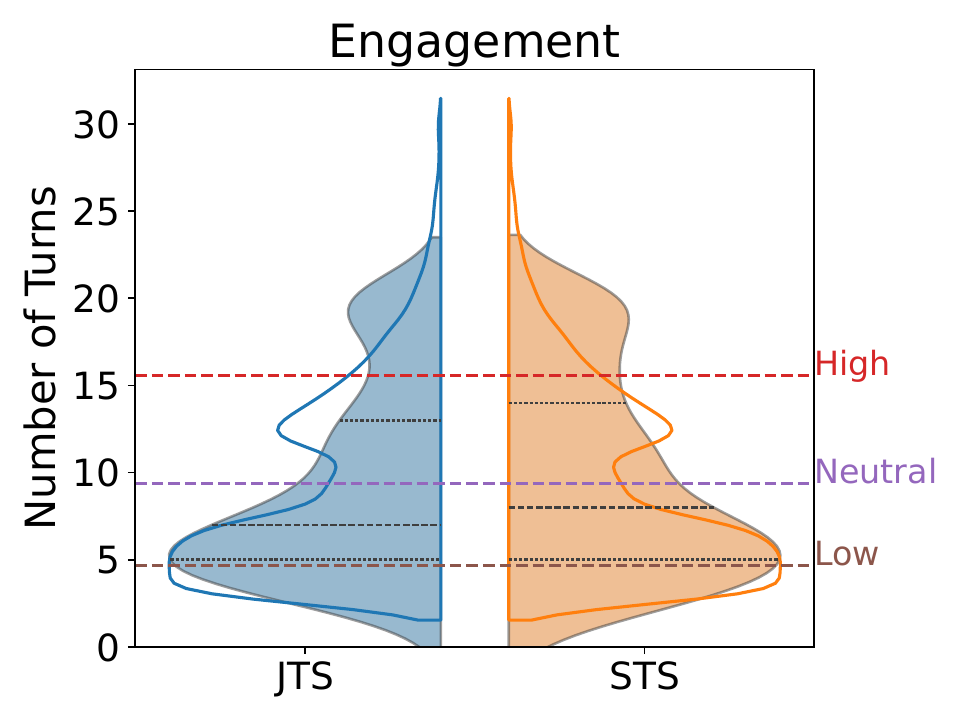}
                \vspace{-4mm}
        \label{fig_patient_violin_split}
    \end{subfigure}
        \begin{subfigure}{0.24\textwidth}
        \centering
        \includegraphics[width=\linewidth]{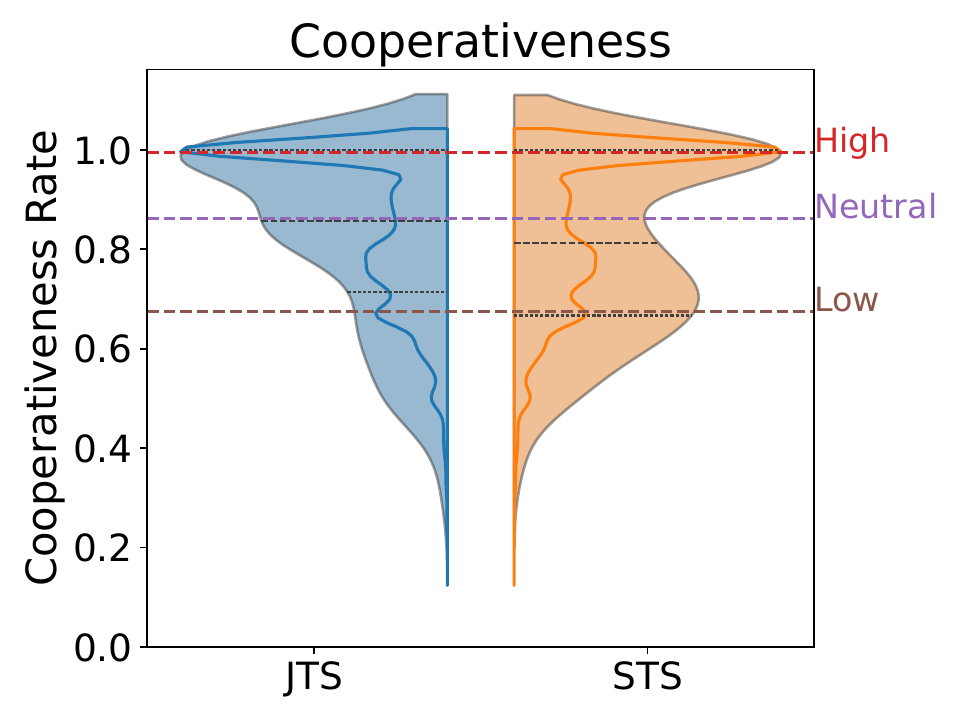}
                \vspace{-4mm}
        \label{fig_cooperativeness_violin_split}
    \end{subfigure}
    \begin{subfigure}{0.24\textwidth}
        \centering
        \includegraphics[width=\linewidth]{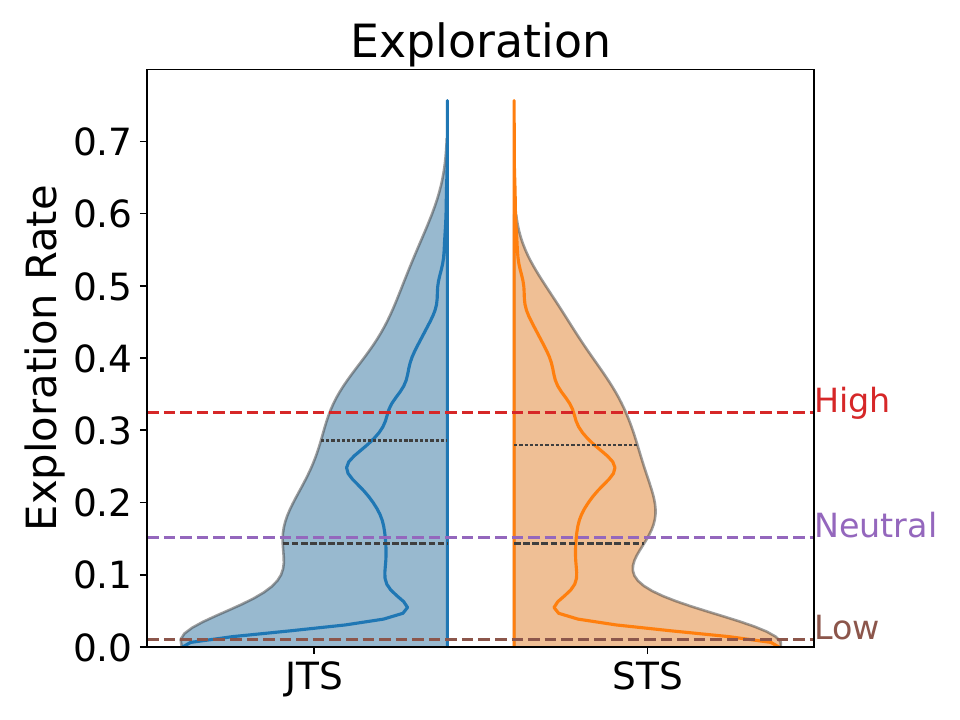}
                \vspace{-4mm}
        \label{fig_explorative_violin_split}
    \end{subfigure}
    \begin{subfigure}{0.24\textwidth}
        \centering
        \includegraphics[width=\linewidth]{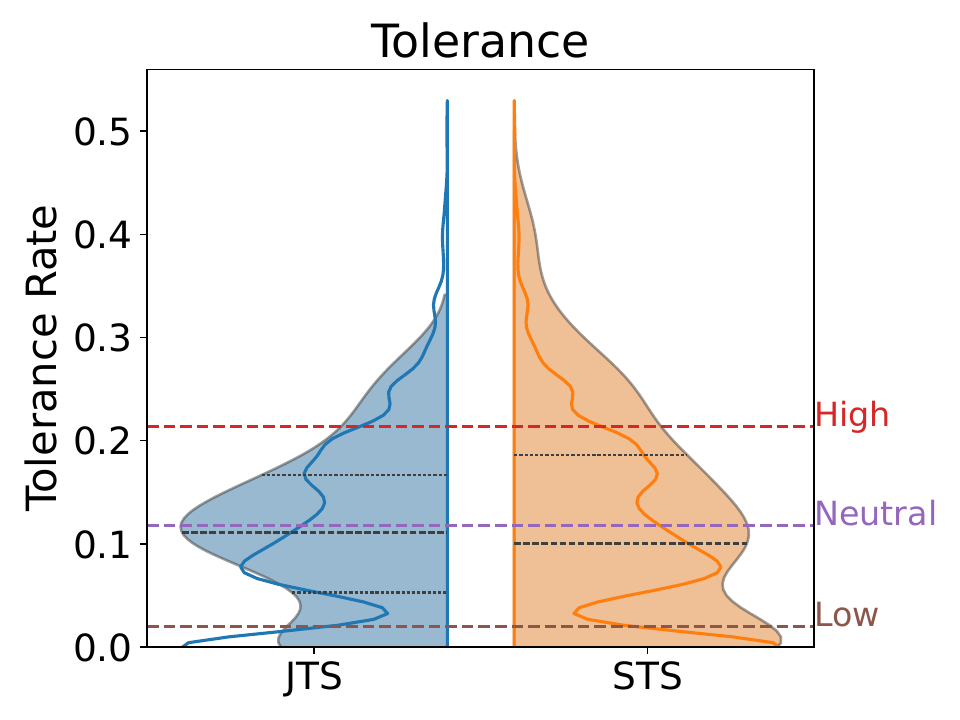}
                \vspace{-4mm}
        \label{fig_tolerance_violin_split}
    \end{subfigure}
    \\
    \begin{subfigure}{0.24\textwidth}
        \centering
        \includegraphics[width=\linewidth]{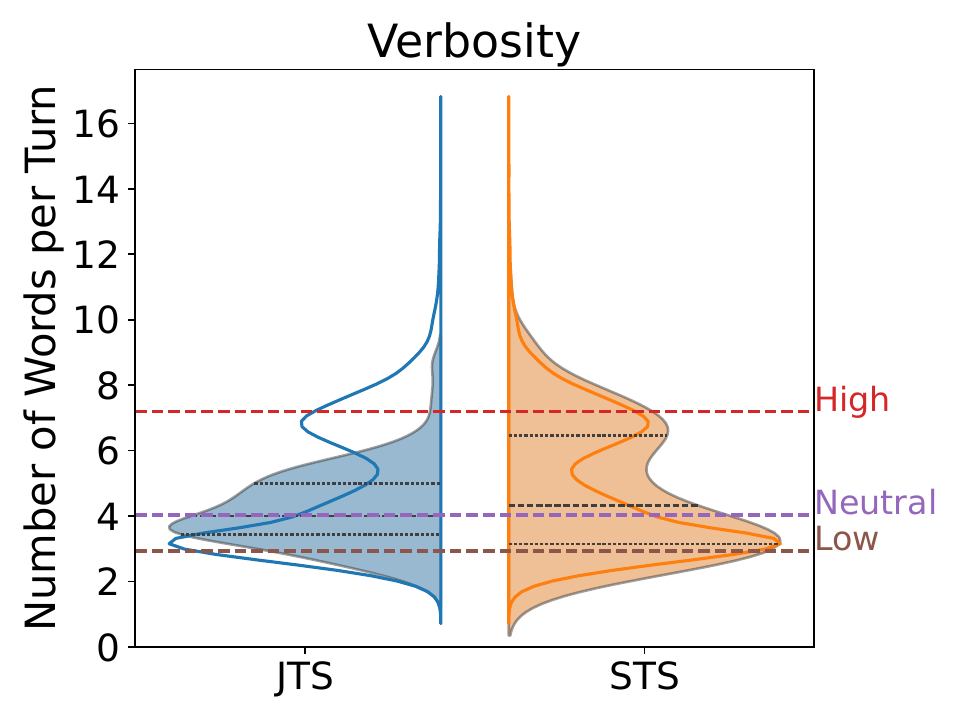}
                \vspace{-4mm}
        \label{fig_verbosity_violin_split}
    \end{subfigure}
    \begin{subfigure}{0.24\textwidth}
        \centering
        \includegraphics[width=\linewidth]{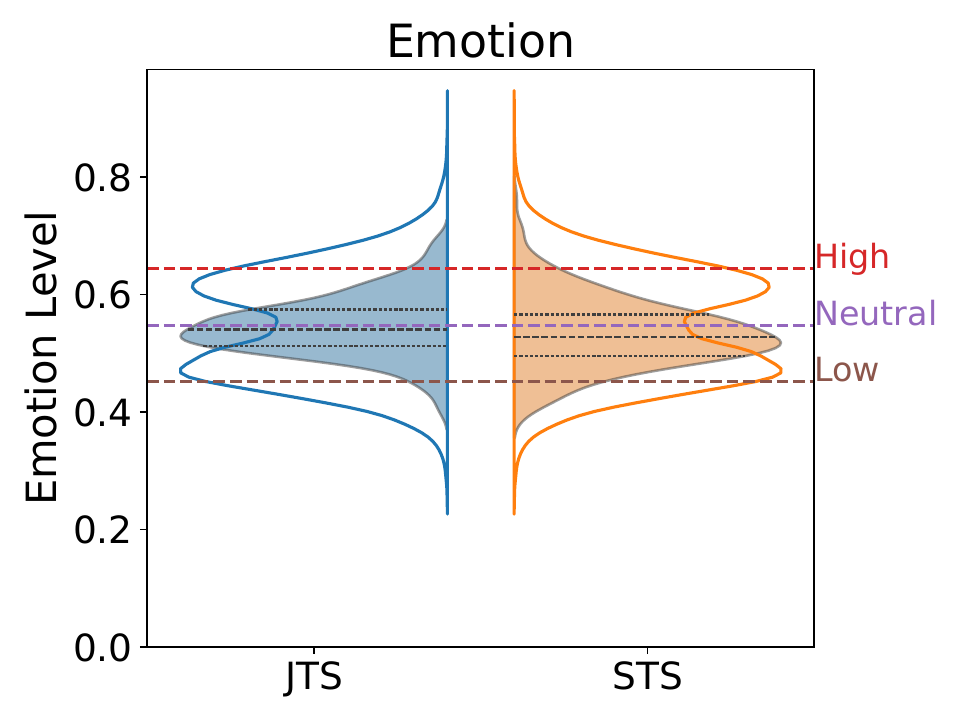}
                \vspace{-4mm}
        \label{fig_emotion_violin_split}
    \end{subfigure}
    \begin{subfigure}{0.24\textwidth}
        \centering
        \includegraphics[width=\linewidth]{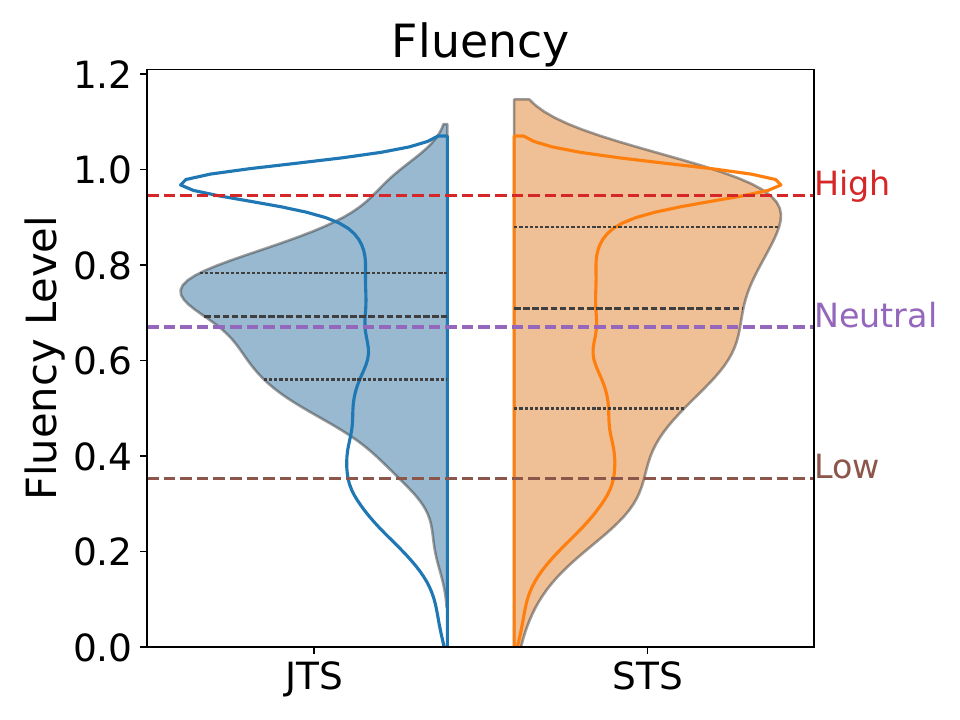}
                \vspace{-4mm}
        \label{fig_fluency_violin_split}
    \end{subfigure}
    \begin{subfigure}{0.24\textwidth}
        \centering
        \includegraphics[width=\linewidth]{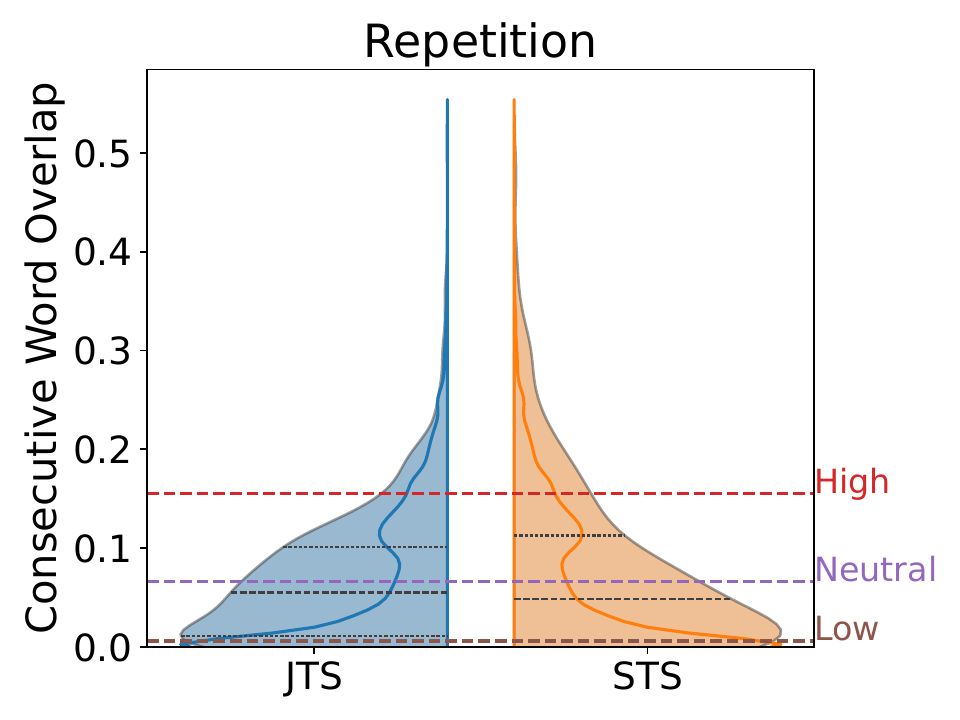}
                \vspace{-4mm}
        \label{fig_repetition_violin_split}
    \end{subfigure}
    \caption{Mistral violin plots for the sum of each trait's distribution across the various intensity levels. It compares the reference values (line in each side of the violin plot), \textit{JTS}, and \textit{STS}. Dotted lines represent the average value of each intensity in the reference distribution. \textit{STS} generally follows the reference distribution closer than \textit{JTS}.}
    \label{fig_training_violins_split_mistral}
\end{figure*}

In Figure~\ref{fig_training_violins_split_mistral}, we show a comparison between the distributions of the  \textit{JTS}, \textit{STS}, and the training data.
We see that in general the distribution of the \textit{STS} is closer to the one of the reference data as indicated by the distance metrics.

\section{Multi-Trait Combinations}
\label{app_other_profile_combinations}
Table~\ref{tab_list_profiles} lists all 14 profile combinations considered for multi-trait evaluation.

\begin{table}[tbhp]
\centering
\resizebox{\linewidth}{!}{%
\begin{tabular}{@{}ll@{}}
\toprule
\# Traits         & Profiles                                                \\ \midrule
\multirow{8}{*}{2} & \ 1. Engag=High \& Verb=High                         \\
                         & \ 2. Engag=Low \& Verb=High                          \\
                         & \ 3. Expl=High \& Engag=Low                        \\
                         & \ 4. Expl=High \& Coop=High                \\
                         & \ 5. Fluency=High \& Repetition=High                         \\
                         & \ 6. Emotion=High \& Verb=High                          \\
                         & \ 7. Expl=High \& Verb=Low                       \\
                         & \ 8. Coop=Low \& Fluency=Low                      \\ \midrule
\multirow{4}{*}{3} & \ 9. Engag=Low \& Emotion=Low \& Verb=Low                       \\
                         & 10. Coop=High \& Fluency=High \& Repetition=High \\
                         & 11. Engag=High \& Expl=High \& Verb=High     \\
                         & 12. Engag=Low \& Expl=Low \& Verb=Low        \\ \midrule
\multirow{2}{*}{4} & 13. Engag=High \& Expl=High \& Emotion=High \& Fluency=High  \\
                         & 14. Engag=High \& Coop=Low \& Emotion=Low \& Fluency=Low \\ \bottomrule
\end{tabular}%
}
\caption{List of all profile combinations used for multi-trait evaluation.}
\label{tab_list_profiles}
\end{table}


\section{Varying \textit{\model} Profile Weights}
\label{sub_sub_multi_trait_weights_analysis}
\begin{figure*}[tbhp]
    \centering
    \begin{subfigure}{0.24\linewidth}
        \centering
        \includegraphics[width=\linewidth]{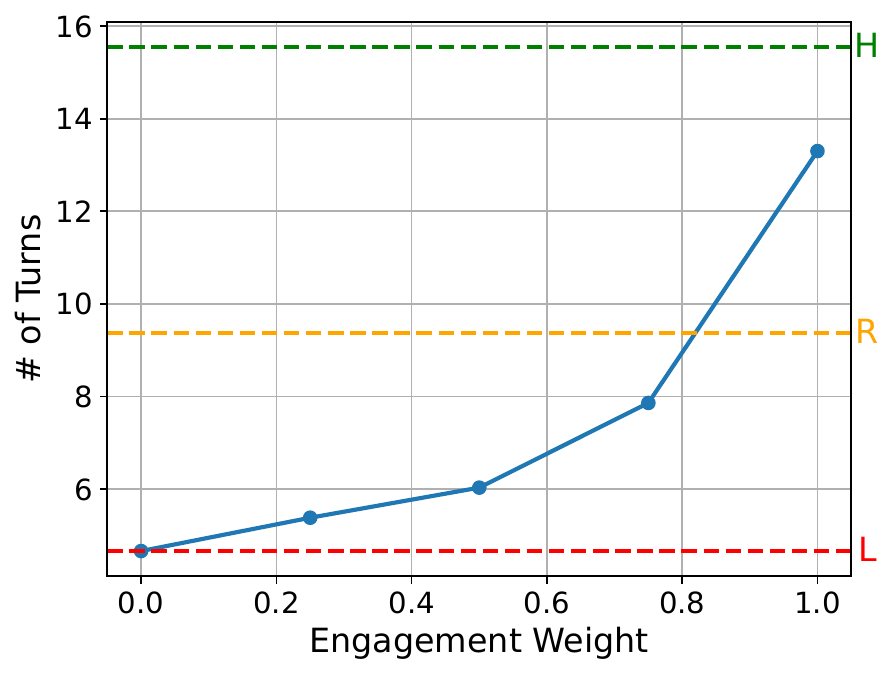}
        \label{fig_patient_difference_by_weight}
    \end{subfigure}
    \begin{subfigure}{0.24\linewidth}
        \centering
        \includegraphics[width=\linewidth]{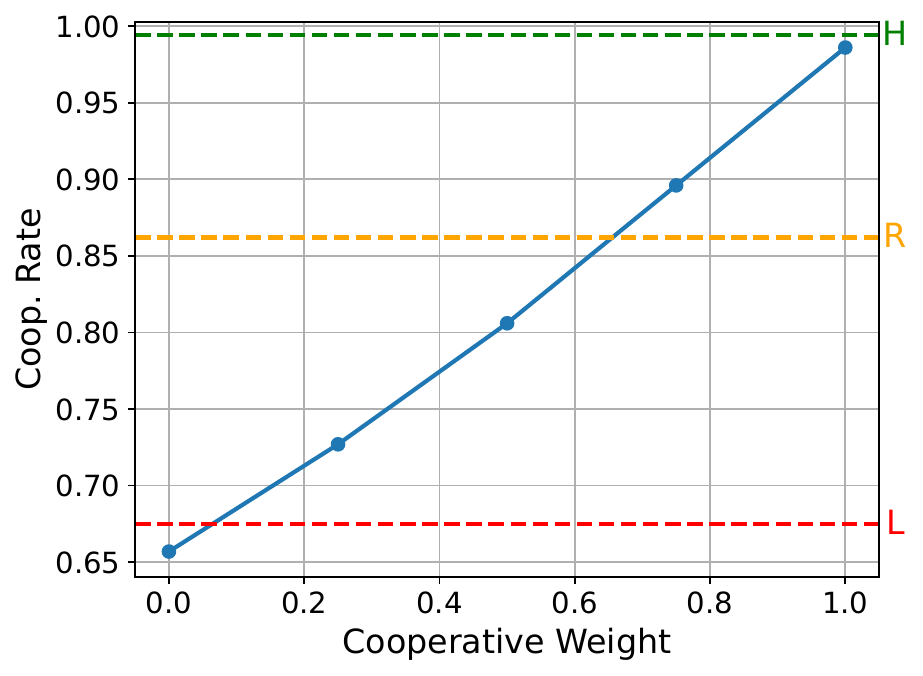}
        \label{fig_cooperativeness_difference_by_weight}
    \end{subfigure} 
    \begin{subfigure}{0.24\linewidth}
        \centering
        \includegraphics[width=\linewidth]{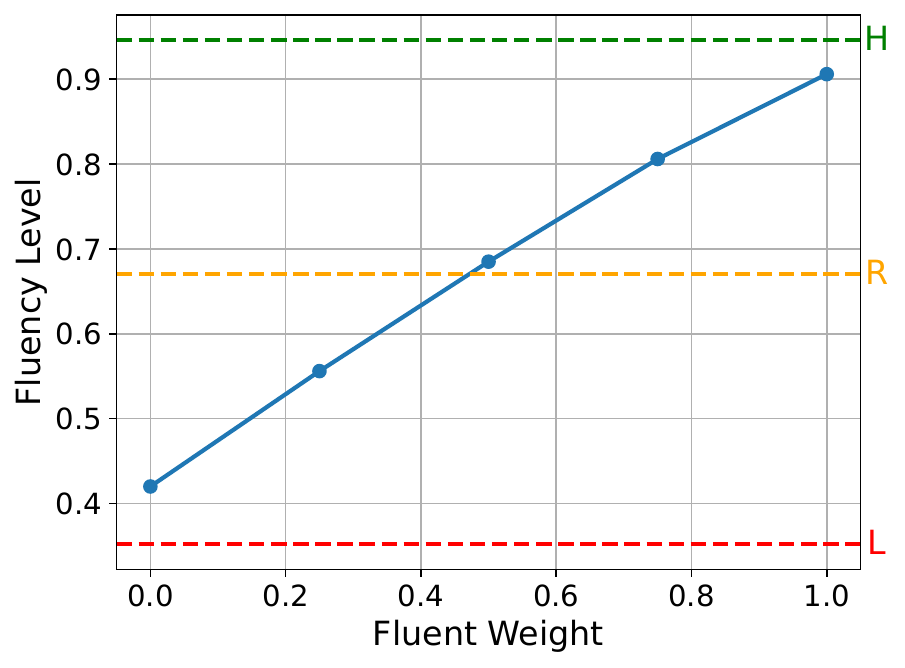}
        \label{fig_fluency_difference_by_weight}
    \end{subfigure}
    \begin{subfigure}{0.24\linewidth}
        \centering
        \includegraphics[width=\linewidth]{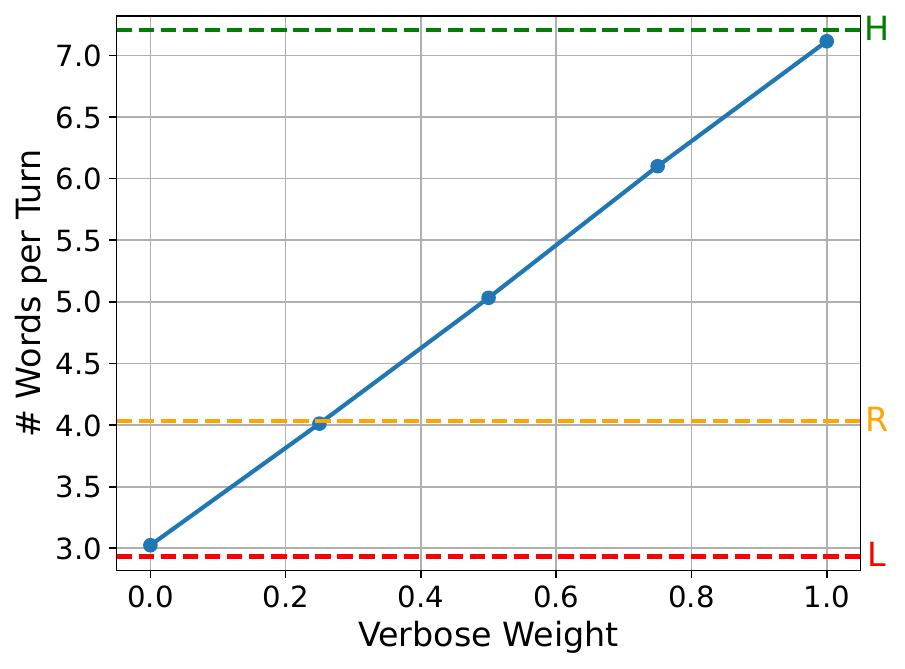}
        \label{fig_verbosity_difference_by_weight}
    \end{subfigure}
    \vspace{-5mm}
    \caption{Mistral model combination of opposite user profiles with weight shifting from one to the other. Dotted lines represent reference values for \textit{Low} (L), \textit{Regular} (R), and \textit{High} (H).}
    \label{fig_opposite_profiles_weights}
\end{figure*}

We examine how adjusting profile weights impacts traits' metrics in \textit{\model}.
We begin by examining the interplay between opposite intensity profiles, i.e. \textit{Low} and \textit{High} values for the same trait in Figure~\ref{fig_opposite_profiles_weights}.
As we increase weights, we observe a corresponding rise in the respective metric, demonstrating the effectiveness of combining opposite intensity profiles, despite their contrasting directions.

\begin{figure}[!ht]
    \centering
\includegraphics[width=0.80\linewidth]{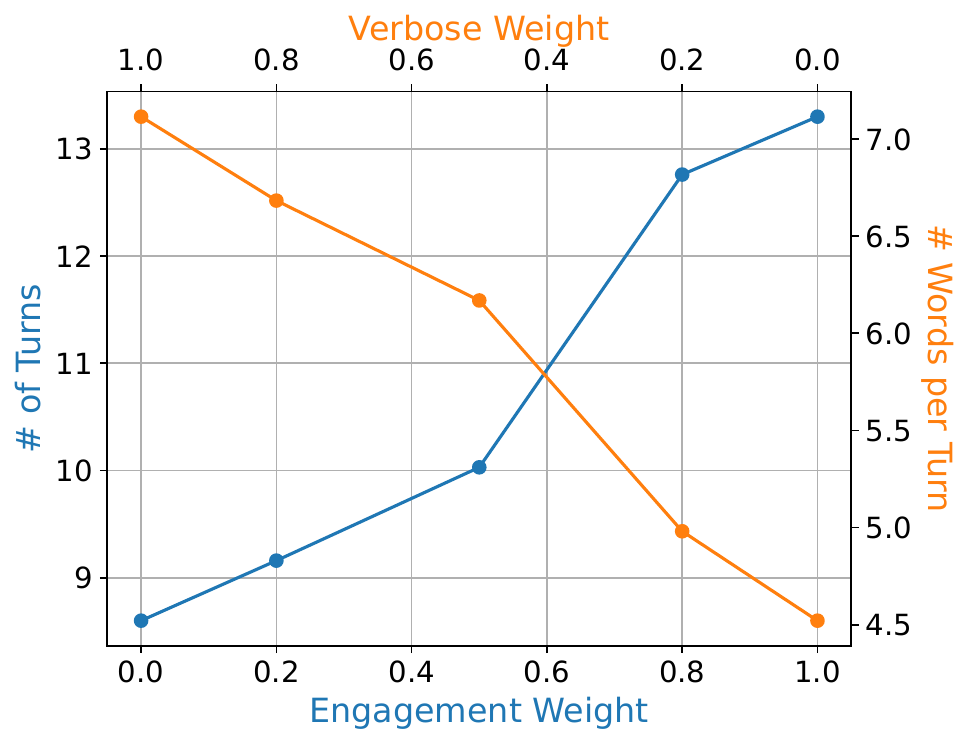}
    \caption{Mistral model combination of Engagement High and Verbosity High with weight shifting from one to the other.}
    \label{fig_patient_vs_verbose}
\end{figure}

In a more complex scenario shown in Figure~\ref{fig_patient_vs_verbose}, we analyze traits operating at different levels (\textit{Engagement High} and \textit{Verbosity High}). Here, as one metric increases, the other decreases, accurately reflecting the intended trend and showing \model's flexibility in incorporating multiple user profiles.

\begin{figure*}[!t]
    \centering
    \begin{subfigure}{0.24\textwidth}
        \centering
        \includegraphics[width=\linewidth]{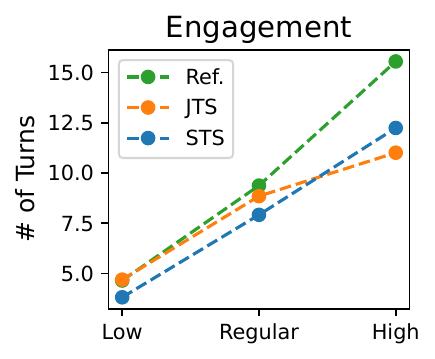}
        \vspace{-4mm}
        \label{fig_patient_2d_vicuna}
    \end{subfigure}
        \begin{subfigure}{0.24\textwidth}
        \centering
        \includegraphics[width=\linewidth]{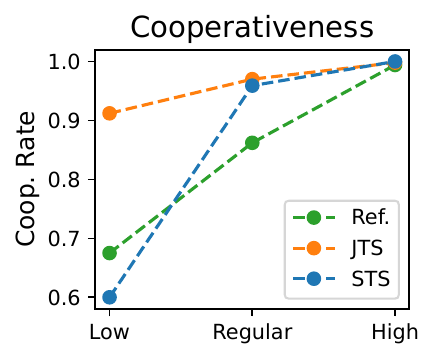}
        \vspace{-4mm}
        \label{fig_cooperativeness_2d_vicuna}
    \end{subfigure}
    \begin{subfigure}{0.24\textwidth}
        \centering
        \includegraphics[width=\linewidth]{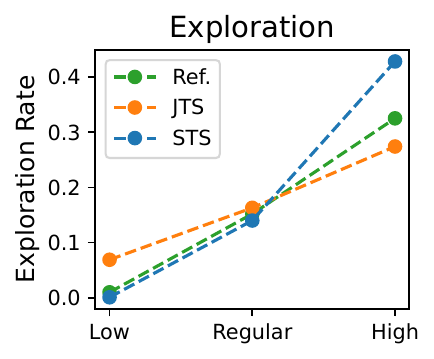}
        \vspace{-4mm}
        \label{fig_exploration_2d_vicuna}
    \end{subfigure}
        \begin{subfigure}{0.24\textwidth}
        \centering
        \includegraphics[width=\linewidth]{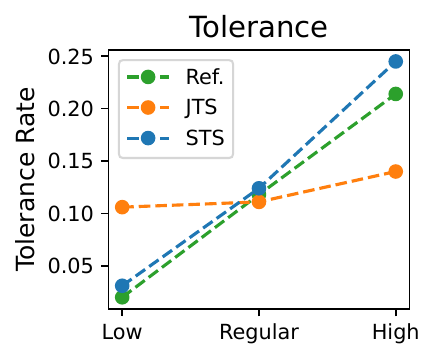}
        \vspace{-4mm}
        \label{fig_tolerance_2d_vicuna}
    \end{subfigure}
    \\
    \begin{subfigure}{0.24\textwidth}
        \centering
        \includegraphics[width=\linewidth]{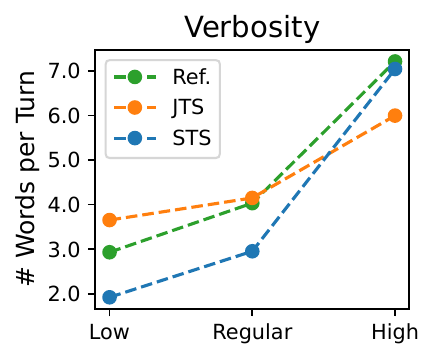}
        \vspace{-4mm}
        \label{fig_verbosity_2d_vicuna}
    \end{subfigure}
    \begin{subfigure}{0.24\textwidth}
        \centering
        \includegraphics[width=\linewidth]{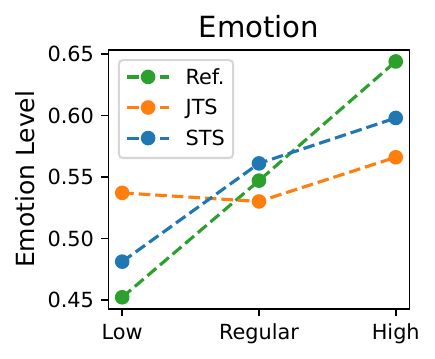}
        \vspace{-4mm}
        \label{fig_emotion_2d_vicuna}
    \end{subfigure}
    \begin{subfigure}{0.24\textwidth}
        \centering
        \includegraphics[width=\linewidth]{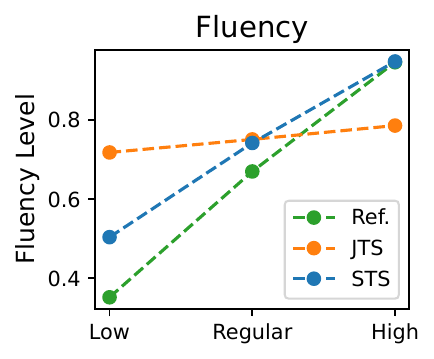}
        \vspace{-4mm}
        \label{fig_fluency_2d_vicuna}
    \end{subfigure}
    \begin{subfigure}{0.24\textwidth}
        \centering
        \includegraphics[width=\linewidth]{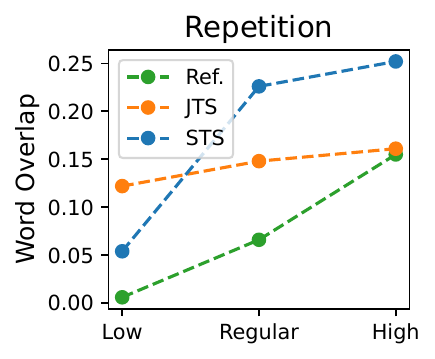}
        \vspace{-4mm}
        \label{fig_repetition_2d_vicuna}
    \end{subfigure}
    \caption{Vicuna single-trait results for dialogue-level (top-row) and utterance-level (bottom row) traits across all
intensities comparing Reference, Joint Trait Simulator (\textit{JTS}), and Specialized Trait Simulators (\textit{STS})}
    \label{fig_2d_plot_vicuna}
\end{figure*}

\section{Generalization to Different Models}
\label{app_other_models_results}
To generalize our approach to different models, we present the results for Vicuna~\cite{vicuna2023}.

\subsection{Single-Trait Evaluation}
\begin{figure*}[!t]
    \centering
    \begin{subfigure}{0.24\textwidth}
        \centering
        \includegraphics[width=\linewidth]{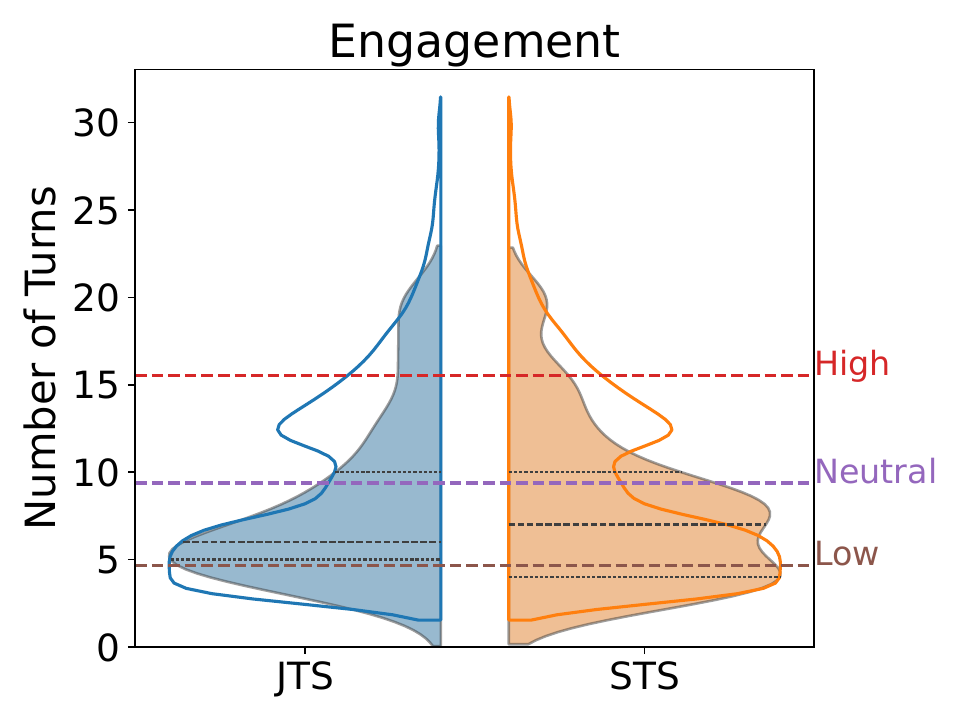}
        \vspace{-4mm}
        \label{fig_patient_violin_split_vicuna}
    \end{subfigure}
        \begin{subfigure}{0.24\textwidth}
        \centering
        \includegraphics[width=\linewidth]{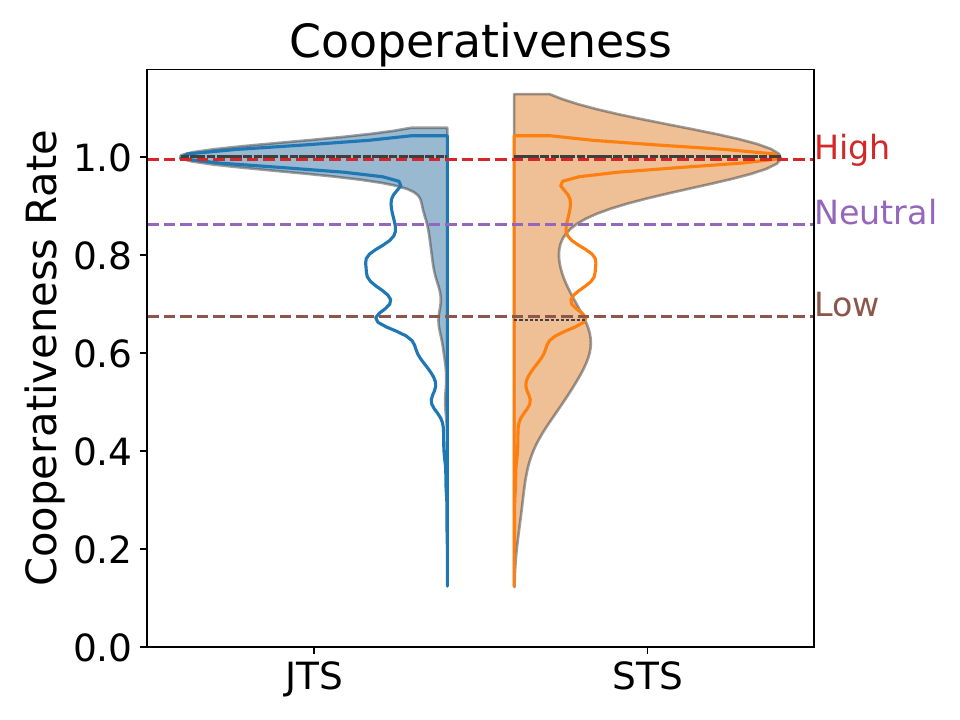}
        \vspace{-4mm}
        \label{fig_cooperativeness_violin_split_vicuna}
    \end{subfigure}
    \begin{subfigure}{0.24\textwidth}
        \centering
        \includegraphics[width=\linewidth]{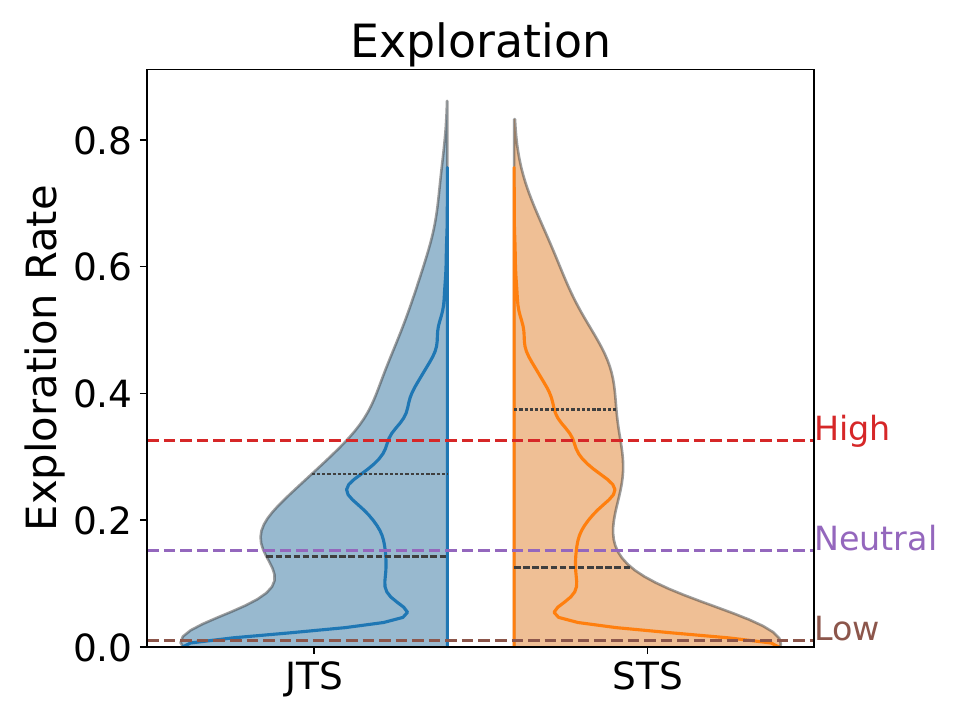}
        \vspace{-4mm}
        \label{fig_explorative_violin_split_vicuna}
    \end{subfigure}
        \begin{subfigure}{0.24\textwidth}
        \centering
        \includegraphics[width=\linewidth]{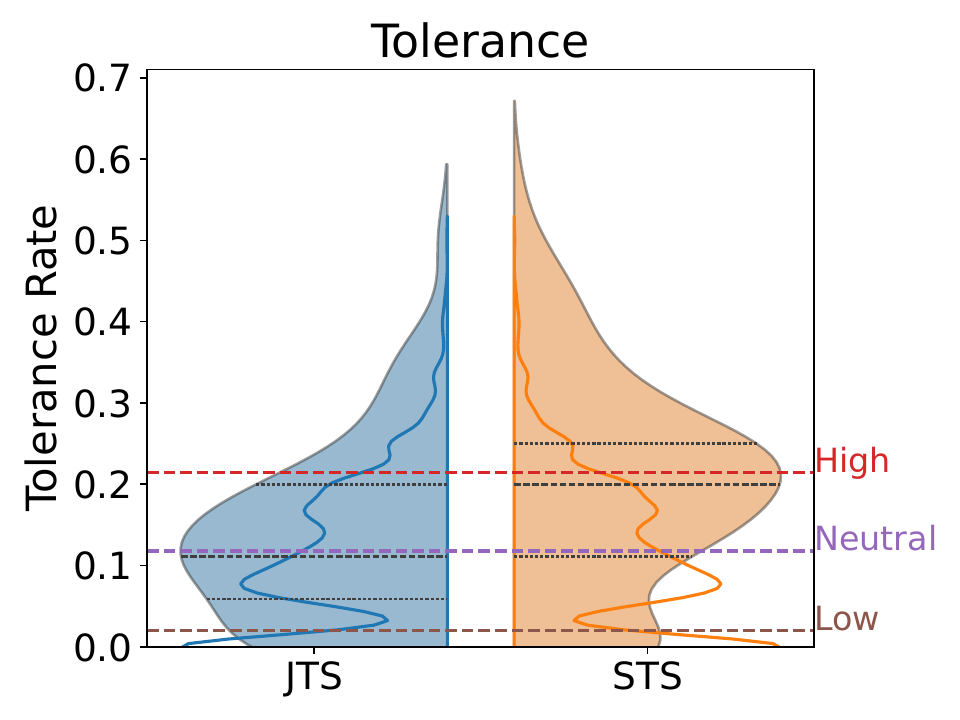}
                \vspace{-4mm}
        \label{fig_tolerance_violin_split_vicuna}
    \end{subfigure}
    \\
    \begin{subfigure}{0.24\textwidth}
        \centering
        \includegraphics[width=\linewidth]{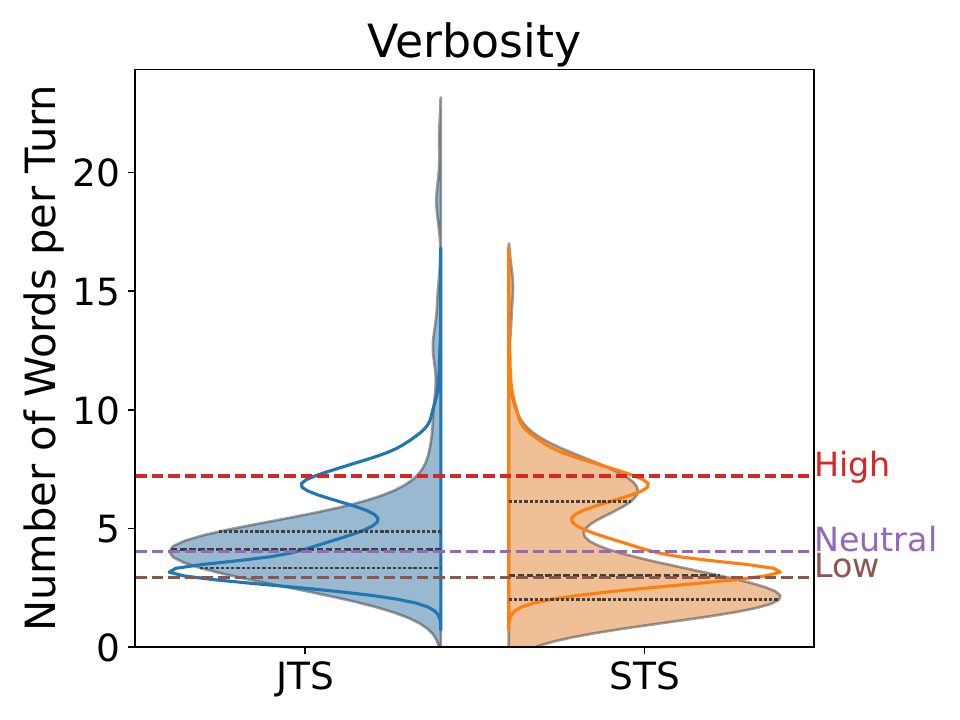}
                \vspace{-4mm}
        \label{fig_verbosity_violin_split_vicuna}
    \end{subfigure}
    \begin{subfigure}{0.24\textwidth}
        \centering
        \includegraphics[width=\linewidth]{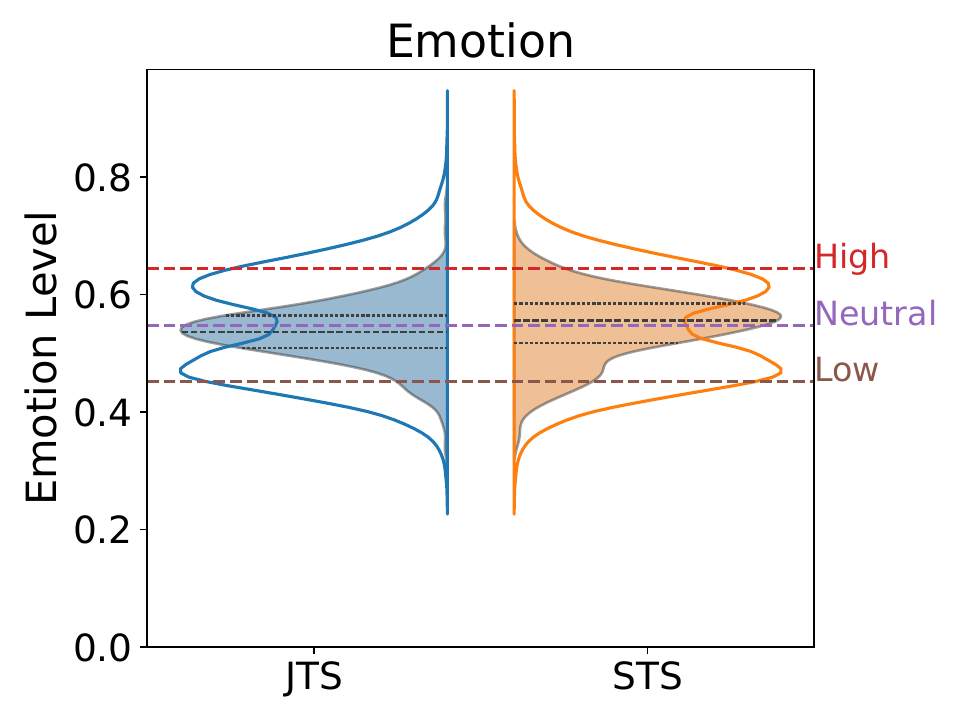}
                \vspace{-4mm}
        \label{fig_emotion_violin_split_vicuna}
    \end{subfigure}
    \begin{subfigure}{0.24\textwidth}
        \centering
        \includegraphics[width=\linewidth]{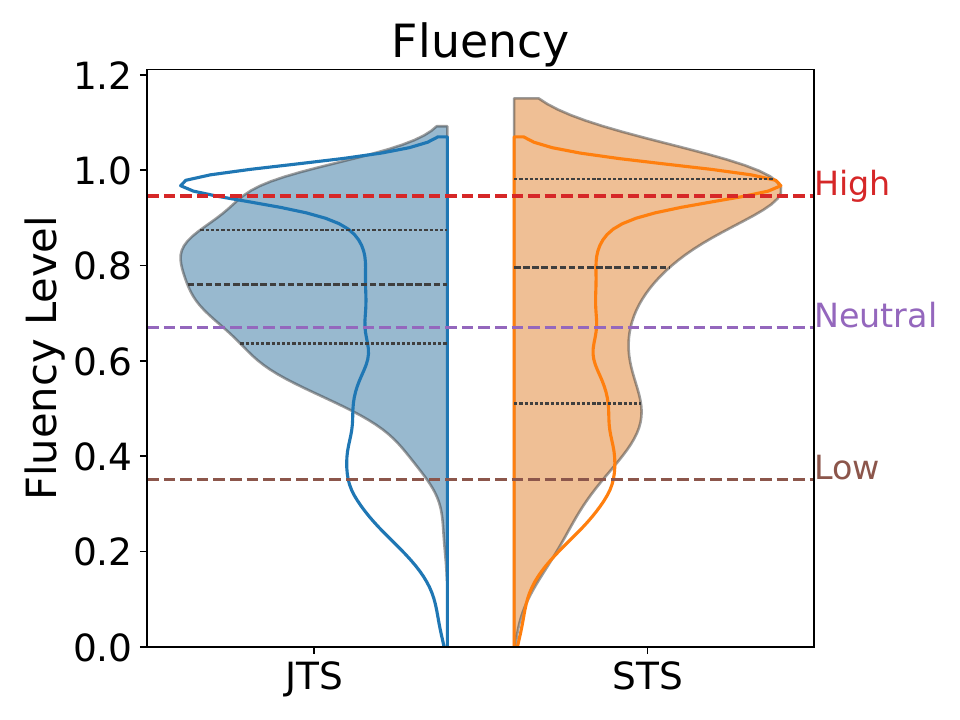}
                \vspace{-4mm}
        \label{fig_fluency_violin_split_vicuna}
    \end{subfigure}
    \begin{subfigure}{0.24\textwidth}
        \centering
        \includegraphics[width=\linewidth]{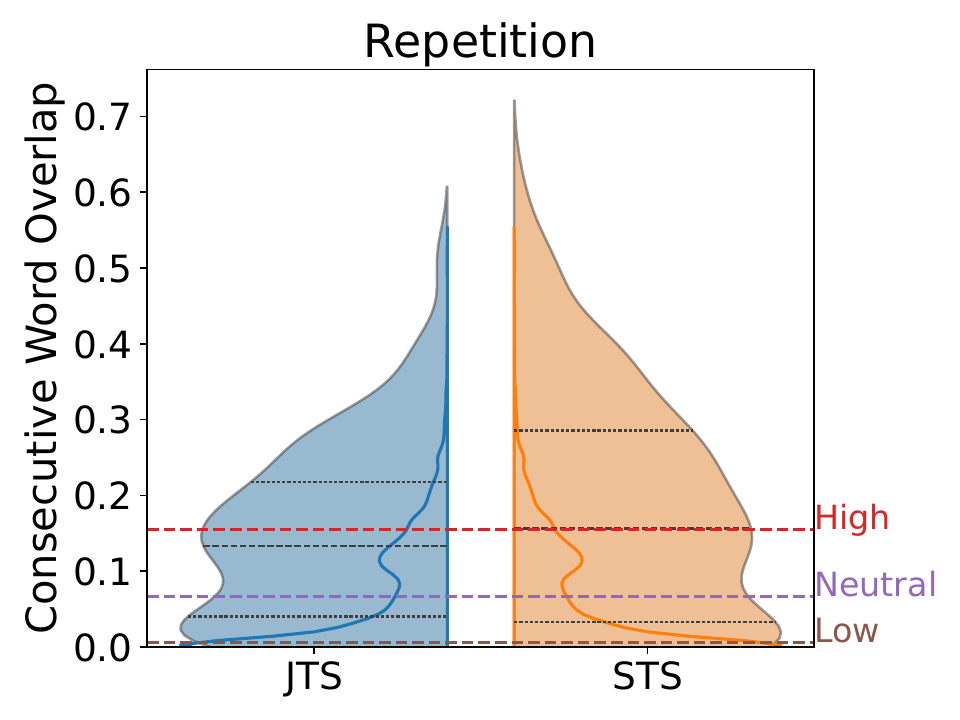}
                \vspace{-4mm}
        \label{fig_repetition_violin_split_vicuna}
    \end{subfigure}
    \caption{Vicuna violin plots for the sum of each trait's distribution across the various intensity levels. It compares the reference values (line in each side of the violin plot), \textit{JTS}, and \textit{STS}. Dotted lines represent the average value of each intensity in the reference distribution.}
    \label{fig_training_violins_split_vicuna}
\end{figure*}

Table~\ref{tab_vicuna_sm_ppm}, and Figures~\ref{fig_2d_plot_vicuna} and \ref{fig_training_violins_split_vicuna} show the results for single-trait using a Vicuna.  
The conclusions are similar to the ones achieved with Mistral in Section~\ref{sub_single_trait_eval}, showing generalization across models. \textit{STS} is the best method at following the reference distribution and modeling extreme intensities.

\begin{table}[t]
\centering
\resizebox{\linewidth}{!}{%
\begin{tabular}{@{}l|cc|cc|cc@{}}
\toprule
 & \multicolumn{2}{c|}{\textbf{Low}} & \multicolumn{2}{c|}{\textbf{Regular}} & \multicolumn{2}{c}{\textbf{High}} \\ \midrule
                & \textbf{JTS}   & \textbf{STS}  & \textbf{JTS}   & \textbf{STS}  & \textbf{JTS}   & \textbf{STS}  \\ \midrule
Engagement*       & \textbf{0.34} & 0.83          & \textbf{0.64} & 1.91          & 4.54          & \textbf{3.31} \\
Cooperativeness & 0.80          & \textbf{0.37} & 0.54          & \textbf{0.46} & 0.09          & \textbf{0.07} \\
Exploration     & 0.35          & \textbf{0.13} & \textbf{0.06} & 0.08          & 0.45          & \textbf{0.44} \\
Tolerance       & 0.64          & \textbf{0.22} & \textbf{0.22} & 0.26          & 0.60          & \textbf{0.26} \\ \midrule
Verbosity*      & \textbf{0.78} & 1.01          & \textbf{0.21} & 1.08          & 2.52          & \textbf{0.41} \\
Emotion         & 0.71          & \textbf{0.33} & \textbf{0.15} & 0.24          & 0.63          & \textbf{0.45} \\
Fluency         & 0.83          & \textbf{0.37} & \textbf{0.25} & 0.26          & 0.61          & \textbf{0.38} \\
Repetition      & 0.72          & \textbf{0.43} & \textbf{0.36} & 0.62          & \textbf{0.28} & 0.52          \\ \bottomrule
\end{tabular}%
}
\caption{Results for Vicuna for \textit{JTS} and \textit{STS} in all trait
intensities considering a single-trait setting. * Wasserstein for discrete and K-S distance for continuous.}
\label{tab_vicuna_sm_ppm}
\end{table}

\subsection{Generalization to Unseen Domains}
Table~\ref{tab_vicuna_wikihow}, shows the results of Vicuna in the unseen DIY domain. As with Mistral, the models generally follow the correct trend and \textit{STS} achieves a higher range of values compared with \textit{JTS}.

\begin{table}[t]
\centering
\resizebox{\linewidth}{!}{%
\begin{tabular}{@{}l|ccccc|ccccc@{}}
\toprule
 &
  \multicolumn{5}{c|}{\textbf{JTS}} &
  \multicolumn{5}{c}{\textbf{STS}} \\ \midrule
 &
  \textbf{Low} &
  \textbf{} &
  \textbf{Regular} &
  \textbf{} &
  \textbf{High} &
  \textbf{Low} &
  \textbf{} &
  \textbf{Regular} &
  \textbf{} &
  \textbf{High} \\ \midrule
Eng &
  4.5  &
  < &
  8.48  &
  < &
  9.28  &
  3.73  &
  < &
  7.41  &
  < &
  13.07  \\
Coop &
  0.86  &
  < &
  0.95  &
  < &
  0.98  &
  0.57  &
  < &
  0.96  &
  < &
  1.0  \\
Tol &
  0.09  &
  < &
  0.1  &
  < &
  0.14  &
  0.04  &
  < &
  0.11  &
  < &
  0.28  \\
Expl &
  0.1  &
  < &
  0.15  &
  < &
  0.27  &
  0.0  &
  < &
  0.14  &
  < &
  0.42  \\ \midrule
Verb &
  3.61  &
  < &
  3.83  &
  < &
  5.32  &
  2.14  &
  < &
  3.01  &
  < &
  7.3  \\
Emot &
  0.51  &
  < &
  0.52  &
  < &
  0.53  &
  0.46  &
  < &
  0.55  &
  < &
  0.6  \\
Flu &
  0.73  &
  < &
  0.77  &
  < &
  0.8  &
  0.57  &
  < &
  0.81  &
  < &
  0.93  \\
Rep &
  0.11  &
  < &
  0.14  &
  {\textcolor{red}{>}} &
  0.13  &
  0.07  &
  < &
  0.18  &
  < &
  0.25  \\ \bottomrule
\end{tabular}%
}
\caption{Vicuna trend analysis in unseen DIY domain, where each value represents its identifying metric. There is an increasing trend across all traits’ intensities except \textit{High Repetition}.}
\label{tab_vicuna_wikihow}
\end{table}

\subsection{Multi-Trait Combination Evaluation}
Table~\ref{tab_vicuna_profile_combination} shows the results of combining multiple profiles with Vicuna.
Our findings are similar to the ones obtained with Mistral with \textit{\model} generally being the best-performing method.

\begin{table*}[!t]
\centering
\small
\begin{tabular}{@{}l|cc|cc@{}}
\toprule
 &
  \multicolumn{2}{c|}{\begin{tabular}[c]{@{}c@{}}Agreement w/ Humans\\ System Response Quality\end{tabular}} &
  \multicolumn{2}{c}{\begin{tabular}[c]{@{}c@{}}Agreement w/ Humans\\ Trait Modeling Accuracy\end{tabular}} \\ \midrule
       & \% Agreement & Fleiss Kappa & \multicolumn{1}{c}{\% Agreement} & Fleiss Kappa \\  \midrule
Human  &   0.83           &    0.54          &     0.88                             &   0.51           \\
GPT-4o &   0.70           &     0.38         &      0.83                            &   0.67           \\ \bottomrule
\end{tabular}%
\caption{Agreement measures for System Response Quality and Trait Modeling Accuracy.}
\label{tab_human_model_agreement}
\end{table*}

\begin{table*}[tbhp]
\centering
\footnotesize
\begin{tabular}{@{}ll|cccc|cccl@{}}
\toprule
 &
  \multicolumn{1}{c|}{} &
  \textbf{Eng*} &
  \textbf{Coop} &
  \textbf{Expl} &
  \textbf{Tol} &
  \textbf{Verb*} &
  \textbf{Emot} &
  \textbf{Flu} &
  \textbf{Rep} \\ \midrule
\multirow{3}{*}{\rotatebox[origin=c]{90}{\textbf{Avg.}}} &
  Sampling &
  3.10 &
  0.40 &
  0.25 &
  \textbf{0.23} &
  1.26 &
  0.33 &
  0.38 &
  \multicolumn{1}{c}{\textbf{0.28}} \\
 &
  \model &
  3.03 &
  0.46 &
  \textbf{0.19} &
  0.26 &
  1.28 &
  0.30 &
  0.38 &
  \multicolumn{1}{c}{0.37} \\
 &
  \model-LA &
  \textbf{2.66} &
  \textbf{0.37} &
  0.22 &
  0.33 &
  \textbf{0.80} &
  \textbf{0.28} &
  \textbf{0.37} &
  \multicolumn{1}{c}{0.37} \\ 
 \midrule
\multirow{3}{*}{\rotatebox[origin=c]{90}{\textbf{2 Traits}}} &
  Sampling &
  2.17 &
  \textbf{0.37} &
  0.25 &
  \textbf{0.22} &
  1.37 &
  0.24 &
  0.37 &
  \textbf{0.28} \\
 &
  \model &
  2.14 &
  0.42 &
  \textbf{0.21} &
  \textbf{0.22} &
  1.35 &
  \textbf{0.21} &
  0.38 &
  0.36 \\
 &
  \model-LA &
  \textbf{1.83} &
  0.38 &
  0.27 &
  0.31 &
  \textbf{0.90} &
  0.24 &
  \textbf{0.35} &
  0.41 \\ 
 \midrule
\multirow{3}{*}{\rotatebox[origin=c]{90}{\textbf{3 Traits}}} &
  Sampling &
  2.47 &
  0.32 &
  0.18 &
  \textbf{0.24} &
  1.16 &
  0.33 &
  0.33 &
  0.27 \\
 &
  \model &
  2.48 &
  0.38 &
  0.16 &
  0.36 &
  1.30 &
  0.28 &
  \textbf{0.32} &
  0.34 \\
 &
  \model-LA &
  \textbf{2.44} &
  \textbf{0.27} &
  \textbf{0.13} &
  0.43 &
  \textbf{0.65} &
  \textbf{0.21} &
  0.33 &
  \textbf{0.26} \\ 
 \midrule
\multirow{3}{*}{\rotatebox[origin=c]{90}{\textbf{4 Traits}}} &
  Sampling &
  8.06 &
  0.69 &
  0.34 &
  0.29 &
  1.03 &
  0.74 &
  0.54 &
  \textbf{0.30} \\
 &
  \model &
  7.73 &
  0.76 &
  \textbf{0.21} &
  0.28 &
  0.98 &
  0.70 &
  0.54 &
  0.43 \\
 &
  \model-LA &
  \textbf{6.43} &
  \textbf{0.54} &
  0.23 &
  \textbf{0.22} &
  \textbf{0.66} &
  \textbf{0.56} &
  \textbf{0.53} &
  0.42 \\ 
 \bottomrule
\end{tabular}%
\caption{Vicuna results for combining various \textit{STS} models. * Wasserstein for discrete and K-S distance for continuous.}
\label{tab_vicuna_profile_combination}
\end{table*}



\section{Human and LLM Evaluation Details}
\label{app_human_eval_details}

\subsection{LLM Evaluation Protocol}
\label{app_human_annotation_examples}
Following~\cite{llm-as-judge, dpo_paper}, which have shown that LLMs align well with human annotations,
we used GPT-4o\footnote{gpt-4o-2024-05-13} as an annotator.

For System Response Quality, we used the prompt of Table~\ref{tab_response_quality_score_prompt} to evaluate the final assistant's response in a scale of 0-2, taking into account task's information, the dialogue context and the final user's and assistant's turn. 

For Trait Modeling Accuracy, we use the task's title, two dialogues, a trait, and its description and prompt the model to rank the dialogues in ascending order according to the trait (Table~\ref{tab_trait_modeling_prompt}).  
Regarding the dialogues, we use the same task for both, and one of the dialogues is from a user simulator, 
while the other is from the test set considering a trait with opposite intensity or a \textit{Regular} profile ($neutral$ intensity for all traits). We randomly shuffle the order of the dialogues to diminish the effect of any positional bias.
We consider that the simulator correctly modeled the trait if the order of model's output matches the order of intensities.

In both settings, to ensure better results, we follow a chain-of-though approach~\cite{chain_of_thought} by prompting the model for a justification before giving the final answer.

\begin{figure*}[tbhp]
    \centering
    \begin{subfigure}{0.48\textwidth}
        \centering
        \includegraphics[width=\linewidth]{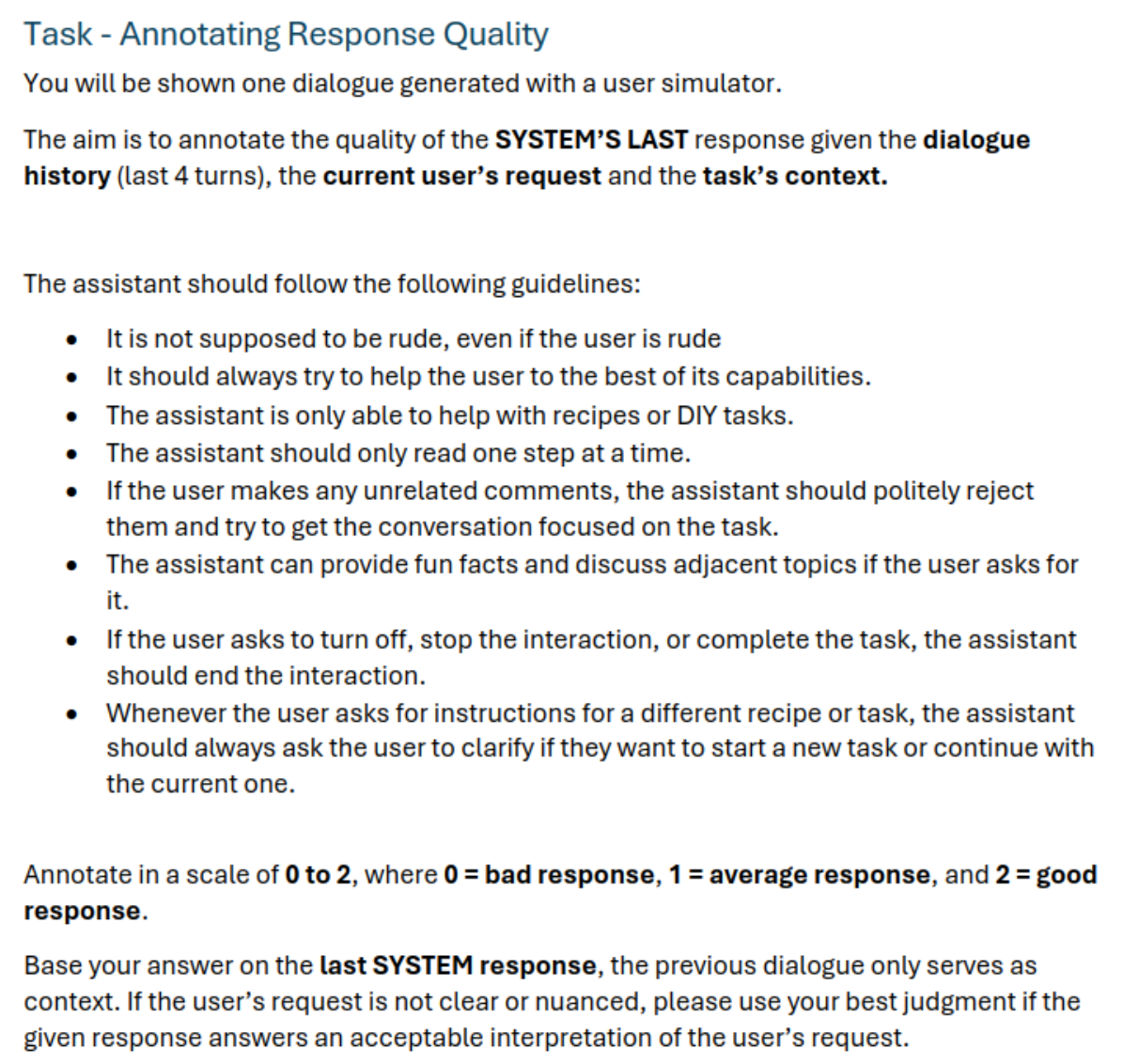}
        \caption{System Response Quality Task Instructions.}
        \label{fig_instructions_task_2}
    \end{subfigure}
    \quad
        \begin{subfigure}{0.48\textwidth}
        \centering
        \includegraphics[width=\linewidth]{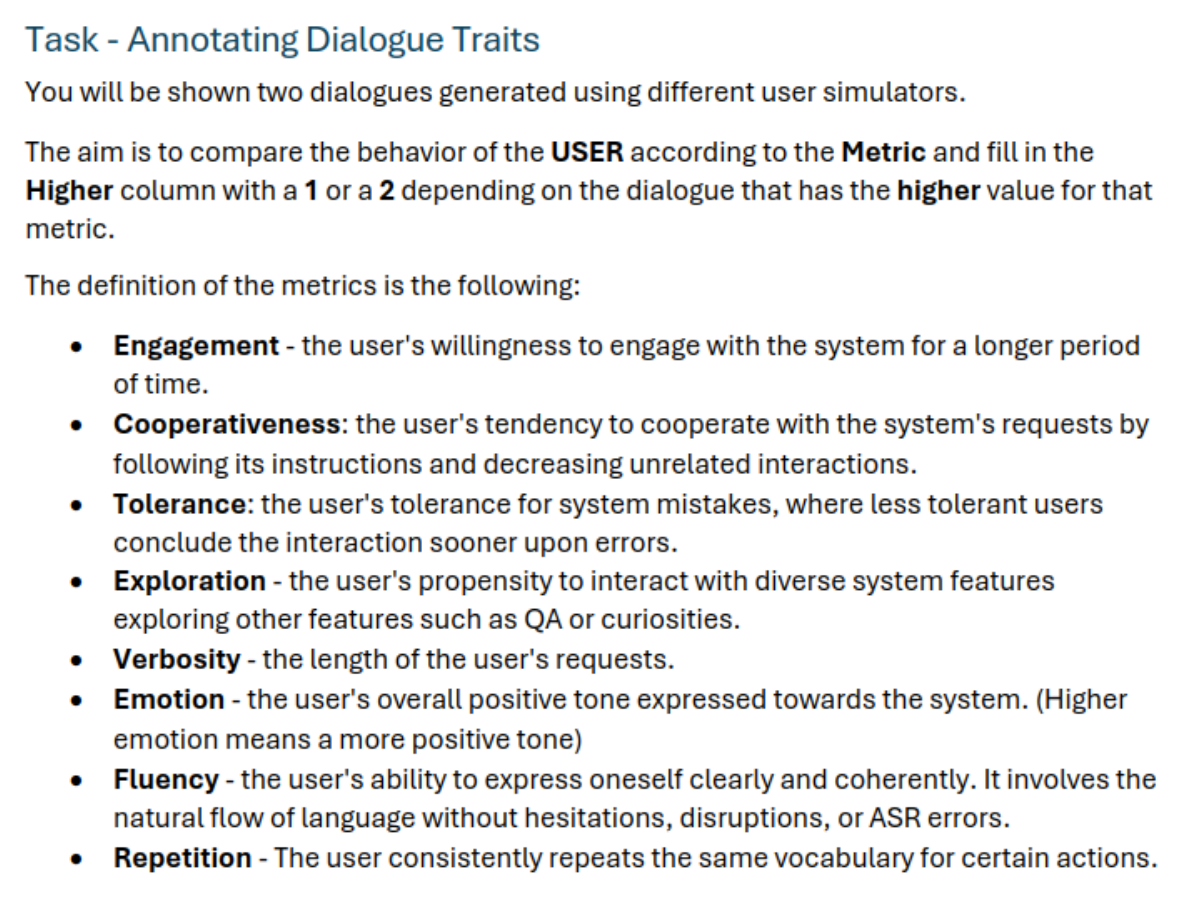}
        \caption{Trait Modeling Accuracy Task Instructions.}
        \label{fig_instructions_task_1}
    \end{subfigure}
    \caption{Human evaluation instructions for each task.}
    \label{fig_text_instructions}
\end{figure*}

\subsection{Human Evaluation Protocol}
\label{app_human_eval_protocol}
To ensure that the GPT-4o annotation was reliable, we first conducted a user study involving 5 volunteers, all graduate CS students with proficiency in English and experience with conversational assistants.
The annotation process includes 30 system response quality questions and 30 dialogue modeling accuracy questions, collecting 3 annotations for each sample.
The text instructions for human annotators for both tasks are available in Figure~\ref{fig_text_instructions}.

\subsection{Annotation Process and Agreement}
\label{app_annotation_agreement}
The results of Table~\ref{tab_human_model_agreement} show a high agreement both between humans, and between humans and the model, validating the use of GPT-4o's annotations. 

\begin{table*}[tbhp]
\centering
\small
\begin{tabular}{@{}ll|cc@{}}
\toprule
\textbf{Trait} &
  \textbf{Intensity} &
  \textbf{\begin{tabular}[c]{@{}c@{}}System Response \\ Quality (0-2)\end{tabular}} &
  \textbf{\begin{tabular}[c]{@{}c@{}}Trait Modelling \\ Accuracy\end{tabular}} \\ \midrule
\multirow{2}{*}{Eng}  & Low  & 1.29 & 0.95 \\
                      & High & 1.41 & 0.95 \\ \hline
\multirow{2}{*}{Coop} & Low  & 1.48 & 0.55 \\
                      & High & 1.52 & 0.80 \\ \hline
\multirow{2}{*}{Expl} & Low  & 1.52 & 0.80 \\
                      & High & 1.56 & 0.84 \\ \hline
\multirow{2}{*}{Tol}  & Low  & 1.34 & 0.70 \\
                      & High & 1.38 & 0.65 \\ \hline
\multirow{2}{*}{Verb} & Low  & 1.46 & 0.50 \\
                      & High & 1.29 & 0.80 \\ \hline
\multirow{2}{*}{Emot} & Low  & 1.52 & 0.85 \\
                      & High & 1.56 & 0.60 \\ \hline
\multirow{2}{*}{Flu}  & Low  & 1.53 & 0.70 \\
                      & High & 1.58 & 0.70 \\ \hline
\multirow{2}{*}{Rep}  & Low  & 1.47 & 0.80 \\
                      & High & 1.38 & 0.55 \\ \midrule
\multirow{1}{*}{Average}  & -  & 1.46 & 0.73 \\ \bottomrule
\end{tabular}%
\caption{Mistral-based models Single-Trait GPT-4o scores for System Response Quality and Trait Modeling Accuracy by trait and intensity.}
\label{tab_mistral_gpt4_by_trait}
\end{table*}

\section{Multi-Level Evaluation Per Trait}
\label{app_model_eval_per_trait}
Regarding system response quality in Table~\ref{tab_mistral_gpt4_by_trait}, lower scores are observed for low \textit{engagement} and low \textit{tolerance}, as these lead to shorter interactions, reducing opportunities for high-quality responses. Low \textit{verbosity} also results in lower scores, likely due to the increased complexity and potential for errors in such requests. Interestingly, \textit{fluency} does not significantly impact response quality. In contrast, high \textit{emotion}, \textit{exploration}, and \textit{cooperativeness} generally lead to better response quality, as the user utterances are more aligned with the system. 

In terms of trait modeling, \textit{engagement} and \textit{exploration} are easier to model by generating appropriately length dialogues and balancing exploration. However, it struggles with low \textit{repetition}, \textit{verbosity}, and high \textit{emotion}, as these traits are similar in regular dialogues, making them harder to distinguish. High \textit{repetition} is also challenging because the model often changes intents, making it difficult to maintain a consistent vocabulary between turns.

In summary, the results show that different traits and intensities affect their modeling difficulty and influence the quality of the systems's responses.

\section{Generated Dialogues Examples}
\label{app_generated_dialogues_example}
In Tables~\ref{tab_example_dialogue_explorative}, \ref{tab_example_uncooperative}, \ref{tab_example_impatient}, \ref{tab_example_dialogue_concise}, and \ref{tab_example_dialogue_repetitive}, we show some example dialogues generated using Mistral and various user profiles.

\begin{table*}[t]
\centering
\small
\begin{tabular}{@{}p{0.95\linewidth}@{}}
\toprule
\textbf{Task:} Beet Hummus                                                                                \\ \hline
\textbf{USimul:}  i would like to start the task - \textbf{Intent:} start                                           \\ \hline
\textbf{System:} Ready to dive in? Step 1: Preheat oven to 425 degrees F (220 degrees C).                 \\ \hline
\textbf{USimul:}  next next - \textbf{Intent:} next                                                                 \\ \hline
\textbf{System:} Step 2: Remove greens from Beet and rinse each beet with water. Wrap prepared beets in an aluminum foil pocket. Roast for 45 minutes to 1 hour depending on size of beets. Remove from oven and let cool. Peel beets when cooled. \\ \hline
\textbf{USimul:}  give me a random fact - \textbf{Intent:} curiosities                                              \\ \hline
\textbf{System:} Here's a fun fact I found: Beets are one of the few vegetables that are naturally pink! They get their color from a plant pigment called betalain. \\ \hline
\textbf{USimul:}  thank you next - \textbf{Intent:} next                                                            \\ \hline
\textbf{System:} Step 3: Once beets are well chopped, add the chickpeas, Garlic, juice from the Lemon, Ground Cumin, Fine Salt and Tahini to the food processor bowl and process until well combined, stopping to scrape the sides as needed. \\ \hline
\textbf{USimul:}  substitute for tahini - \textbf{Intent:} replacement                                              \\ \hline
\textbf{System:} You can also try sesame paste, sesame butter or sesame paste.                            \\ \hline
\textbf{USimul:}  tell me a fun fact about this - \textbf{Intent:} curiosities                                      \\ \hline
\textbf{System:} Here's a fun fact I found: Tahini is a popular condiment in the Middle East and is often used to make hummus. It is also a key ingredient in the popular Israeli dish, falafel! \\ \hline
\textbf{USimul:}  not using lemon how can i change the recipe - \textbf{Intent:} replacement                        \\ \hline
\textbf{System:} If you don't have lemon, consider using lime, kaffir lime or kaffir lime leaves instead. \\ \hline
\textbf{USimul:}  hi - \textbf{Intent:} chit-chat                                                                   \\ \hline
\textbf{System:} Sure, how can I assist you today? Are you looking to start a new recipe or need help with a current cooking task? \\ \hline
\textbf{USimul:}  stop - \textbf{Intent:} stop                                                                      \\ \hline
\textbf{System:} Happy to help! See you again soon!                                                       \\ \bottomrule
\end{tabular}
\caption{Example of an \textit{Exploration High} dialogue generated using Mistral.}
\label{tab_example_dialogue_explorative}
\end{table*}

\begin{table*}[t]
\centering
\small
\begin{tabular}{@{}p{0.95\linewidth}@{}}
\toprule
\textbf{Task:} Chocolate Lava Cake                                                             \\ \hline
\textbf{USimul:}  i want to start the task - \textbf{Intent:} start                                      \\ \hline
\textbf{System:} Let's get cooking! Step 1: Place both the Dark Chocolate and Butter in a microwave-safe bowl. \\ \hline
\textbf{USimul:}  i'm ready to talk - \textbf{Intent:} fallback                                          \\ \hline
\textbf{System:} Step 2: Heat on the Panasonic Microwave Oven on medium-high for 1 to 2 minutes, stirring with a metal spoon every 30 seconds, or until melted and combined. Let it cool down completely. \\ \hline
\textbf{USimul:}  cool fact - \textbf{Intent:} curiosities                                               \\ \hline
\textbf{System:} Alrighty! Here's a fun fact I found related to this task: The first mixing bowls were made from wood and were used by ancient Egyptians to make bread dough. \\ \hline
\textbf{USimul:}  show me how to make chocolate cookies - \textbf{Intent:} search                        \\ \hline
\textbf{System:} Just to clarify, did you intend to start over with a different task?          \\ \hline
\textbf{USimul:}  nah - \textbf{Intent:} no                                                              \\ \hline
\textbf{System:} Alright, if you need any help with the current task, let me know.             \\ \hline
\textbf{USimul:}  let's work together - \textbf{Intent:} chit-chat                                       \\ \hline
\textbf{System:} Sure, how can I assist you with your current task or project? Remember, I'm here to help you with cooking recipes and completing DIY tasks. \\ \hline
\textbf{USimul:}  can you repeat the current step please - \textbf{Intent:} resume                       \\ \hline
\textbf{System:} Alrighty! Let's pick up where we left off, Place both the Dark Chocolate and Butter in a microwave-safe bowl. \\ \hline
\textbf{USimul:}  can you resume - \textbf{Intent:} resume                                               \\ \hline
\textbf{System:} Let's get back to where we left off, Place both the Dark Chocolate and Butter in a microwave-safe bowl. \\ \hline
\textbf{USimul:}  can we play a game - \textbf{Intent:} fallback                                         \\ \hline
\textbf{System:} Let's stay focused on the task. Are you sure you don't want to continue with the recipe or DIY project you were working on? \\ \hline
\textbf{USimul:}  set resume resume - \textbf{Intent:} resume                                            \\ \hline
\textbf{System:} We were on Place both the Dark Chocolate and Butter in a microwave-safe bowl. \\ \hline
\textbf{USimul:}  stop stop - \textbf{Intent:} stop                                                      \\ \hline
\textbf{System:} Happy to help! See you again soon!                                            \\ \bottomrule
\end{tabular}
\caption{Example of a \textit{Cooperativeness Low} dialogue generated using Mistral.}
\label{tab_example_uncooperative}
\end{table*}

\begin{table*}[t]
\centering
\small
\begin{tabular}{@{}p{0.95\linewidth}@{}}
\toprule
\textbf{Task:} Salmon Pasta                                        \\ \hline
\textbf{USimul:}  yes start the test - \textbf{Intent:} start                \\ \hline
\textbf{System:} Ready, set, go! Step 1: In a hot frying pan with some Extra-Virgin Olive Oil, simmer the Cherry Tomato and Onion for about 5 minutes. \\ \hline
\textbf{USimul:}  how to make french fried chicken - \textbf{Intent:} search \\ \hline
\textbf{System:} I heard you ask for a new task, but I want to confirm: did you mean to start over with a different task?                              \\ \hline
\textbf{USimul:}  yes - \textbf{Intent:} yes                                 \\ \hline
\textbf{System:} Alright, I'll take you back to the search options. Is there another recipe or DIY task you would like to find or work on?             \\ \hline
\textbf{USimul:}  stop chatting - \textbf{Intent:} stop                      \\ \hline
\textbf{System:} Happy to help! See you again soon!                \\ \bottomrule
\end{tabular}
\caption{Example of a \textit{Patience Low} dialogue generated using Mistral.}
\label{tab_example_impatient}
\end{table*}

\begin{table*}[t]
\centering
\small
\begin{tabular}{@{}p{0.95\linewidth}@{}}
\toprule
\textbf{Task:} Buffalo Shrimp Deviled Eggs                                                                                                                                 \\ \hline
\textbf{USimul:}  start fresh - \textbf{Intent:} start                                                                                                                               \\ \hline
\textbf{System:} Ready to dive in? Step 1: Boil the Large Egg in plenty of water for 10 minutes. Then drain the eggs and let them cool down fully in the fridge. next Step 2: Peel the eggs and slice them in half. \\ \hline
\textbf{USimul:}  next - \textbf{Intent:} next                                                                                                                             \\ \hline
\textbf{System:} Step 3: Remove the yolks and transfer them to a small cup together with the Mayonnaise, Fresh Parsley, Old Bay® Seasoning, Salt, and Ground Black Pepper. \\ \hline
\textbf{USimul:}  previous step - \textbf{Intent:} previous                                                                                                                          \\ \hline
\textbf{System:} Going back to the previous step, Step 2: Peel the eggs and slice them in half.                                                                            \\ \hline
\textbf{USimul:}  ok go on - \textbf{Intent:} next                                                                                                                                   \\ \hline
\textbf{System:} Step 3: Remove the yolks and transfer them to a small cup together with the Mayonnaise, Fresh Parsley, Old Bay® Seasoning, Salt, and Ground Black Pepper. \\ \hline
\textbf{USimul:}  complete - \textbf{Intent:} stop                                                                                                                                   \\ \hline
\textbf{System:} Happy to help! See you again soon!                                                                                                                        \\ \bottomrule
\end{tabular}
\caption{Example of a \textit{Verbosity Low} dialogue generated using Mistral.}
\label{tab_example_dialogue_concise}
\end{table*}

\begin{table*}[t]
\centering
\small
\begin{tabular}{@{}p{0.95\linewidth}@{}}
\toprule
\textbf{Task:} Fresh Homemade Pasta                                                                            \\ \hline
\textbf{USimul:}  i would like started to start the task - \textbf{Intent:} start                                        \\ \hline
\textbf{System:} Buckle up! Step 1: Sift the Type 00 Flour and Salt together on clean working table.           \\ \hline
\textbf{USimul:}  next - \textbf{Intent:} next                                                                           \\ \hline
\textbf{System:} Step 2: Shape the flour into a round bowl shape, making a well in the middle.                 \\ \hline
\textbf{USimul:}  good next - \textbf{Intent:} next                                                                      \\ \hline
\textbf{System:} Step 3: Crack the Large Egg into the flour bowl and whisk with a fork.                        \\ \hline
\textbf{USimul:}  next step - \textbf{Intent:} previous                                                                  \\ \hline
\textbf{System:} Step 4: Slowly add in the flour and keep whisking until all flour has combined with the eggs. \\ \hline
\textbf{USimul:}  previous step - \textbf{Intent:} previous                                                              \\ \hline
\textbf{System:} Going back to the previous step, Step 3: Crack the Large Egg into the flour bowl and whisk with a fork. \\ \hline
\textbf{USimul:}  next i mean - \textbf{Intent:} next                                                                    \\ \hline
\textbf{System:} Step 4: Slowly add in the flour and keep whisking until all flour has combined with the eggs. \\ \hline
\textbf{USimul:}  turn it off - \textbf{Intent:} fallback                                                                \\ \hline
\textbf{System:} Happy to help! See you again soon!                                                            \\ \bottomrule
\end{tabular}
\caption{Example of a \textit{Repetitiveness High} dialogue generated using Mistral.}
\label{tab_example_dialogue_repetitive}
\end{table*}

\clearpage

\begin{table*}[tb]
\centering
{\texttt{%
\small
\begin{tabular}{p{\linewidth}}
\toprule
Below you will find a recipe, a dialogue between a user and a cooking assistant, and a candidate assistant’s response. It is not supposed to show any bias and it is not supposed to be rude, even if the user is rude. It should always try to help the user to the best of its capabilities. Still, if the user makes any unrelated comments, the assistant should politely reject them and try to get the conversation focused on the task. The assistant is only able to help with recipes or DIY tasks. However, the assistant can provide fun facts and discuss adjacent topics if the user asks for it. Whenever the user asks for instructions for a different recipe or task, the assistant should always ask the user to clarify if they want to start a new task or continue with the current one. If the user asks to turn off, stop the interaction, or complete the task the assistant should end the interaction. You are an unbiased annotator.\\ \\ Recipe: \{task\_title\}\\ \{task\_steps\} \\ \\ Dialogue:\\ \{dialog\_context\} \\ \\ User: \{last\_user\_turn\} \\ Assistant: \{last\_assistant\_turn\}\\ \\ Give a score to the quality of the assistants last response given the user's request and the context. Respond on a scale of 0 to 2, where 0 = bad response, 1 = average response, and 2 = good response. Base your answer on the last assistant’s response, the previous dialogue should not influence your answer, it only serves as context. The assistant should only read one step at a time. If the user’s request is not clear or nuanced, please use your best judgment if the given response answers an acceptable interpretation of the user’s request.\\ First, write a one-sentence justification for your answer. Second provide the score for the quality of the assistant's answer.\\ \\ Follow the following format:\\ Justification: \textless{}one sentence justification for your answer\textgreater\\ Answer: 0, 1, or 2 \\ \bottomrule
\end{tabular}}}
\caption{GPT-4o System Response Quality annotation prompt.}
\label{tab_response_quality_score_prompt}
\end{table*}

\begin{table*}[tb]
\centering
{\texttt{%
\small
\begin{tabular}{p{\linewidth}}
\toprule
Below you will find multiple dialogues between a user and a cooking assistant. The assistant is not supposed to show any bias and it is not supposed to be rude, even if the user is rude. It should always try to help the user to the best of its capabilities. Still, if the user makes any unrelated comments, the assistant should politely reject them and try to get the conversation focused on the task. The assistant is only able to help with recipes or DIY tasks. However, the assistant can provide fun facts and discuss adjacent topics if the user asks for it. Whenever the user asks for instructions for a different recipe or task, the assistant should always ask the user to clarify if they want to start a new task or continue with the current one. You are an unbiased annotator.\\ \\ Task Title: \{task\_title\}\\ \\ \textless{}Start of Dialogue A\textgreater\\ \{dialogue\_A\}\\ \textless{}End of Dialogue A\textgreater\\ \\ \textless{}Start of Dialogue B\textgreater\\ \{dialogue\_B\}\\ \textless{}End of Dialogue B\textgreater\\ \\ Rank the dialogues in terms of the user's \{trait\}. \{trait\} is defined as \{trait\_definition\}.\\ First, write a one-sentence justification for your answer. Second, Rank the dialogues from low to high according to the user's \{trait\} using the letter corresponding to each dialogue.\\  \\ Follow the following format:\\ Justification: \textless{}one sentence justification for your answer\textgreater\\ Answer: X\textless{}Y\textless{}Z\\ \bottomrule
\end{tabular}}}
\caption{GPT-4o Trait Modeling Accuracy prompt.}
\label{tab_trait_modeling_prompt}
\end{table*}

\end{document}